\documentclass[conference]{IEEEtran}
\IEEEoverridecommandlockouts
\usepackage{cite}
\usepackage{amsmath,amssymb,amsfonts}
\usepackage{algorithmic}
\usepackage{graphicx}
\usepackage{textcomp}
\usepackage{xcolor}
\def\BibTeX{{\rm B\kern-.05em{\sc i\kern-.025em b}\kern-.08em
    T\kern-.1667em\lower.7ex\hbox{E}\kern-.125emX}}

\newcommand{\rechecked}{REC$\checkmark$D}
\newcommand{\iou}{\mathit{IoU}}

\usepackage{subcaption}
\usepackage{url}
\usepackage{hyperref}
\usepackage{booktabs}
\usepackage{placeins}

\newcommand{\ie}{\textit{i.e.}}
\newcommand{\eg}{\textit{e.g.}}

\begin{document}

\makeatletter

\newcommand{\linebreakand}{%
  \end{@IEEEauthorhalign}
  \hfill\mbox{}\par
  \mbox{}\hfill\begin{@IEEEauthorhalign}
}

\title{From Label Error Detection to Correction:\\
A Modular Framework and Benchmark for \\ Object Detection Datasets
}

\author{
  \IEEEauthorblockN{1\textsuperscript{st} Sarina Penquitt$^\ast$\thanks{$^\ast$ Equal contribution.} }
  \IEEEauthorblockA{\textit{Institute of Computer Science} \\
    \textit{Osnabrück University}\\
    Osnabrück, Germany \\
    sarina.penquitt@uni-osnabrueck.de}
  \and
  \IEEEauthorblockN{1\textsuperscript{st} Jonathan Klees$^\ast$}
  \IEEEauthorblockA{\textit{Institute of Computer Science} \\
    \textit{Osnabrück University}\\
    Osnabrück, Germany \\
    jonathan.klees@uni-osnabrueck.de}
  \and
  \IEEEauthorblockN{3\textsuperscript{rd} Rinor Cakaj}
  \IEEEauthorblockA{\textit{Quality Match GmbH} \\
    Heidelberg, Germany \\
    rinor.cakaj@quality-match.com}
  \linebreakand % <------------- \and with a line-break
  \IEEEauthorblockN{4\textsuperscript{th} Daniel Kondermann}
  \IEEEauthorblockA{\textit{Quality Match GmbH} \\
    Heidelberg, Germany \\
    dk@quality-match.com}
  \and
  \IEEEauthorblockN{5\textsuperscript{th} Matthias Rottmann}
  \IEEEauthorblockA{\textit{Institute of Computer Science} \\
    \textit{Osnabrück University}\\
    Osnabrück, Germany \\
    matthias.rottmann@uni-osnabrueck.de}
  \and
  \IEEEauthorblockN{6\textsuperscript{th} Lars Schmarje}
  \IEEEauthorblockA{\textit{Quality Match GmbH} \\
    Heidelberg, Germany \\
    lars.schmarje@quality-match.com}
}

\maketitle

\begin{abstract}

Object detection has advanced rapidly in recent years, driven by increasingly large and diverse datasets. However, label errors often compromise the quality of these datasets and affect the outcomes of training and benchmark evaluations. Although label error detection methods for object detection datasets now exist, they are typically validated only on synthetic benchmarks or via limited manual inspection. How to correct such errors systematically and at scale remains an open problem. We introduce a semi-automated framework for label error correction called \rechecked~(Rechecked). Building on existing label error detection methods, their error proposals are reviewed with lightweight, crowd-sourced microtasks. We apply \rechecked\, to the class pedestrian in the KITTI dataset, for which we crowdsourced high-quality corrected annotations. We detect 18\% of missing and inaccurate labels in the original ground truth. We show that current label error detection methods, when combined with our correction framework, can recover hundreds of errors with little human effort compared to annotation from scratch. However, even the best methods still miss up to 66\% of the label errors, which motivates further research, now enabled by our released benchmark.

\end{abstract}

\section{Introduction}
\label{sec:intro}
% Teaser Figure
\begin{figure}[htbp]
    \centering

%\hfill
    \begin{subfigure}[b]{0.24\linewidth}
        \includegraphics[trim=26 0 13 0, clip, width=\textwidth]{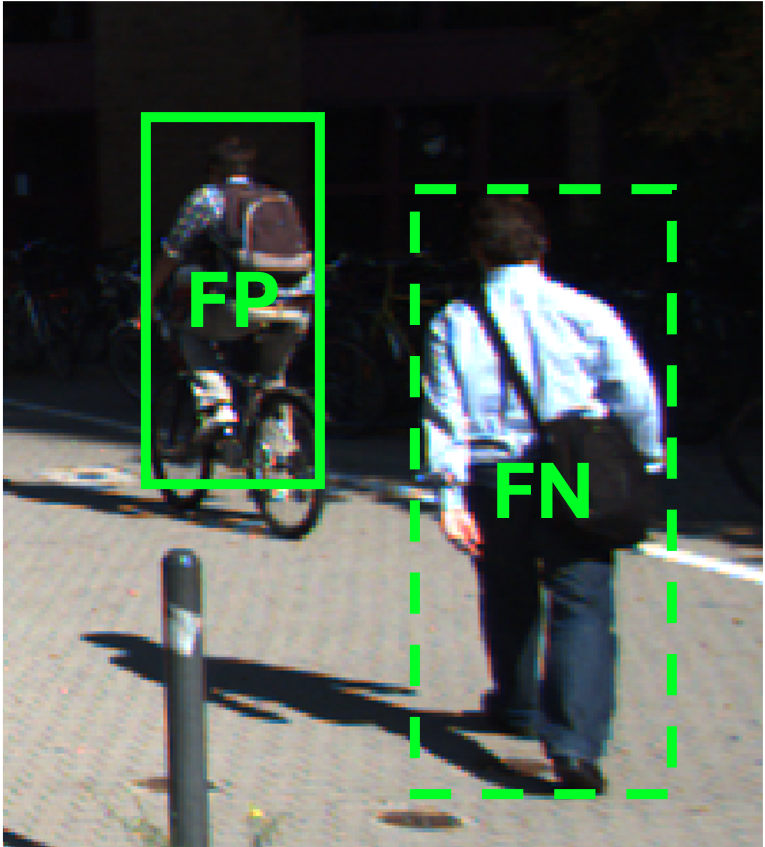}
        \caption{\centering Original Ground Truth}
    \end{subfigure}
    %~~
    %\hfill
    \begin{subfigure}[b]{0.24\linewidth}
        \includegraphics[trim=26 0 13 0, clip, width=\textwidth]{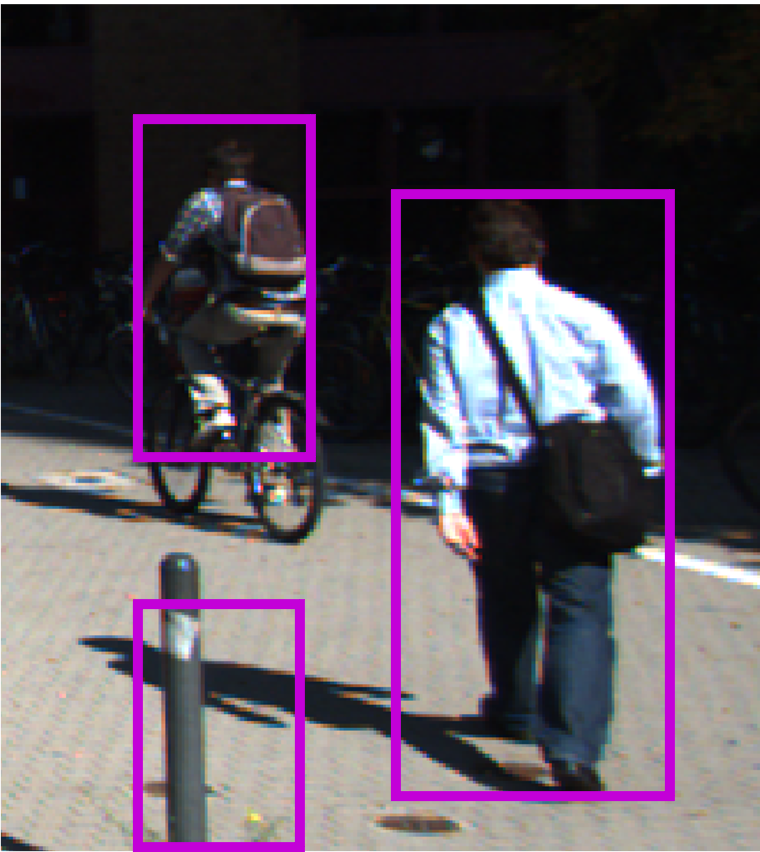}
        \caption{\centering Object Detection}
    \end{subfigure}
    %\\
    %\hfill
    %\hfill
    \begin{subfigure}[b]{0.24\linewidth}
        \includegraphics[trim=26 0 13 0, clip, width=\textwidth]{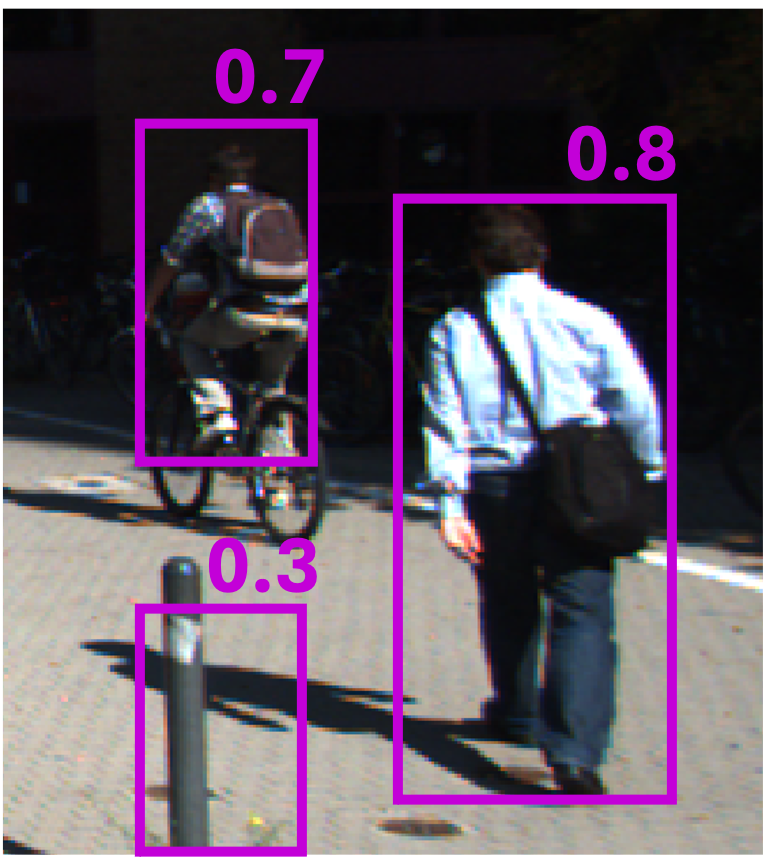}
        \caption{\centering Error Detection}
    \end{subfigure}
    %~~
     %\hfill
    \begin{subfigure}[b]{0.24\linewidth}
        \includegraphics[trim=26 0 13 0, clip, width=\textwidth]{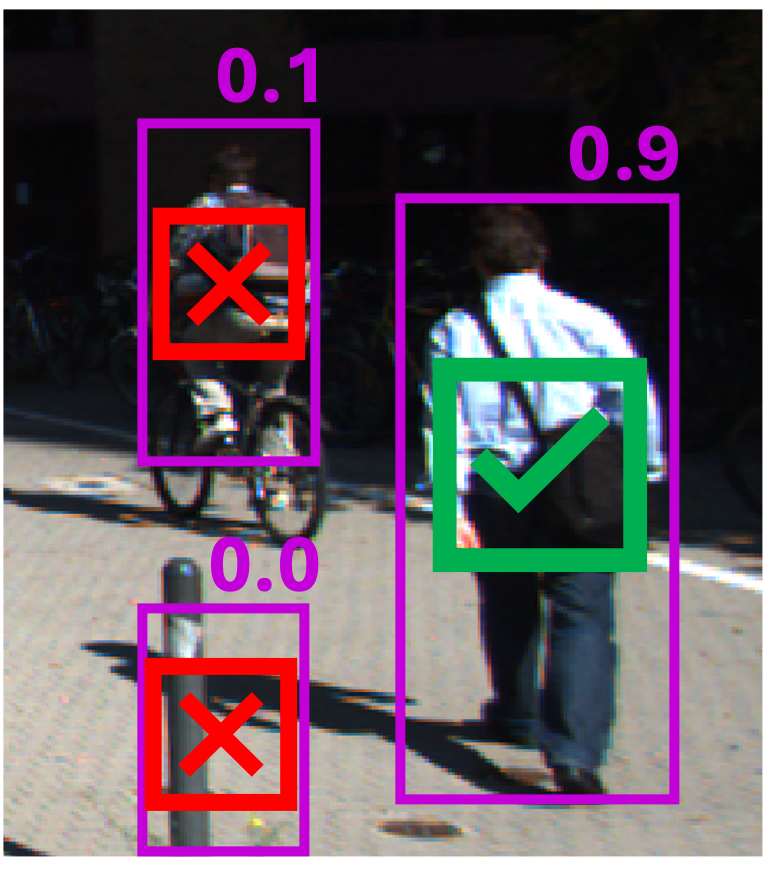}
        \caption{\centering Error Correction}
    \end{subfigure}

    \caption{
    Illustration of our framework \rechecked. (a) Original ground truth (GT) annotations in object detection datasets often contain errors such as missing bounding boxes (false negatives, FN) or incorrect extra boxes (false positives, FP). (b) Pretrained object detectors yield bounding box predictions. (c) These boxes are scored by a label error detection method estimating the probability for a label error. (d) Human microtasks are used to validate each box by multiple annotators, resulting in a soft label that can be used to correct label errors.
    }
    \label{fig:teaser}
\end{figure}

Deep neural networks (DNNs) have become state-of-the-art for extracting information from large-scale visual data \cite{kuutti2020survey,feng2020deep,lundervold2019mri,schmarje2021semi-supervised, hussain2018autonomous, oktay2018attention}. Training them for specialized and safety-critical domains, such as autonomous driving~\cite{kuutti2020survey, feng2020deep, hussain2018autonomous} or medical imaging~\cite{lundervold2019mri, oktay2018attention,schmarje2022mros,Kamnitsas2017Efficient}, requires vast amounts of high-quality annotated data \cite{sun2017revisiting, schmarje2023highquality, kaur2021survey, Agnew2024AnnotQual}.
Crowdsourcing offers a cost- and time-effective solution for building large-scale vision datasets~\cite{Kovashka2016Crowdsourcing} and enables scalable annotation~\cite{CrowdsourcingApplications, Marchesoni2023, Human-in-the-loop-systems}, particularly when tasks are broken down into simple, quick interactions known as \emph{microtasks}~\cite{CrowdsourcingApplications, Kovashka2016Crowdsourcing}.
The quality of labels is critical because label errors such as missing labels, incorrect classes, or poor localization degrade model performance, generalization capability~\cite{Song2023LearningFromNoisyLabels, Nahum2024LLM, jakubik2024Improve,Barragan2021QualityQuantity,deep_fish,noisy-labels-comparison,Wei2021cifar10n,Tomasev2022Relicv2} and benchmark validity~\cite{northcutt2021confident, Song2023LearningFromNoisyLabels}. Although recent research has proposed automated methods for detecting such errors in image classification~\cite{northcutt2021confident,northcutt2021pervasive, thyagarajan2022}, semantic segmentation~\cite{Rottmann2023AutomatedDetection}, and object detection~\cite{hu2022probability}, few works have focused on correcting them~\cite{northcutt2021pervasive, Jeon2025Active, Ma2022Correction, Kim2024Correction} and none use error detection methods to reduce the re-annotation costs.

We propose \rechecked~(see figure~\ref{fig:teaser}), a semi-automated framework that transitions from label error detection to correction for object detection datasets. \rechecked~combines automated label error proposals from existing methods~\cite{metadetect, loss-based-method, objectlab} with lightweight validation via microtasks.
Aggregated responses capture both label correctness and ambiguity, enabling scalable and cost-effective correction.
We apply \rechecked~to the class pedestrian in the KITTI dataset~\cite{kitti}.
The evaluation is intentionally limited to one class in order to establish proof of concept and control variables during the early stages of benchmarking. 
Our annotations reveal at least 18\% of previously unlabeled or incorrectly located pedestrian labels. These corrected annotations will be made publicly available as a new benchmark for evaluating methods for detecting and correcting label errors.
Our analysis shows that current detection methods can be rechecked faster than humans can draw the bounding boxes for labeling. However, up to 66\% of errors remain undetected, highlighting the need for improved label error detection strategies, which is now supported by our benchmark.

\noindent Our main contributions can be summarized as follows:
\begin{itemize}
\item We propose \rechecked{}, a scalable and cost-efficient framework that utilizes state-of-the-art label error detection methods and extends them to label error correction in object detection datasets, enabling the correction of hundreds of errors faster than humans can annotate the bounding boxes.
\item We introduce ambiguity-aware labeling through soft labels yielding high quality object labels. We demonstrate that lower quality data might introduce more errors than it is fixing. We release corrected pedestrian annotations for KITTI, enabling the evaluation of label error detection methods on real-world label noise.
\item With our framework, we detected at least 18\% of missing and inaccurate pedestrian annotations. 
However, we also find that up to 66\% of the errors were not identified by current label error correction methods, which highlights the importance of this research, now enabled by our benchmark.
\end{itemize}
Our source code is publicly available at \\
\url{https://github.com/JonathanKlees/rechecked}.

\section{Related Work}
\label{sec:related work}

\emph{Identifying Label Errors in Object Detection Datasets.} In recent years, there has been a research focus on label error detection in object detection. Cleanlab's tool ObjectLab~\cite{objectlab} assesses the quality of labels in object detection datasets by comparing predicted bounding boxes with GT, and provides a score describing the quality of the labels on an image. Another method, based on the instance-wise object detection loss~\cite{loss-based-method}, uses a two-stage object detector to detect real label errors by monitoring the regression and classification losses and using the discrepancy between predictions and ground truth labels. MetaDetect~\cite{metadetect} is a method that performs meta classification and meta regression for predictions. Potential label errors are those predicted boxes that are evaluated as false positive ($\iou<0.5$ with GT), but the meta classifier estimates a high probability for the prediction being correct. The method LidarMetaDetect~\cite{Lmd2025IJCV} is an extension of MetaDetect that detects label errors in 3D object detection datasets. \cite{Bär2023Noisy} developed a label refinement network that improves noisy localization labels. 
While we build on this recent research, we go one step further and aim at correcting these labels instead of just detecting them.

\emph{Label Error Correction.} In addition to detecting label errors, correcting them has also become an important step in improving label quality. \cite{Bernhardt2022Clean} developed an active label cleaning framework to prioritize samples for re-annotation by comparing estimated label correctness and label difficulty. CROWDLAB~\cite{goh2022crowdlab} is another tool of Cleanlab that classifies crowdsourced labels for refinement to identify, which samples require re-annotation. In the work of~\cite{Ma2022Correction}, the datasets COCO~\cite{lin2014microsoft} and Google Open Images~\cite{GoogleData} were re-annotated to compare model performance on original and re-annotated data. \cite{Kim2024Correction} combine foundation models with label corrections provided by human annotators to create PASCAL+, an updated version of the Pascal VOC dataset for semantic segmentation and\cite{Yun2021Correction} applied their re-labeling strategy, \emph{ReLabel}, to re-annotate the ImageNet training set and published a corrected version. 
Our work focuses on the process of label error correction based on error detection, with the aim of reducing annotation costs.
 
\emph{Ambiguity and Soft Labels in Computer Vision Datasets.} In common computer vision benchmark datasets, images or objects are usually associated with a single hard class label.
However, recent research shows that this concept is not representative of the ambiguity in real data \cite{tailception,mammo-variability,schoening2020Megafauna,medicinecrowdsource,Vasudevan2022DoughBagel,planktonUncertain,Davani2022BeyondMajority,Basile2021ConsiderDisagree,schmarje2022dc3,Wei2021cifar10n} and even harms model performance \cite{Barragan2021QualityQuantity,deep_fish,noisy-labels-comparison,schmarje2022benchmark,Wei2021cifar10n,Tomasev2022Relicv2}.
In contrast to this, soft or probabilistic labels~\cite{Peng2014LearningOP} indicate a probability distribution across several classes, reflecting that the correct label for an instance is not always straightforward and multiple annotators may disagree on the label. Soft labels in object detection datasets are rare but there exist methods that simulate soft labels based on single-class annotations through deep learning~\cite{Li2024SoftLabelsOD, Gao2024PSTTL, Nguyen2022LabelDistillation}.
We extend this research by showing that the label quality is essential for correcting labels and poor quality data might even introduce more errors than it corrects. 
Furthermore, we release these soft labels as a benchmark for future label error detection and correction research.

\section{Method \rechecked}
\label{sec:method}
The most challenging task in label error detection is detecting missing labels, \ie, objects present in the image that are missing in the GT annotations. Unlike inaccurate bounding box localizations and incorrect class labels, which only require verifying existing annotations and consequently, are easier to identify, detecting missing labels requires a complete review of each image. We propose a semi-automatic framework that combines label error detection methods with a microtask-based review process to efficiently correct label errors. 
As depicted in figure~\ref{fig:method-graph}, our framework relies on the predictions of object detectors, which are then evaluated through label error detection methods resulting in a much smaller number of label error proposals to be manually verified and corrected.

\begin{figure}[t]
    \centering
    \includegraphics[width=\linewidth]{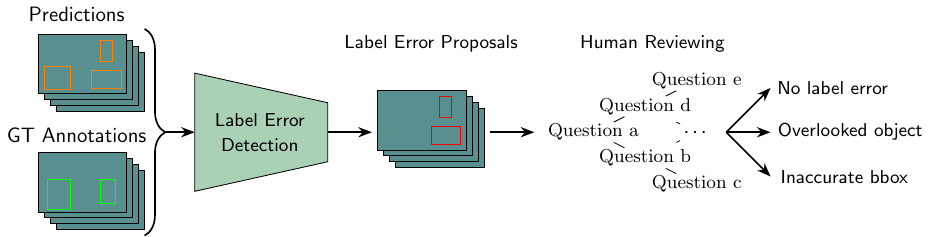}
    \caption{Overview of our suggested workflow \rechecked~to detect and correct label errors in object detection datasets.}
    \label{fig:method-graph}
\end{figure}

\subsection{Label Error Detection Methods}
\label{subsec:method, label error detection methods}
\rechecked~is model-agnostic. Any object detector can be used to infer initial predictions. In our experiments, we use YOLOX~\cite{yolox2021}, a one-stage detector, and Cascade R-CNN~\cite{CascadeRCNN}, a two-stage detector. 
We combine them with existing label error detection methods including MetaDetect~\cite{metadetect}, loss-based instance-wise scoring~\cite{loss-based-method} and ObjectLab~\cite{objectlab}.
The considered methods compare the predictions of an object detector with the GT annotations to identify overlooked objects or misfitting bounding boxes. The predictions are then scored by the label error detection method, reflecting the likelihood of a label error being present. This way, we obtain label error proposals, which are then reviewed manually in the second stage.

\begin{figure}[!ht] % tb -> different layout
     \centering
     \scalebox{1.0}{\includegraphics[width=\linewidth]{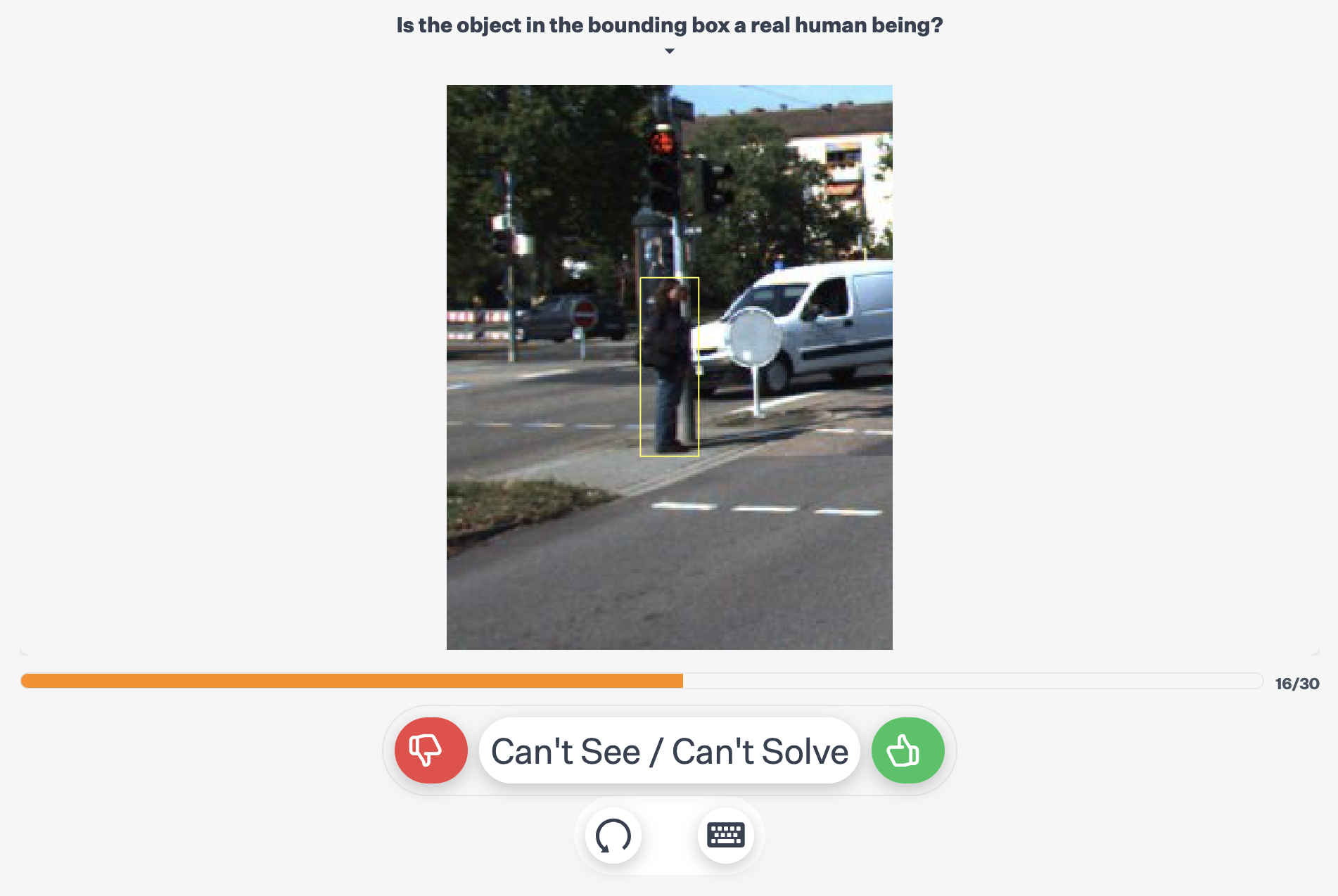}}
\caption{Microtask interface for verifying whether the object in the bounding box is a real human being. The interface shows only the relevant region and minimal surrounding context. A clear highlight guides the annotator’s attention, and no unnecessary elements are shown that could distract from the decision.}
     \label{fig:example_nanotask}
 \end{figure}
\subsection{Error Correction}
\label{subsec:method, nano-task correction design}

From the previous stage, we obtain a collection of label error proposals that need to be manually evaluated. We created short and easy-to-answer questions consisting of a single visual task, \eg, whether the highlighted object is a human or not, as illustrated in figure~\ref{fig:example_nanotask}. These kinds of micro-questions are also referred to as microtasks~\cite{GoodNWS15}. With microtasks, annotators can quickly answer one simple question at a time.
We repeat the microtasks with several people, which results in an estimate of the underlying distribution of the outcomes associated to this question.
More complex questions, such as 'What is a pedestrian?', are broken down into multiple microtasks, such as 'Is it a human?' and 'What is the person doing?'. See \ref{subsec:experimental setup, validated gt generation} for the microtasks used to construct the validated annotations of the KITTI dataset. 
We can use the calculated soft label probability to determine whether a proposed label error is indeed erroneous or not.

\section{Experimental Setup}
\label{sec:experimental setup}
In the following, we describe the dataset used in our study, the object detectors evaluated, the design of our microtasks, and the metrics employed for evaluation.

\subsection{Dataset Construction}
\label{subsec:experimental setup, datasets}

We apply \rechecked~to the KITTI~\cite{kitti} 2D object detection dataset.
We focus on the class pedestrian as a challenging and safety-relevant use case. 
The  7,481 training images include annotations for the classes: car, van, truck, pedestrian, person (sitting), cyclist, tram and misc. In addition, some regions are labeled as 'don't care' to mark areas with unlabeled objects.
Since no GT annotations are provided for the test set, we randomly split the annotated training set into training (80\%) and validation (20\%) subsets. To ensure representative coverage, we repeated the sampling process until approximately 80\% of all GT pedestrian annotations were assigned to the training set and 20\% to the validation set. The resulting dataset consists of 5,984 training images with 3,591 pedestrian objects and 1,497 validation images with 896 pedestrian annotations.

\begin{table}[tb]
\caption{Bounding box annotation strategies. The number of aggregations refers to how many annotator-provided boxes were combined into the final box. The combined strategy merges results from both methods. Costs are reported as average per box in seconds. $\dagger$ Cost for keypoint annotation alone is 11.12 s. $\ddagger$ Manual effort for removal of duplicate boxes is not included.}
\centering
\resizebox{\linewidth}{!}{  
\begin{tabular}{lccc}
\toprule
 & \textbf{\# Bounding Boxes} &  \textbf{\# Aggregations} & \textbf{Costs [s]} \\
\midrule
Direct box & 2833 & 1.0 & 44.11 \\
Keypoint-to-box &  2099 & 3.0 &  92.67$\dagger$ \\
Combined box  & 3078 & 2.4 & 103.79$\ddagger$ \\
\bottomrule
\end{tabular}
}

\label{tab:box_strategy}
\end{table}

\begin{figure}[tb]
     \centering
     \scalebox{1.0}{\includegraphics[width=\linewidth]{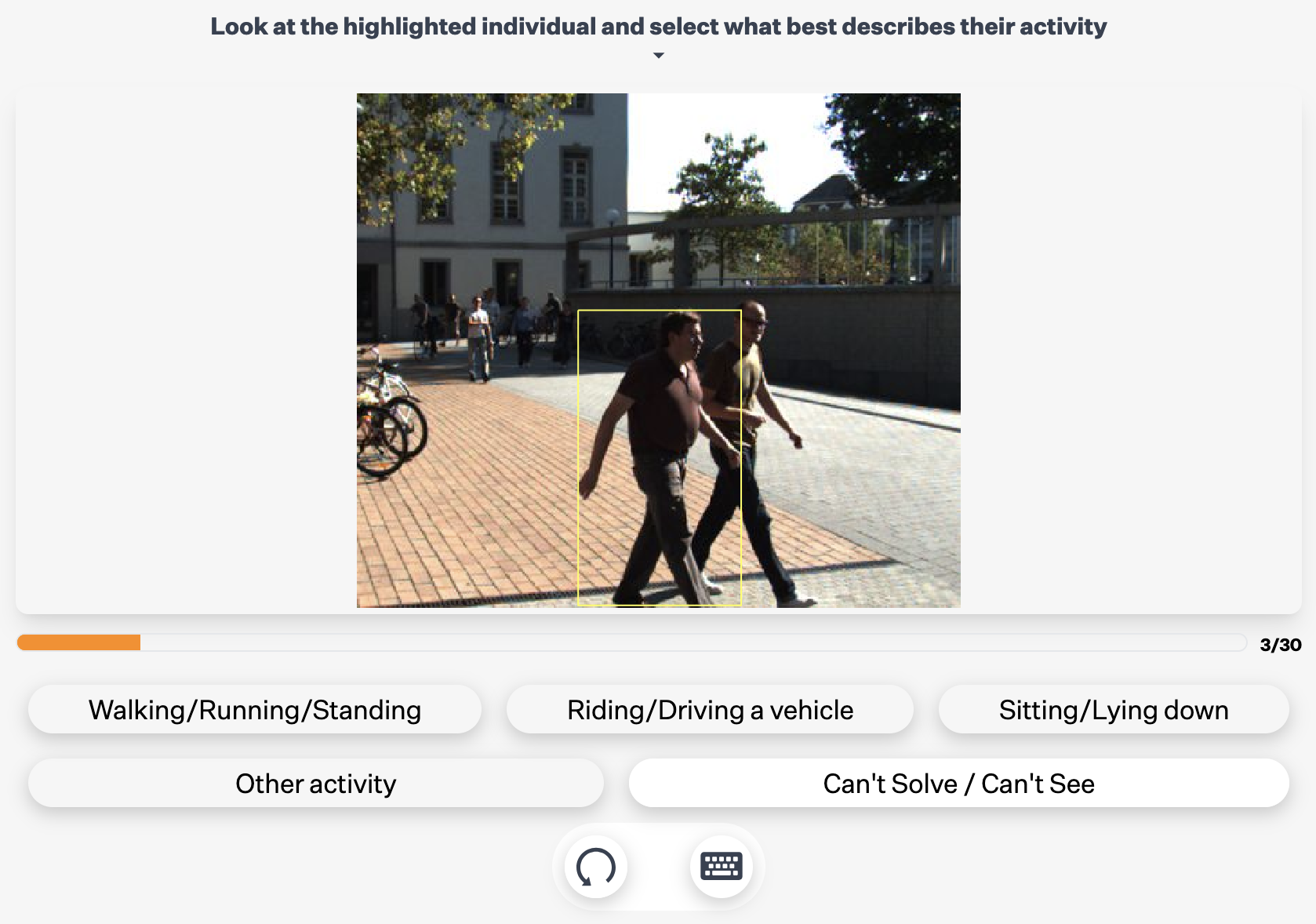}}
\caption{Annotator interface for microtask 6: \emph{Activity classification}.}
     \label{fig:example_nanotask_2}
\end{figure} 

\subsection{Validated Ground Truth Generation}
\label{subsec:experimental setup, validated gt generation}

To establish a reliable evaluation benchmark, we constructed a high-quality \textit{validated ground truth} (VGT) that achieves a very high recall by capturing every visible pedestrian.
Each bounding box is associated with a probability score that reflects the likelihood of containing a pedestrian, rather than relying on a single hard label that may be erroneous.

We use a two-stage microtask pipeline. In the first stage, we localize pedestrian instances. In the second stage, we validate the semantic content of each bounding box, generating soft labels that reflect annotator's uncertainty.

\paragraph{Stage 1: Bounding box collection.}
Our first aim is to identify all possible human instances in the data. To achieve this, we used two complementary task configurations, each with three different annotators.
\begin{itemize}[left=0pt]
    \item \textbf{Direct box annotation} (\textbf{Microtask 1}): All annotators draw tight bounding boxes around humans that have not yet been marked in the image.
    \item \textbf{Keypoint-to-box annotation} (\textbf{Microtasks 2 \& 3}): Each annotator first places a single keypoint on unmarked humans. Then, around each resulting group of keypoints, each annotator draws a bounding box. The drawn bounding boxes are then aggregated.
\end{itemize}

Running both strategies in separate task configurations allows us to compare annotation effort and resulting recall. For constructing the VGT, we combined both methods to ensure maximal coverage of pedestrian instances. We manually removed duplicate boxes resulting from the different strategies. Further details and visual representations of the tasks are provided in table~\ref{tab:box_strategy} and figure~\ref{fig:example_nanotask_2}.

\begin{table}[tb]
\caption{Semantic validation strategies. Column 'AR' indicates whether ambiguity refinement was applied. Confidence interval (CI) widths are based on the Wilson score interval and the delta method for 95\% confidence. Costs are reported as average per bounding box in seconds. $\ddagger$ Manual effort for duplicate removal is not included.}
\centering
\resizebox{\linewidth}{!}{  
\begin{tabular}{lcccc}
\toprule
 & \textbf{AR} & \textbf{\# Annotations} & \textbf{CI Width} & \textbf{Costs [s]}\\
\midrule
Is pedestrian? &  & $11.0 \pm 0.0$ &  $0.41 \pm 0.08$  & 37.87 \\
Is human \& stand/walk? & & $22.0 \pm 0.0$ &   $0.29 \pm 0.19$  &  57.03\\
Is human \& stand/walk?  & X & $33.3 \pm 11.1$  &   $0.26 \pm 0.16$  & 77.85\\
Validated GT & X &  $48.2 \pm 18.2$ &   $0.24 \pm 0.12$ & 124.75$\ddagger$ \\
\bottomrule
\end{tabular}
}
\label{tab:attribute_strategy}
\end{table}

\newpage
\paragraph{Stage 2: Semantic validation.}
In the second stage, we determine whether each bounding box contains a pedestrian. This process also serves as the foundation for label error correction, as described in~\ref{subsec:method, nano-task correction design}. For this purpose, we used the following three microtasks, each of which was answered by 11 different annotators:
\begin{itemize}[left=0pt]
    \item \textbf{Microtask 4}: Is the object a pedestrian? (\textit{Yes}, \textit{No}, \textit{Can't See/Can't Solve})
    \item \textbf{Microtask 5}: Is the object a real human? (\textit{Yes}, \textit{No}, \textit{Can't See/Can't Solve})    
    \item \textbf{Microtask 6}: What is the activity? (\textit{Walking/Running/Standing}, \textit{Riding/Driving a vehicle}, \textit{Sitting/Lying down}, \textit{Other activity}, \textit{Can't See/Can't Solve})
\end{itemize}

While microtask 4 can theoretically provide sufficient information, the combination of microtask 5 and 6 leads to increased annotator focus and reduced ambiguity. However, this approach requires twice the number of responses, as both tasks must be completed, resulting in 22 annotations per bounding box. To further improve soft label quality, we incorporated two additional refinement steps. First, for boxes with high disagreement in microtasks 5 and 6, we added 11 additional annotations per task. 
We refer to this step as \textit{ambiguity refinement} (AR). 
Second, we group duplicates of the same object from stage 1. These duplicate boxes were treated as describing the same object and aggregated accordingly. 
This grouping was applied exclusively to the VGT.
An overview is given in table~\ref{tab:attribute_strategy}.
The final soft label for determining whether a bounding box contains a pedestrian is computed by multiplying the soft label from microtask 5 (real human) with the soft label from microtask 6 (standing or walking).

\subsection{Label Error Detection}
\label{subsec:experimental setup, label error detection}

We use three label error detection methods, which are based on the predictions of YOLOX and Cascade R-CNN. These methods generate proposals for potential label errors.

\emph{MetaDetect.} We use the MetaDetect meta classification model, which applies gradient boosting. We use the default parameter values and make predictions via $5$-fold cross-validation on the respective hold-out set. The $\iou$ threshold, which determines whether predictions are considered to be TP or FP, is set to 0.5. 
For each predicted bounding box, the meta classification model predicts a probability for the prediction being correct, based on which we identify label errors.
Meta classification results are listed in table~\ref{tab:metadetect_results_1}. 
\begin{table}[!ht]
\centering
\caption{Performance of the MetaDetect meta classification model on hold-out data, reported as mean accuracy and \mbox{AUROC} with standard deviations over 5 folds.}
\resizebox{0.9\linewidth}{!}{  
\begin{tabular}{lcc}
\toprule
\textbf{Object Detector} & \textbf{Accuracy} & \textbf{AUROC} \\
\midrule
Cascade R-CNN & $94.01 \pm 0.42\,\% $ & $0.9107 \pm 0.0010 $ \\
YOLOX &  $95.78 \pm 0.90\,\% $ & $0.9892 \pm 0.0051 $ \\
\bottomrule
\end{tabular}}
\label{tab:metadetect_results_1}
\end{table}

\emph{Instance-wise Loss Method.} This method is based on object detection loss at instance level. It requires a two-stage object detector, which is why it is applicable only to Cascade R-CNN predictions. 
Based on images and the corresponding GT annotations, this method generates label error proposal boxes with a corresponding score and class prediction. We consider proposals with a score of at least 0.01 for the class \emph{pedestrian}. 

\emph{ObjectLab.} The detected bounding boxes can be evaluated with respect to the overlooked score with ObjectLab. Originally, only confident predictions with a probability greater than 0.95 were considered, but to infer a score for every detected bounding box, this threshold is removed. All other parameters are set to their default values. Additionally, we consider the poor location scores of the original GT annotations. As there is only a single class under consideration, the swapped class label error does not occur here. As label quality scores are normalized and lower values indicate poorer label quality, we subtract these scores from 1 and interpret the result as a probability for the corresponding label issue. The detected bounding boxes, as well as the original GT boxes, are then reviewed in order of probabilities.

%%%%%%%%%%%%%%%%%%%%%%%%%%%%%%%%%%%%%%%%%%%%%%%%%%%%%%%%%%%%%%%%%%%%%%%%%%%%%%%%%%%%%%%%%%%%%%%%%

\subsection{Evaluation Metrics}
\label{subsec:experimental setup, matching rules and evaluation metrics}

\emph{Label Errors / FN in Original GT:}
We match predicted bounding boxes with the original GT using an $\iou$ threshold of 0.5, and with the VGT using a threshold of 0.1. These thresholds serve different purposes: the $0.5$ threshold is standard for checking agreement with original annotations, while the $0.1$ threshold allows for approximate matching against VGT, which acts as a proxy for re-annotation.
A prediction is considered a label error (\ie, a false negative in the original GT) if it has no match in the original GT but overlaps with the VGT. We classify errors into two types: completely missing ($\iou = 0$) and misaligned boxes ($0 < \iou < 0.5$). Matching is performed greedily by descending $\iou$, and each VGT and GT annotation is matched at most once.

\emph{Introduced Label Errors:}
To assess the risk of new errors introduced by labeling strategies, we compare their outcomes against the VGT. If a semantic validation strategy disagrees with the VGT on a bounding box's label (\eg, one says \emph{pedestrian}, the other \emph{not pedestrian}), the box is misclassified. This may cause errors (FNs or FPs) to be missed or falsely introduced.

\emph{Costs:}
Each labeling strategy incurs costs for creating bounding boxes, as reported in table~\ref{tab:box_strategy}. Label error detection methods avoid these costs by using model predictions. However, all bounding boxes, whether generated by algorithms or created by humans, must be semantically validated. We compute the total costs using the per-box validation costs from table~\ref{tab:attribute_strategy}. A more detailed discussion on assigning costs per individual sample is provided in the appendix.

\begin{figure}[tb]
    \centering
    \scalebox{1.0}{\includegraphics[width=\linewidth]{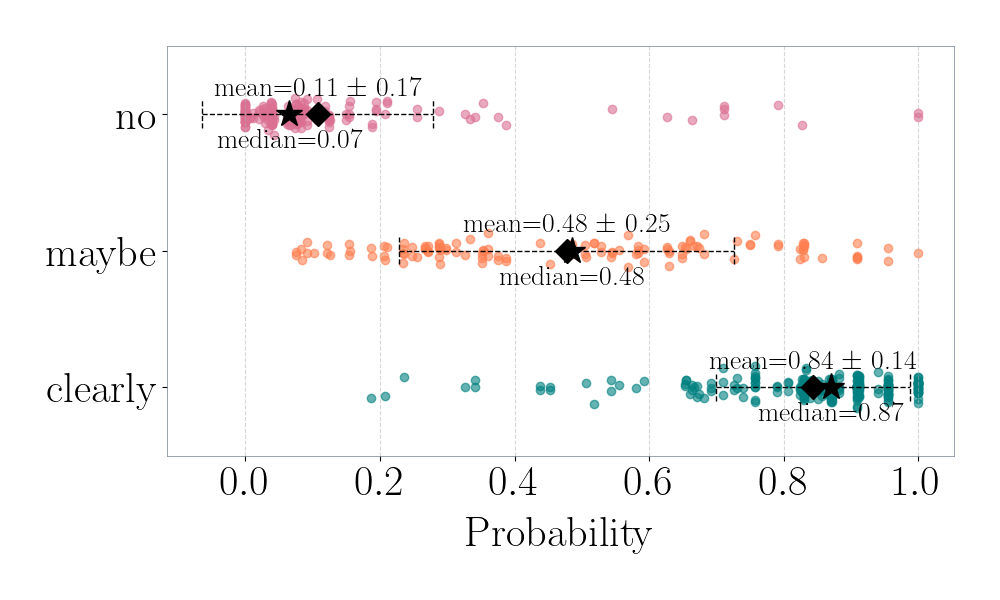}}
    \caption{Comparison of VGT soft label probability and human perception of being a pedestrian.
    Each dot represents an expert annotator annotation and its corresponding soft label probability.
    The diamond represents the mean, the dashed lines the standard deviation and the star the median.}
    \label{fig:probability_to_human_perception}
\end{figure}

\begin{figure}[tb]
  \centering
  \begin{subfigure}[t]{0.49\linewidth}
    \centering
    \includegraphics[width=\linewidth]{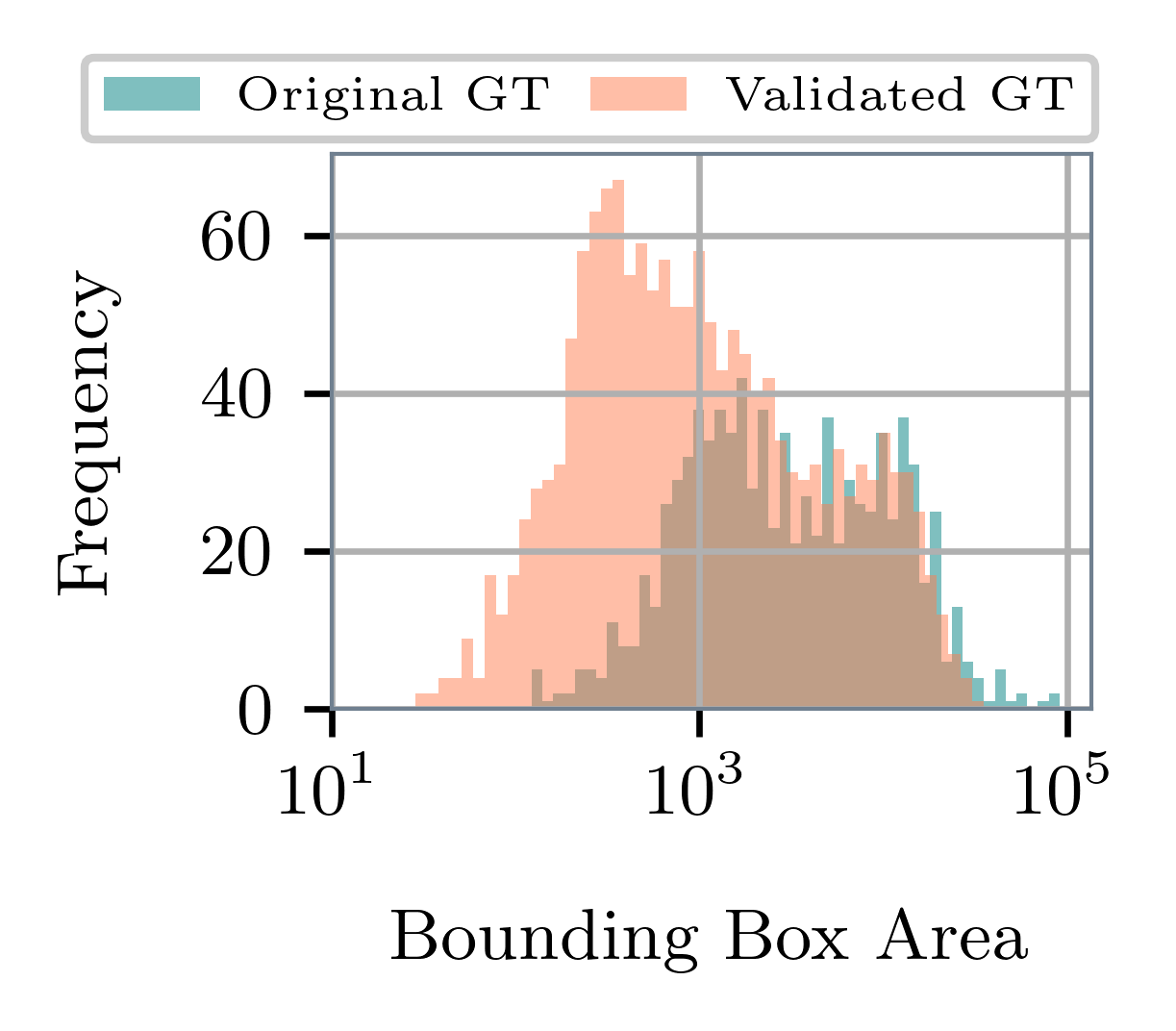}
    \caption{Distribution of bounding box area of original and validated annotations. Here, we consider VGT annotations with a soft label probability of $0.5$ or higher.}
    \label{fig:bbox_area_histogram}
  \end{subfigure}
  \hfill
  \begin{subfigure}[t]{0.49\linewidth}
    \centering
    \includegraphics[width=\linewidth]{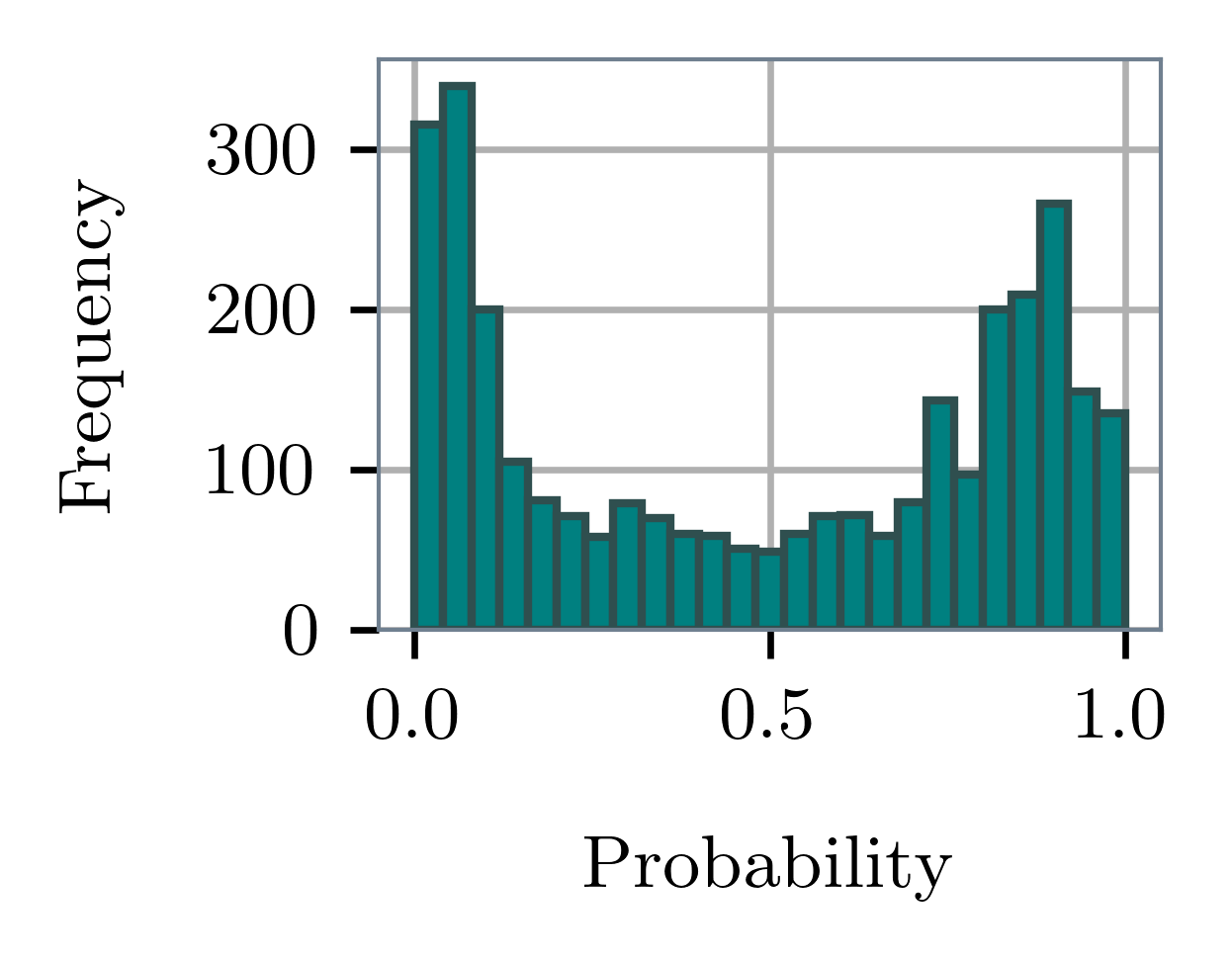}
    \caption{Distribution of VGT soft labels. Through aggregation of multiple responses from annotators a probability for a bounding box containing a pedestrian was inferred.}
    \label{fig:soft_label_distribution}
  \end{subfigure}
  \vspace{0.3cm}
  \caption{Details on VGT annotations including the distribution of bounding box area compared to original annotations and soft labels. We observe that VGT annotations additionally contain many small objects and the soft labels reveal that some ambiguous annotations exist while annotator agreement is usually high. }
  \label{fig:Hist_Bbox_Area_and_soft_label_distr}
\end{figure}

\section{NUMERICAL RESULTS}
\label{sec:numerical results}

We present quantitative results for comparing the original GT with VGT,  identified label errors, performance of label error detection methods with a focus on annotation costs as well as trade-offs for labeling strategies with regard to costs and quality of data.

\begin{figure*}[tb]
    \centering

    \begin{subfigure}[b]{0.41\textwidth}
        \includegraphics[width=\textwidth]{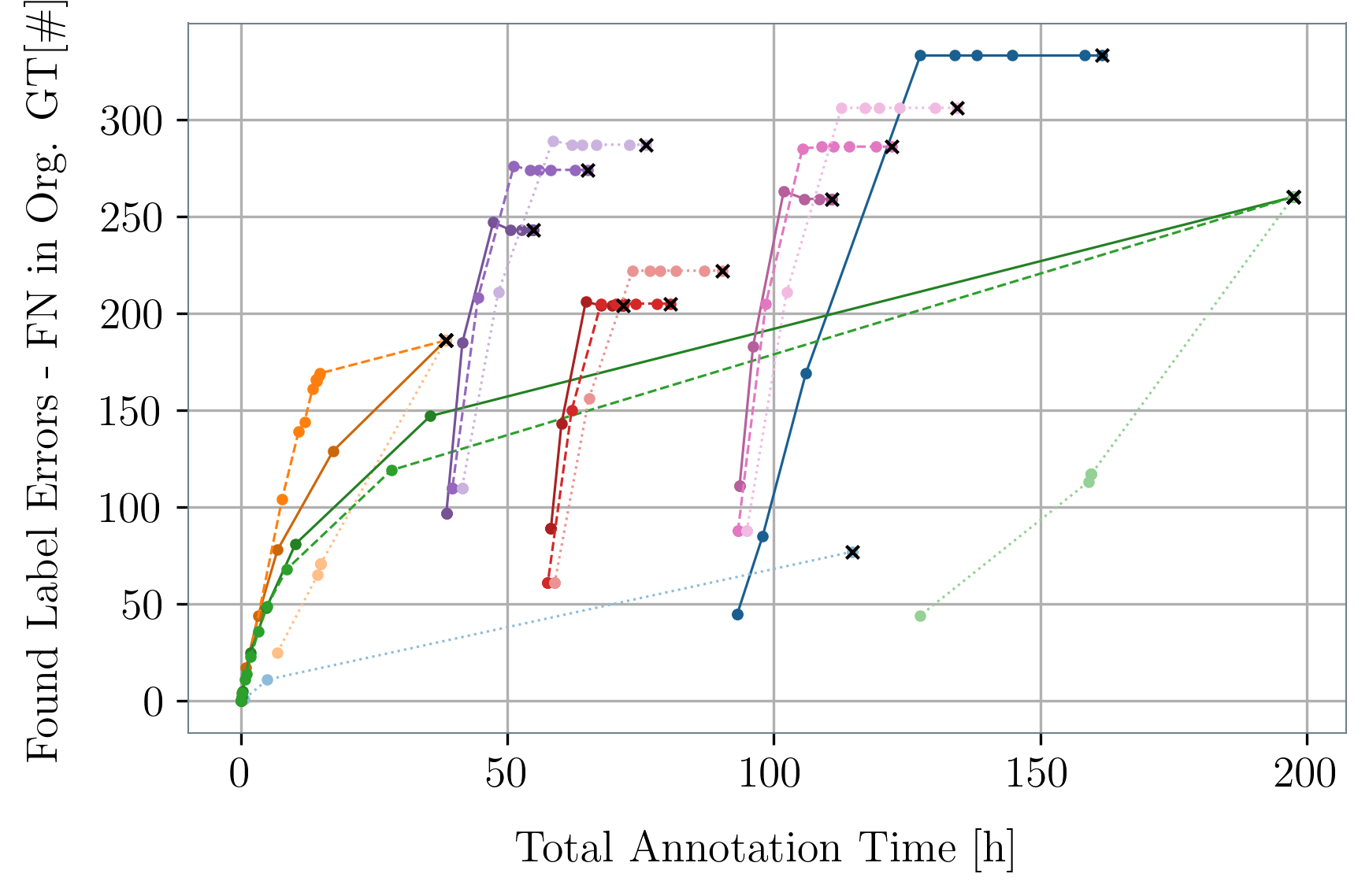}
        \caption{Found Label Errors (FN in original GT)}
        \label{fig:cost_error_comparison-fn}
    \end{subfigure}
    \hfill
    \begin{subfigure}[b]{0.41\textwidth}
        \includegraphics[width=\textwidth]{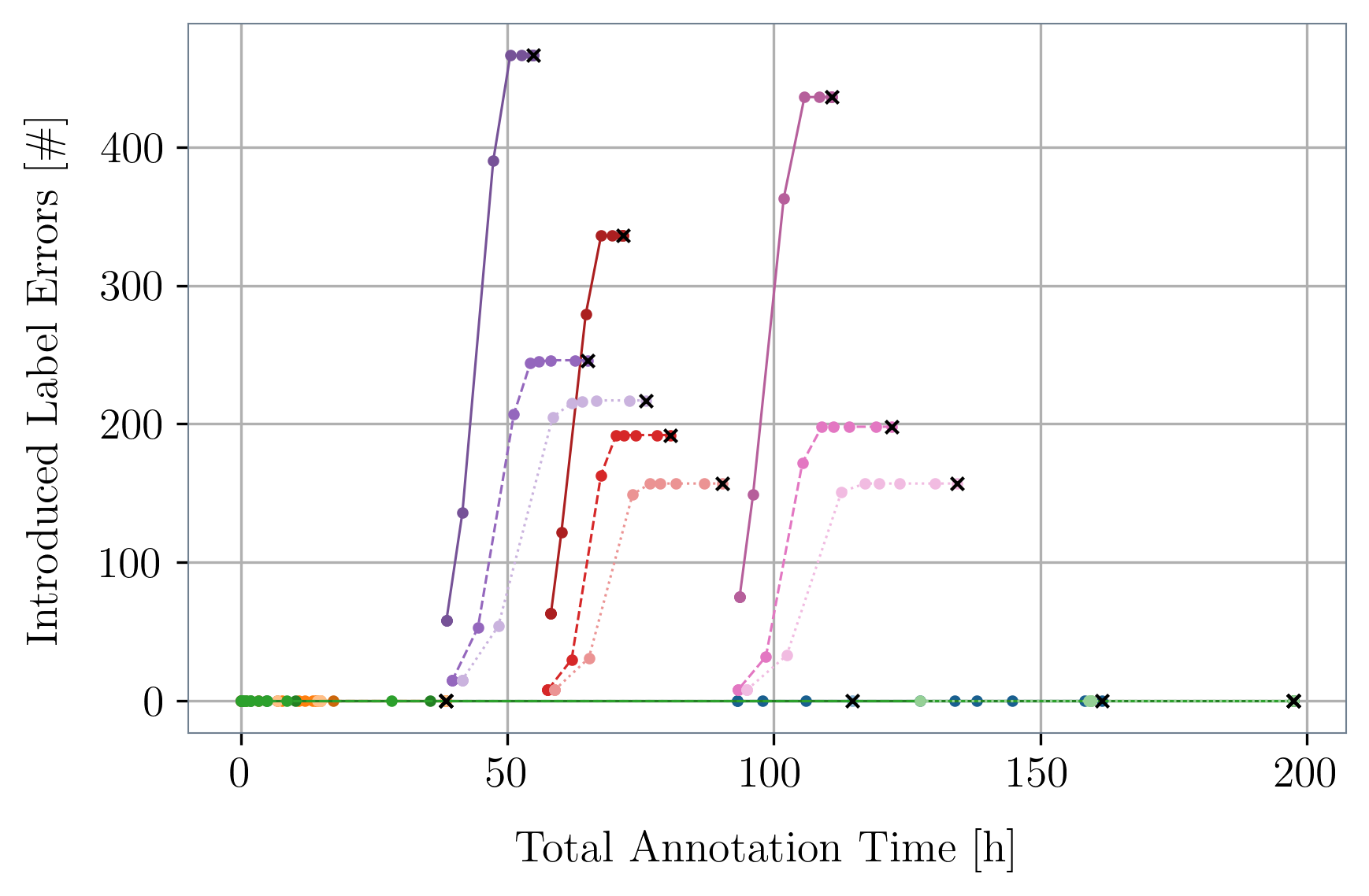}
        \caption{Introduced Label Errors due to low quality labeling strategy}
        \label{fig:cost_error_comparison-introduced}
    \end{subfigure}
    \hfill
    \begin{subfigure}[b]{0.15\textwidth}
        \includegraphics[width=\textwidth]{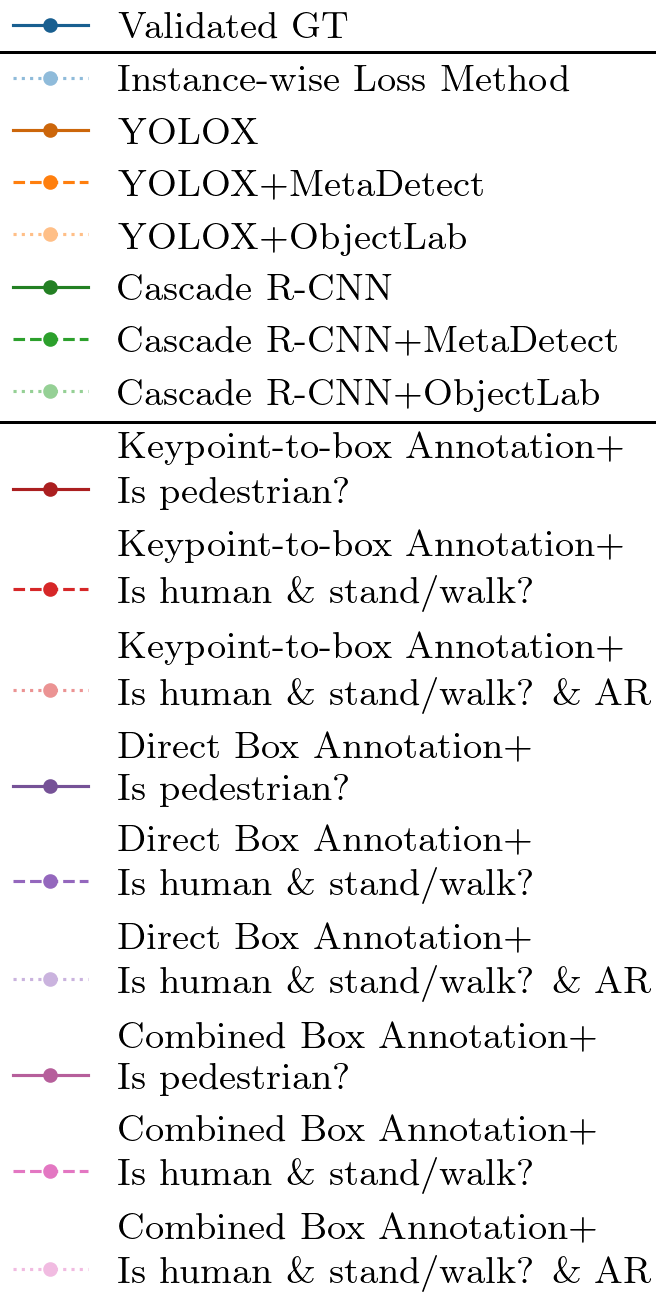}
        \caption{Legend}
        \label{fig:cost_error_comparison-legend}
    \end{subfigure}
    \vspace{0.3cm}
    \caption{
    Comparison of label error detection capabilities across various methods and labeling strategies, along with their respective costs. 
In both plots, the x-axis represents the total annotation time, while the y-axis shows the metric of interest.
Each curve corresponds to a specific method, with dots marking different model confidence thresholds. The point labeled 'x' indicates the lowest threshold, corresponding to the most predictions above the threshold. As the confidence threshold increases, fewer predictions are included.
Methods are grouped into three categories in the legend (c): (1) GT baselines (original and validated), (2) automated label error detection methods, and (3) labeling strategies combining different box annotation and semantic validation techniques.
The left plot (a) shows the number of detected label errors, specifically, false negatives (FNs) in the original GT. 
The center plot (b) illustrates errors introduced by different labeling strategies, demonstrating that lower-cost strategies may lead to new annotation errors.
Note that automated error detection methods assume access to validated GT for evaluation, but also use its high cost.
For a detailed description of the evaluation metrics, refer to \ref{subsec:experimental setup, matching rules and evaluation metrics}. 
All boxes are pre-filtered to a minimum height of 25 pixels, as per KITTI labeling guidelines. A validated GT label is considered a pedestrian if its soft label probability exceeds 0.8.
For more relaxed filtering criteria, see the appendix. 
 }
    \label{fig:cost_error_comparison}
\end{figure*}

\subsection{Comparison of Original and Validated Annotations}
\label{subsec:numerical results, comparison of original and validated annotations}

Before comparing the VGT to the original GT, we first verify its quality. We evaluate three aspects: whether pedestrians were missed, whether duplicate bounding boxes exist, and, how well the soft labels align with human perception.
In a random sample of 200 images, we found that about 0.5\% of pedestrians might have been missed, and about 0.8\% of pedestrian bounding boxes were duplicates. See the appendix for the full analysis.
To evaluate the alignment of soft labels with human perception, we randomly sampled 200 bounding boxes from the VGT and let three expert annotators rate them. Each annotator could choose between the labels 'clearly', 'maybe', or 'no' to indicate whether the bounding box contains a pedestrian. In figure~\ref{fig:probability_to_human_perception}, these labels are compared with the corresponding soft label probability.
The comparison shows that the categories match well with the expected scores. The category 'maybe' has a mean and median of 0.48, which is almost perfectly calibrated, as we would expect a value around 0.5. For the category 'clearly', we would expect a value around 0.8, which is consistent with the results.
We conclude that the validated ground truth contains almost no missing pedestrians (about 0.5\%), almost no duplicate bounding boxes (about 0.8\%), and is well aligned with human perception.

We compare the validated annotations with the original annotations to determine the number of overlooked pedestrians and the number of misfitting or inaccurate bounding boxes. Overlooked pedestrians are VGT annotations that have an $\iou$ of zero with any original annotation. Misfitting boxes are VGT annotations for which there is an intersecting bounding box in the original GT, but with an $\iou$ below 0.5.
We distinguish between large and small objects based on the minimum bounding box height, using thresholds of 25 or 40 pixels, which relate to the difficulty levels defined in the KITTI benchmark. Additionally, we consider two different probability thresholds, $0.5$ and $0.8$, for the soft labels in the VGT. The distribution of soft label probabilities is shown in figure~\ref{fig:soft_label_distribution} and illustrates that while most annotations show high agreement among annotators, some ambiguous cases occur.
It is important to note that, unlike the original annotation protocol, the procedure described in~\ref{subsec:experimental setup, validated gt generation} includes all pedestrians, regardless of their size or whether they fall into defined \emph{don't care} regions. As a result, we observe that the VGT also contains small objects that were not annotated in the original data, as shown in figure~\ref{fig:bbox_area_histogram}.

To ensure a fair comparison, we define a default configuration for the annotations under consideration. We include all objects that have an intersection over bounding box area below 0.5 with any \emph{don't care} region and a bounding box height of at least 25 pixels. For the VGT annotations, we apply a soft label probability threshold of 0.5 to decide whether to include them. The resulting VGT contains 1,567 annotations for the class \emph{pedestrian}, which is 671 more than in the original GT.
Using this configuration, we identified a total of 384 label errors in the subset of 1,497 KITTI training images (table~\ref{tab:number_of_label_errors}). Out of these, 228 pedestrians were missing entirely and 156 were annotated with inaccurate bounding boxes.
We estimate that the original GT misses between 42 and 683 pedestrians, depending on the selected conditions such as minimum object size and probability threshold.
This corresponds to between 4\,–\,76\% relative to the 896 original pedestrian annotations.
If we treat the overlooked and misaligned annotations as false negatives (FNs) relative to the original GT, we identify 206 FNs under strict conditions (applying a high confidence threshold, excluding \emph{don't care} regions, and enforcing size constraints) or 862 FNs when these restrictions are relaxed.
This corresponds to a false negative rate \mbox{(FNs / [FNs + Original GT])} ranging from 18\% to 49\%, depending on the evaluation setting.

\begin{table}[tb]
\centering
\caption{Number of identified label errors for the class \emph{pedestrian} in the KITTI dataset depending on the probability threshold for soft label annotations and the minimal height of considered objects. Through comparison of original and validated annotations, we identify a number of label errors in the original dataset even when considering only evident errors.
For the definition of label errors see~\ref{subsec:experimental setup, matching rules and evaluation metrics}.}
\label{tab:number_of_label_errors}
\resizebox{\linewidth}{!}{
\begin{tabular}{lcc|cc}  
\toprule
Probability threshold & \multicolumn{2}{c|}{$p \geq 0.5$} & \multicolumn{2}{c}{$p \geq 0.8$} \\
\midrule
\hspace{1em} Object Height & Overlooked & Misfitting & Overlooked & Misfitting \\
\midrule
& \multicolumn{4}{c}{\hspace{1em}\emph{Considering all objects}} \\
\addlinespace
\hspace{1em}Arbitrary         & 683 & 179 & 272 & 115 \\
\hspace{1em}$\geq 25$ pixels  & 454 & 162 & 218 & 112 \\
\hspace{1em}$\geq 40$ pixels  & 157 & 118 &  83 & 89 \\
\midrule
& \multicolumn{4}{c}{\hspace{1em}\emph{Considering only objects outside of 'Don't Care' regions}} \\
\addlinespace
\hspace{1em}Arbitrary         & 341 & 173 & 121 & 112 \\
\hspace{1em}$\geq 25$ pixels  & 228 & 156 & 97 & 109 \\
\hspace{1em}$\geq 40$ pixels  & 91 & 114 &  42 & 87 \\
\bottomrule
\end{tabular}
}
\end{table}
\subsection{Label Error Detection and Correction}
\label{subsec:numerical results, performance on real label errors}

In figure~\ref{fig:cost_error_comparison-fn}, we compare the number of FNs in the original GT detected by various label error detection methods and labeling strategies. As outlined in~\ref{sec:method}, we define label errors as FNs in the original GT.
We derive two key insights from the figure: (1) error detection methods make labeling more cost-efficient, and (2) there remains substantial room for improvement.

Among the evaluated methods, only YOLOX+MetaDetect consistently outperforms its base model. Cascade R-CNN+MetaDetect underperforms compared to Cascade R-CNN alone. 
ObjectLab performs significantly worse, likely due to our adaptation of its method; the original version uses a strict confidence threshold of 0.95, which we relaxed for this analysis.
At such high thresholds, models provide highly confident predictions, but they cover fewer objects, limiting error discovery. For instance, using only predictions above 0.95 yields very few detected FNs, even in strong models.
This highlights that restricting to high-confidence predictions alone is too limiting, even when using optimal detection techniques.
Most detection-based approaches including YOLOX, Cascade R-CNN, and their error detection variants are able to find 150–200 FNs at a cost similar to or lower than the cheapest human annotation strategy. 
However, the base models saturate at 186 (YOLOX) and 260 (Cascade R-CNN) FNs, while the VGT contains 330 FNs. This indicates that up to 44\% of missing labels remain undetected. The gap increases under less restrictive matching conditions (\eg, without minimum box size), where up to 66\% of FNs go unnoticed.
This limitation stems primarily from the object detectors themselves, which constrain the set of proposed boxes.
Labeling strategies show similar trends. As summarized in table~\ref{tab:box_strategy}, strategies using keypoint-to-box annotation generate fewer boxes and thus recover fewer FNs. Additional differences can be attributed to variations in label quality across correction strategies (see~\ref{subsec:numerical results, ablation studies} for ablations).
While detectors limit recall, our \rechecked~correction pipeline allows for more precise validation of the boxes they do propose and thus, improves soft labeling quality.
In summary, \rechecked~identifies hundreds of label errors in less time than it takes humans to even annotate the bounding boxes. Yet, our analysis also reveals that up to 66\% of potentially recoverable FNs remain undetected, highlighting significant research opportunities for improving label error detection.

\subsection{Trade-offs Between Labeling Strategies and Label Quality}
\label{subsec:numerical results, ablation studies}

We evaluate the benefits and limitations of various labeling strategies in figure~\ref{fig:cost_error_comparison-introduced}.
As expected, strategies cluster based on their bounding box annotation method discussed in \ref{subsec:experimental setup, validated gt generation}.
This is intuitive, since the number and quality of available boxes heavily influence the ability to identify FNs. Strategies using higher-quality soft labels consistently find more FNs.
However, in less restrictive settings, \eg, relaxing the minimum box size, even low-cost strategies can sometimes perform best.
It can be observed that the number of newly introduced errors decreases with increasing label quality. The lowest-cost semantic strategy ('Is pedestrian?') introduces several hundred errors. In contrast, higher-quality soft labels introduce significantly fewer mistakes.
This aligns with theoretical expectations. More precise soft labels tend to have narrower confidence intervals, which reduces the probability that a sample will cross the decision threshold due to random annotation variation. As a result, label corrections are more stable, and fewer errors are introduced.
We find that low-quality labeling strategies introduce more errors than they remove.

\section{Conclusion}
\label{sec:conclusion}

We introduced \rechecked, a framework for semi-automated detection and correction of label errors in object detection datasets.
For evaluation purposes, we acquired high-quality annotations based on which we identified at least 18\% missing and inaccurate pedestrian annotations in the original KITTI dataset. 

This work demonstrated the applicability of our framework to detect and correct label errors. However, we note that up to 66\% of the existing label errors remain undetected. This highlights research opportunities for improving label error detection methods. Importantly, \rechecked~is detector-agnostic by design and thus, advances in detection architectures can be leveraged to enhance label error detection.
We analyze different annotation procedures with respect to label quality and annotation cost, and show that attempting to correct datasets with poor-quality annotations may introduce more label errors than it fixes.

Future work may extend \rechecked~to additional datasets and classes to assess its transferability, which lies beyond the scope of this work.
We announce the public release of the validated KITTI annotations for the \emph{pedestrian} class. 
Based on these annotations, we provide a benchmark evaluation platform for label error detection methods for object detection datasets.  

\section*{Acknowledgements}
S.P.\ and M.R.\ thank Marius Schubert for discussion and help with baseline implementations.
S.P.\ and M.R.\ acknowledge support by the German Federal Ministry of Research, Technology and Space (BMFTR) within the junior research group project ``UnrEAL'' (grant no.\ 16IS22069).
J.K.\ and M.R.\ acknowledge support by the BMFTR within the project ``RELiABEL'' (grant no.\ 16IS24019B).
R.C., D.K.\ and L.S.\ also acknowledge support within the project ``RELiABEL'' (grant no. 16IS24019A).

%\bibliographystyle{IEEEtran}
%\bibliography{literature}

\twocolumn[
\begin{@twocolumnfalse}
  \begin{center}
    {\Large \bfseries Supplementary Material \par}
    \vspace{1ex}
  \end{center}
\end{@twocolumnfalse}
]

\renewcommand{\thefigure}{A.\arabic{figure}} 
\renewcommand{\thetable}{A.\arabic{table}}   
\setcounter{figure}{0}                        
\setcounter{table}{0}                         

\appendix

\subsection{Supplementary Material Overview}
\label{sec:Appendix}

\emph{More Details about Experimental Setup (B).} 
Describes annotation procedures, validated GT construction, and cost breakdown used to support the main experiments.
\begin{itemize}
    \item Annotators used microtask interfaces for bounding box drawing and activity selection, illustrated with UI screenshots (B.1).
    \item Validated GT was created by merging and manually deduplicating two annotation strategies, with recommendations for future designs (B.2).
    \item Annotation costs were derived from stable task durations using crowd workers paid hourly, ensuring fair and reproducible cost estimation (B.3).
\end{itemize}

\emph{Further Numerical Results (C).} Presents complementary quantitative evaluations to validate the VGT and detector performance.
\begin{itemize}
    \item Expert review of 200 images confirmed the VGT has low miss and duplicate rates, indicating high precision and recall (C.1).
    \item Bounding box size distribution analysis shows detectors miss smaller objects compared to the VGT (C.2).
    \item False positives are analyzed through qualitative examples, revealing typical error patterns in detection outputs (C.3).
    \item Less restrictive evaluation confirms findings across broader subsets (C.4).
\end{itemize}

\emph{Visual Examples (D).} 
Provides qualitative insights into labeling ambiguity and the effectiveness of the corrected labels.
\begin{itemize}
    \item Ambiguous cases with low agreement motivate using soft labels to capture uncertainty and improve model training (D.1).
    \item Visual comparisons between original and validated GT highlight missed, misaligned, and ambiguous annotations (D.2 - D.6).
\end{itemize}

\begin{figure}[!ht]
     \centering
     \includegraphics[width=\linewidth]{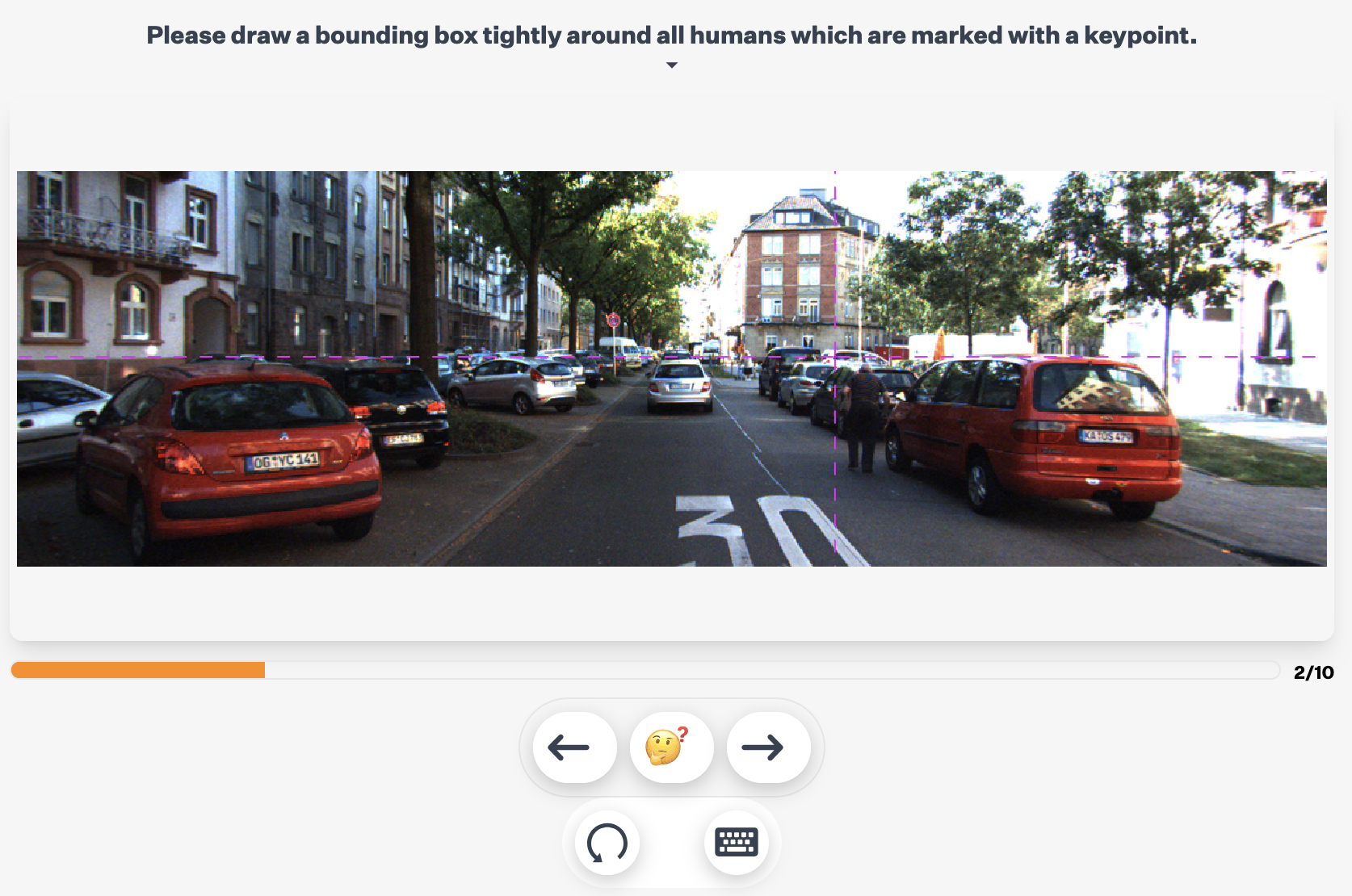}
\caption{Annotator interface for microtask 1: Direct box annotation, where multiple annotators draw tight bounding boxes around all real humans not yet marked in the image.}
     \label{fig:example_loc_nanotask}
 \end{figure}

  \begin{figure}[!ht]
     \centering
     \includegraphics[width=\linewidth]{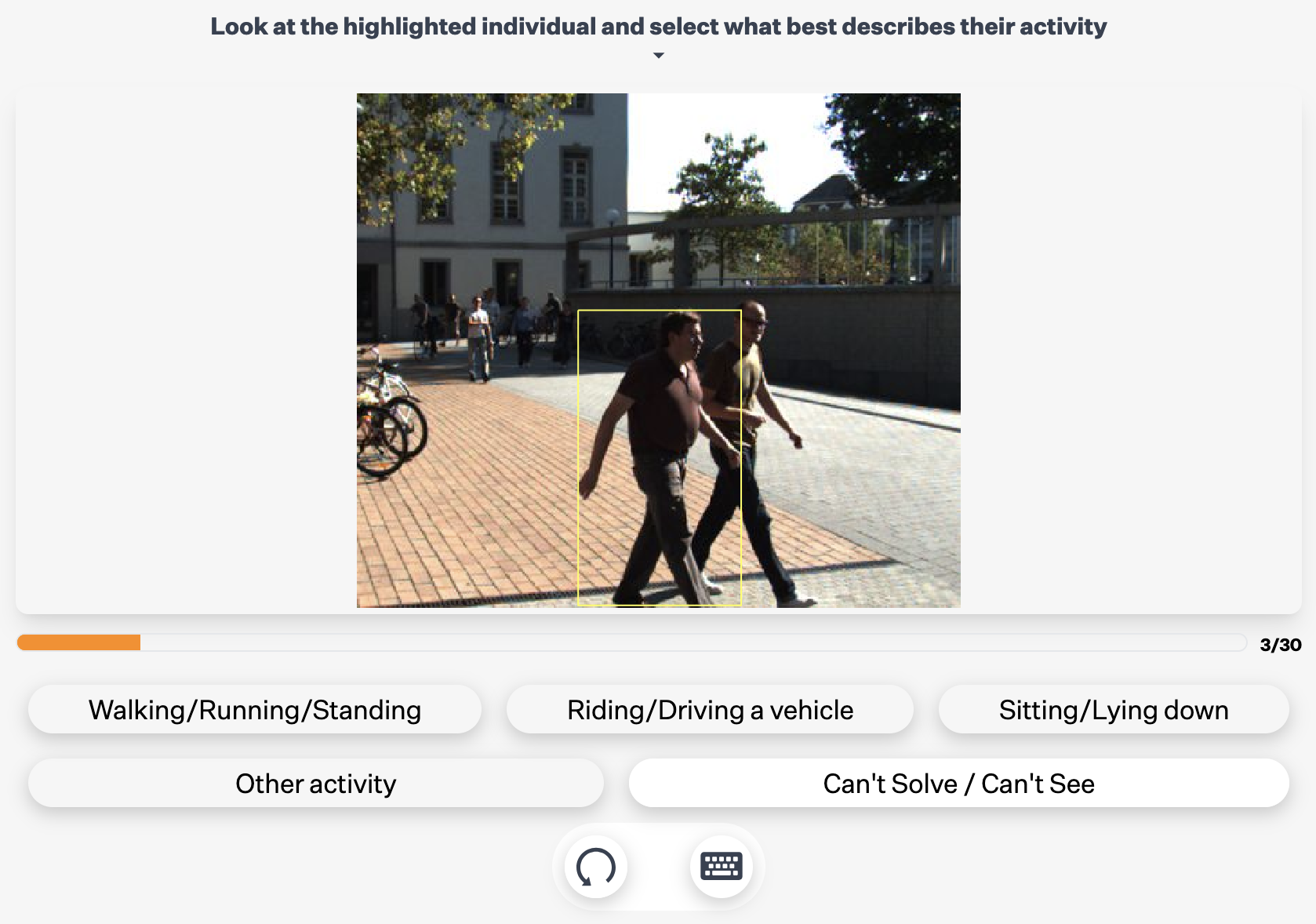}
\caption{Annotator interface for microtask 6: Activity classification, where annotators select the activity of a person from predefined options: \textit{Walking/Running/Standing}, \textit{Riding/Driving a vehicle}, \textit{Sitting/Lying down}, or \textit{Other}.}
     \label{fig:example_nanotask_3}
 \end{figure}  
 
\subsection{More Details about Experimental Setup}

\subsubsection{Validated GT}

In this section, we detail the process for generating the validated ground truth (VGT), with a focus on duplicate removal during the merging of two bounding box annotation strategies.

To identify duplicates, we considered all bounding box pairs across the two strategies with an $\iou$ greater than 0.25. Each candidate pair was manually reviewed, and boxes were marked as either duplicate or non-duplicate.
The combined set of bounding boxes was created by starting with the annotations from the \emph{Keypoint-to-Box} strategy and adding non-duplicate boxes from the \emph{Direct Box Annotation} strategy. We chose Keypoint-to-Box as the base because these annotations were aggregated across multiple annotators and generally more precise. The keypoint-first workflow also helped annotators achieve consensus on all visible human figures before defining bounding boxes.
In contrast, the Direct Box strategy resulted in frequent duplicates, often because annotators missed existing boxes or interpreted bounding boundaries differently.

Overall, we found that duplicate removal is difficult and time-consuming. In future annotation efforts, we recommend avoiding the merging of multiple annotation strategies and instead relying solely on the Keypoint-to-Box approach, with two improvements:
\begin{enumerate}
    \item Increase the number of annotators in the keypoint stage and verify coverage \eg by checking the false negative rate before proceeding to box drawing.
    \item Use multiple annotators to draw bounding boxes for each identified keypoint group and add a final check to ensure no boxes were missed.
\end{enumerate}
Although box aggregation generally worked well, some boxes were mistakenly merged or missing due to overlooked keypoints.
So we would recommend an additional round of bounding box annotation without aggregation to catch errors.
Finally, we recommend including clear annotation guidelines regarding occlusion. Some annotators estimated full-body extents (including occluded regions), while others labeled only visible parts. Both are valid, but the choice should be aligned with the intended downstream task.
Overall, both  strategies proved highly effective for identifying all human instances, including pedestrians.

\subsubsection{Annotation Interface}

In figure~\ref{fig:example_loc_nanotask} you can see an example of the used Annotation interface.

 \begin{figure}[!ht]
     \centering
     \includegraphics[width=\linewidth]{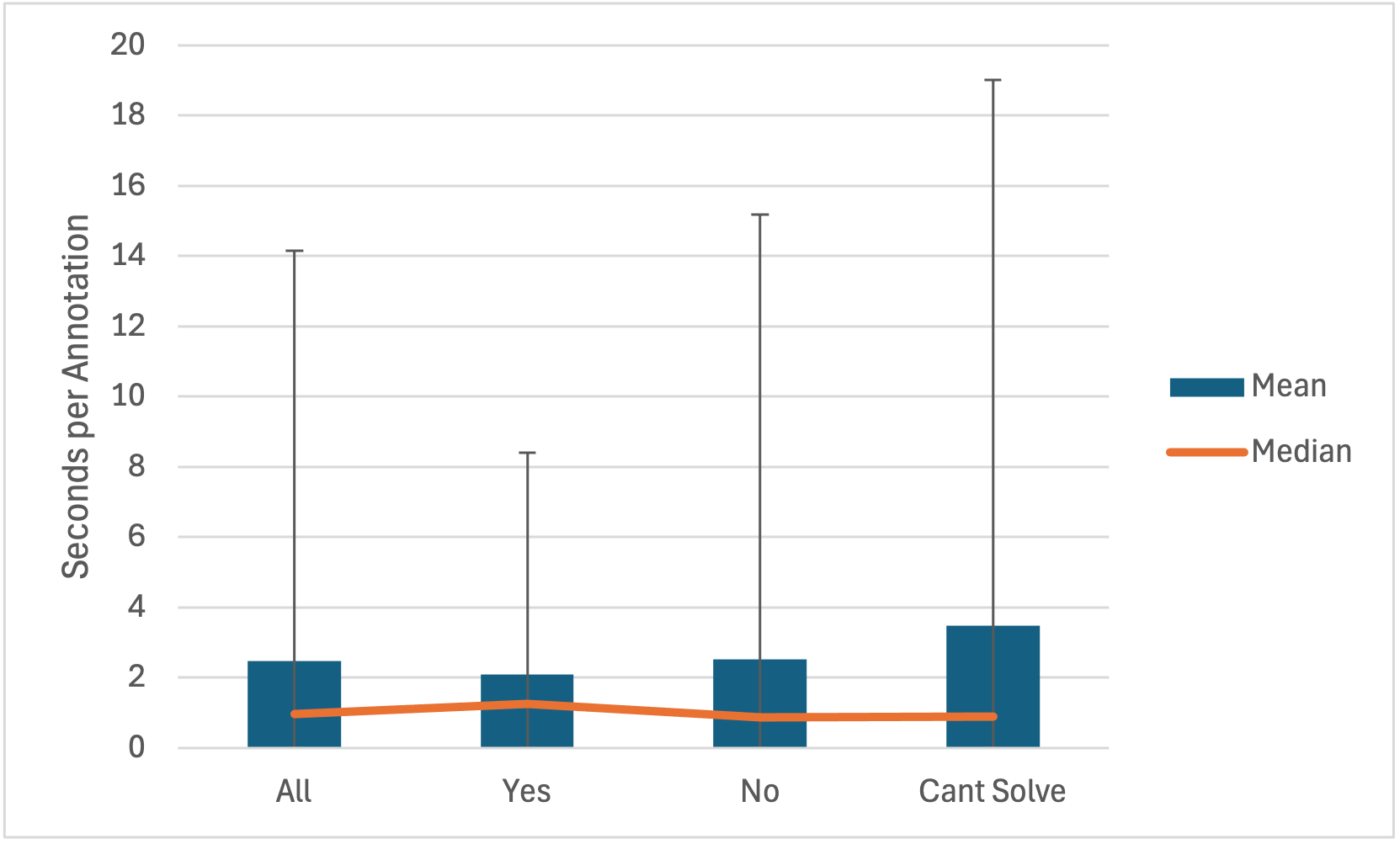}
\caption{Mean, median, and variance of annotation times for microtask 4 (“Is the object in the bounding box a pedestrian?”)}
     \label{fig:variance_annotation_times}
 \end{figure}

\subsubsection{Costs}

We used a crowdsourced labeling provider operating in Ukraine, with a compensation rate of 4.5\,€ per hour per annotator. Annotation costs are calculated per bounding box correction, based on the measured annotation time for each individual label.

A detailed cost analysis is provided in the main paper, where we report the total annotation cost per corrected bounding box. A further breakdown by answer option is not necessary, as the annotation times and thus the costs do not vary significantly across answer types. This is shown in \autoref{fig:variance_annotation_times} for microtask~4 (“Is the object in the bounding box a pedestrian?”), where the average annotation times for the options \emph{Yes}, \emph{No}, and \emph{Can't Solve} show only minor differences. This consistency justifies the use of a unified cost estimate across all response types.

\begin{figure}[!ht]
    \centering
    \begin{subfigure}[t]{0.49\linewidth}
        \includegraphics[width=\linewidth,height=4.8cm,keepaspectratio]{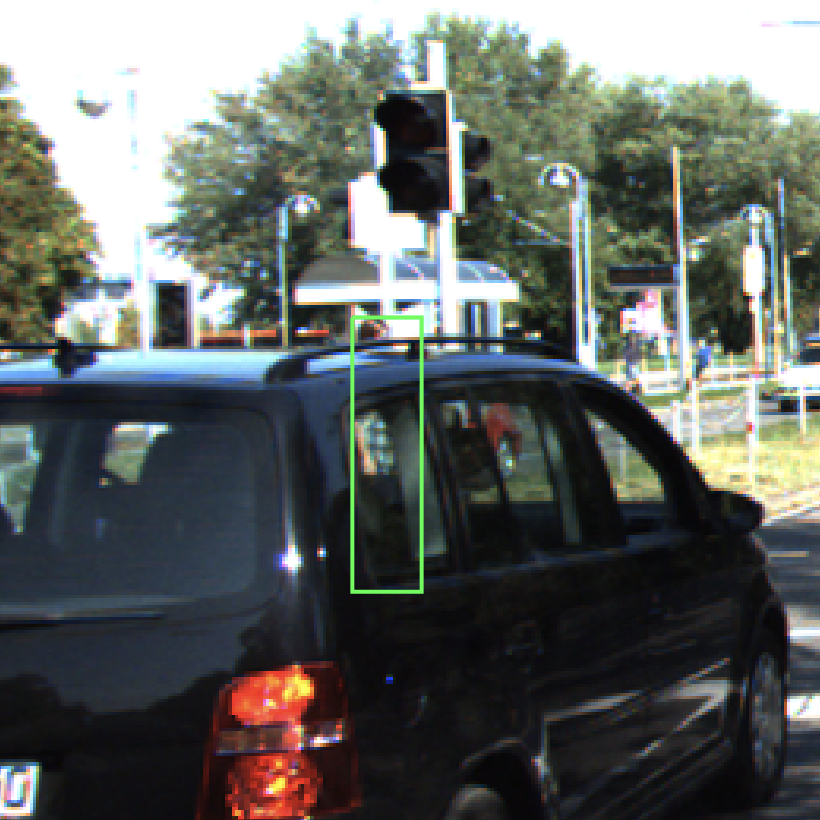}
        \caption{}
        \label{fig:missed_person_1}
    \end{subfigure}
    \hfill
    \begin{subfigure}[t]{0.49\linewidth}
        \includegraphics[width=\linewidth,height=4.8cm,keepaspectratio]{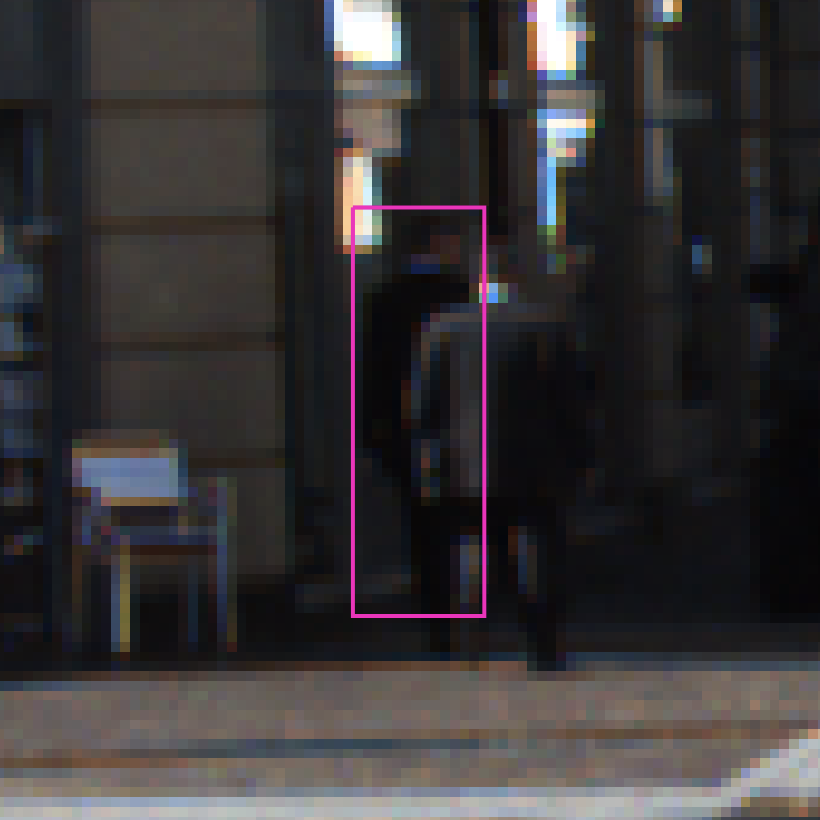}
        \caption{}
        \label{fig:missed_person_2}
    \end{subfigure}
    \caption{Two missing humans in the validated GT. Green box pedestrian in original data, pink bounding box found humans in Validated GT. On the left side, the human is almost completely hidden by the car and thus was most likely missed by the annotators. Since the human was detected in the original GT, we could potentially add this box by merging our boxes with the original GT. On the right side, it looks like a second person is right of the found person. In both cases the persons are extremely difficult to see. }
    \label{fig:missed_person_examples}
\end{figure}
\begin{figure}[!ht]
    \centering
    \begin{subfigure}{0.39\linewidth}
        \includegraphics[width=\linewidth]{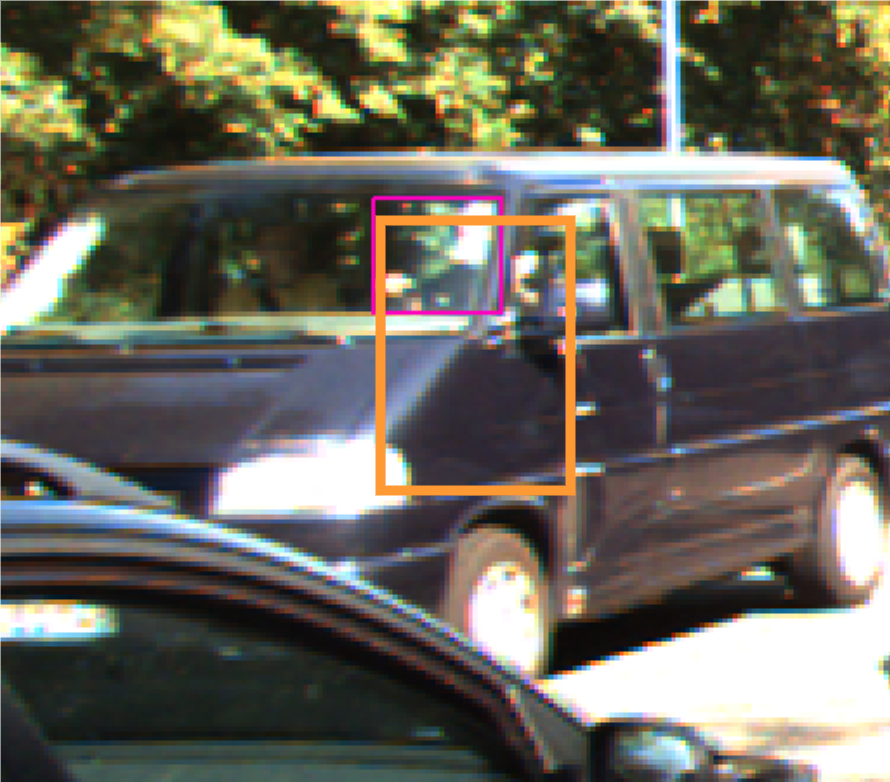}
        \caption{}
        \label{fig:duplicates-0}
    \end{subfigure}
    \hfill
    \begin{subfigure}{0.58\linewidth}
        \includegraphics[width=\linewidth]{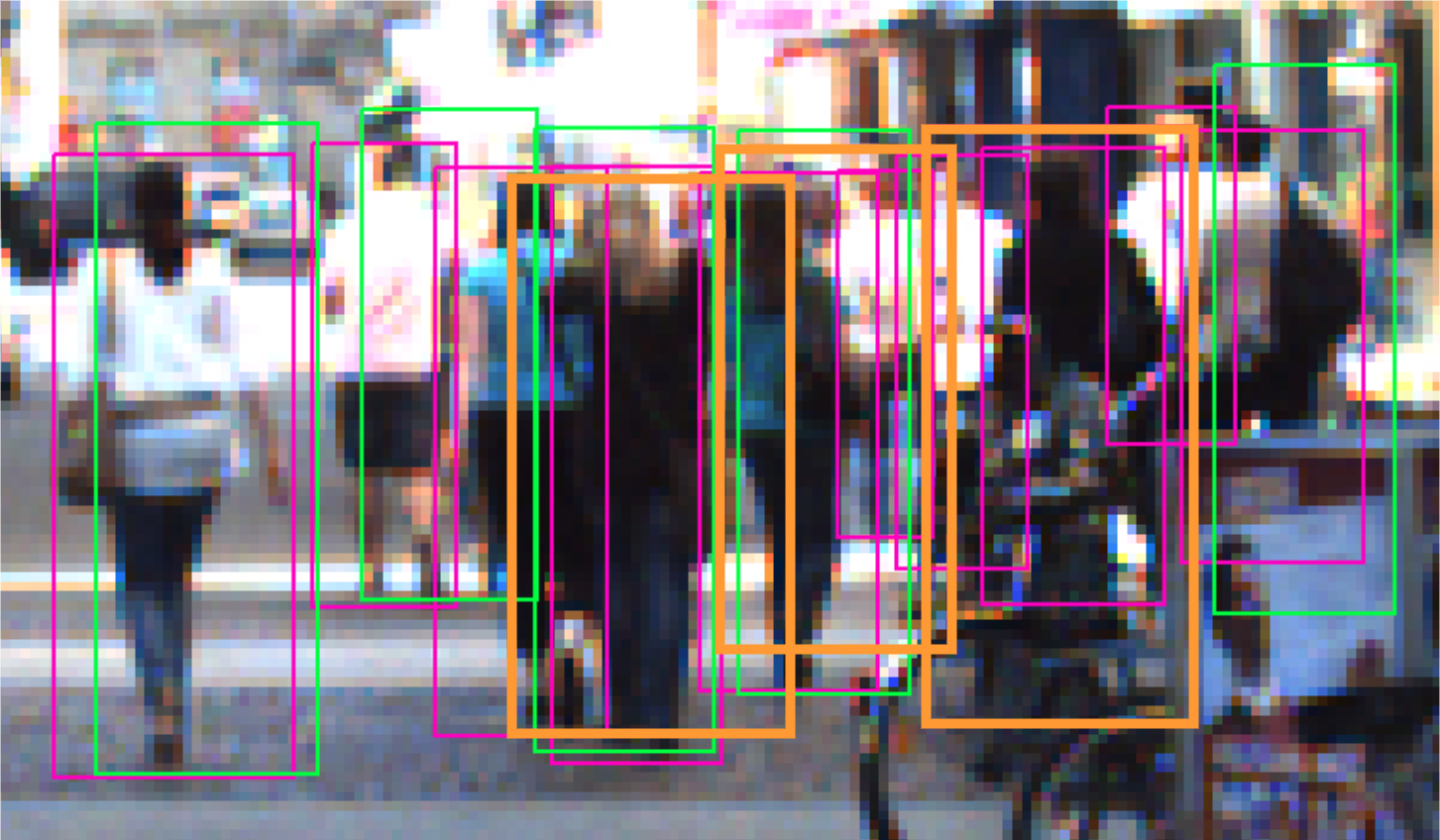}
        \caption{}
        \label{fig:duplicates-1}
    \end{subfigure}
    \caption{Potential duplicate bounding boxes in the validated GT. 
    Duplicates can occur for largely different bounding boxes \eg for humans in cars, where one annotator annotates only the visible part of the human and another annotates the estimated shape of the human see \autoref{fig:duplicates-0}.
    These duplicates are not an issue for the remainder of the analysis because the probability of these bounding boxes being a pedestrian are $<0.5$ and thus excluded from the evaluations anyways.
    For one other image out of the 200 randomly, we found potentially duplicate pedestrian entries (see \autoref{fig:duplicates-1}).
    This scene shows a group of people in the background. Due to the resolution, it is quite difficult to see where one person ends and another begins.
    Based on a rating of an expert annotator, we assume that three bounding boxes (marked in orange) are actually duplicates here mainly due to imprecise bounding box drawing. }
    \label{fig:duplicates_vgt}
\end{figure}

\begin{figure}[!ht]
    \centering
    \includegraphics[width=\linewidth]{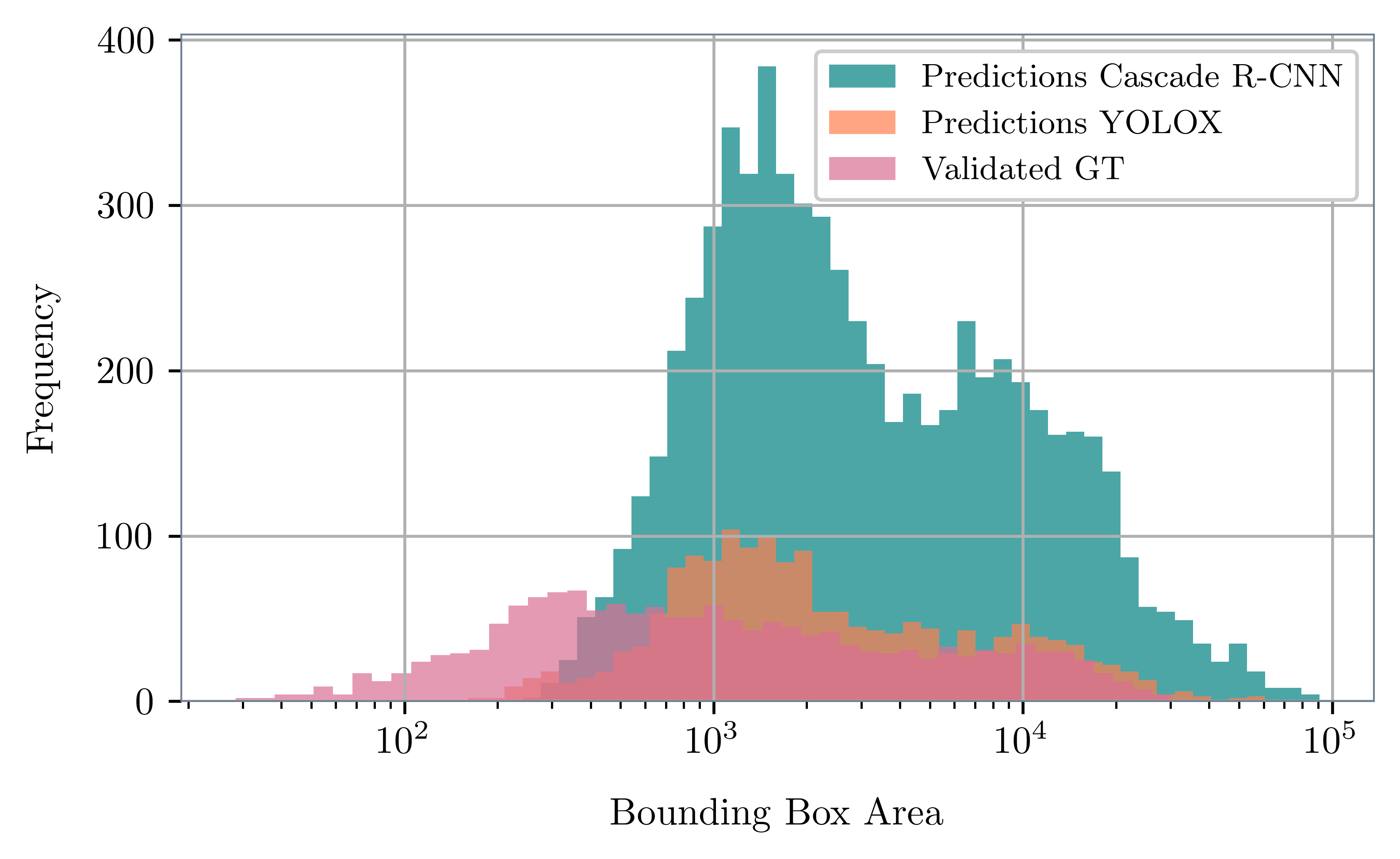}
    \caption{Distribution of bounding box area of validated annotations and predictions of the object detectors YOLOX and Cascade R-CNN for the class \emph{pedestrian}.}
    \label{fig:object_detector_size_of_preds}
\end{figure}

\subsection{Further Numerical Results}

\subsubsection{Quality of Validated GT}

To verify the quality of the validated GT we needed to ensure that we did not miss (FN) or duplicated (FP) pedestrians during the annotation process.
First, we manually audited 200 randomly selected VGT images by expert annotators to check if any pedestrian is missing a bounding box in the dataset.
We only found two examples where one could argue that a pedestrian is missing see \autoref{fig:missed_person_examples}.
Considering that there are 377 VGT boxes with a soft label probability above 0.5 in this sample and we found only two additional pedestrians, this yields a miss rate of \(2/379 \approx 0.5\%\) and confirms the high recall of the VGT. Second, we checked if duplicate bounding boxes exist for 200 randomly selected VGT images.
These are mainly introduced due to the merging of the bounding box strategies.
We found one minor general issue. If the bounding boxes are too different in appearance we may not have caught them in the duplicate detection.
We found potentially 3 duplicate bounding boxes with a score above 0.5 out of all bounding boxes in the sample of 200 images, see \autoref{fig:duplicates_vgt}.
This implies that \(3/377 \approx 0.8\%\) of our boxes above 0.5 are actually not pedestrians.

\subsubsection{Prediction Bounding Box Sizes}
When we analyze the size of the objects predicted by the detectors used in this study, we observe a discrepancy with the validated annotations in \autoref{fig:object_detector_size_of_preds}. Predictions for small objects are rare because these models did not encounter many small objects in the training data.

\begin{figure}[!ht]
    \centering
    \scalebox{0.8}{\includegraphics[trim=100 100 100 160, clip, width=\linewidth]{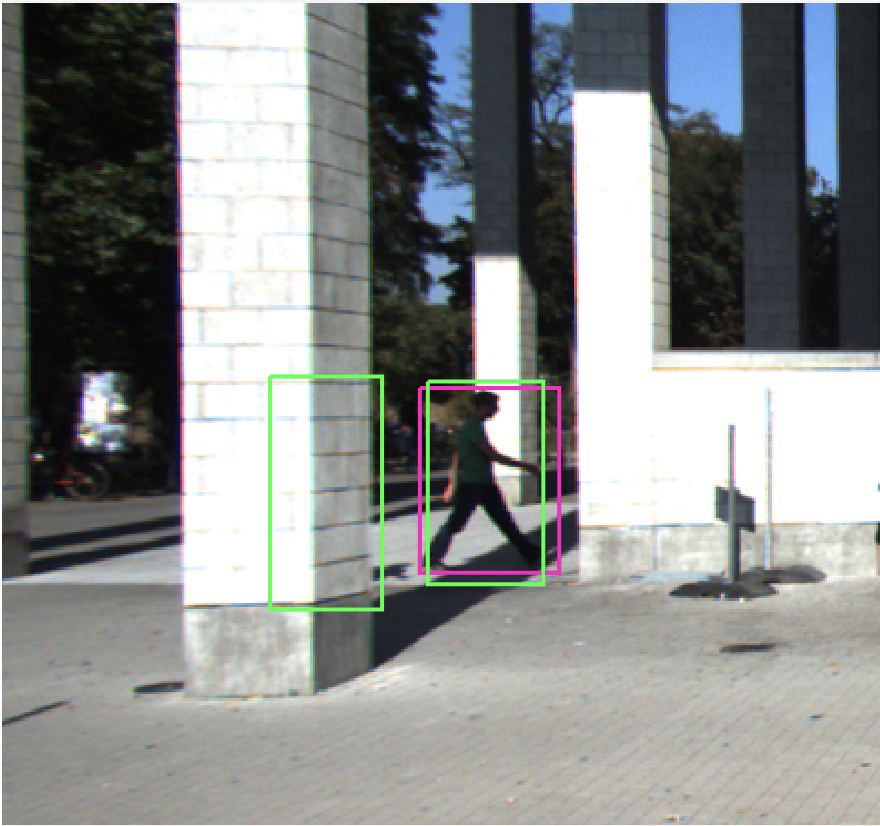}}
    \caption{False positive detection in the KITTI dataset: A bounding box is predicted on a background region where no pedestrian is present.}
    \label{fig:false_positive_1}
\end{figure}

\begin{figure}[!ht]
    \centering
    \scalebox{0.8}{\includegraphics[trim=160 60 100 160, clip, width=\linewidth]{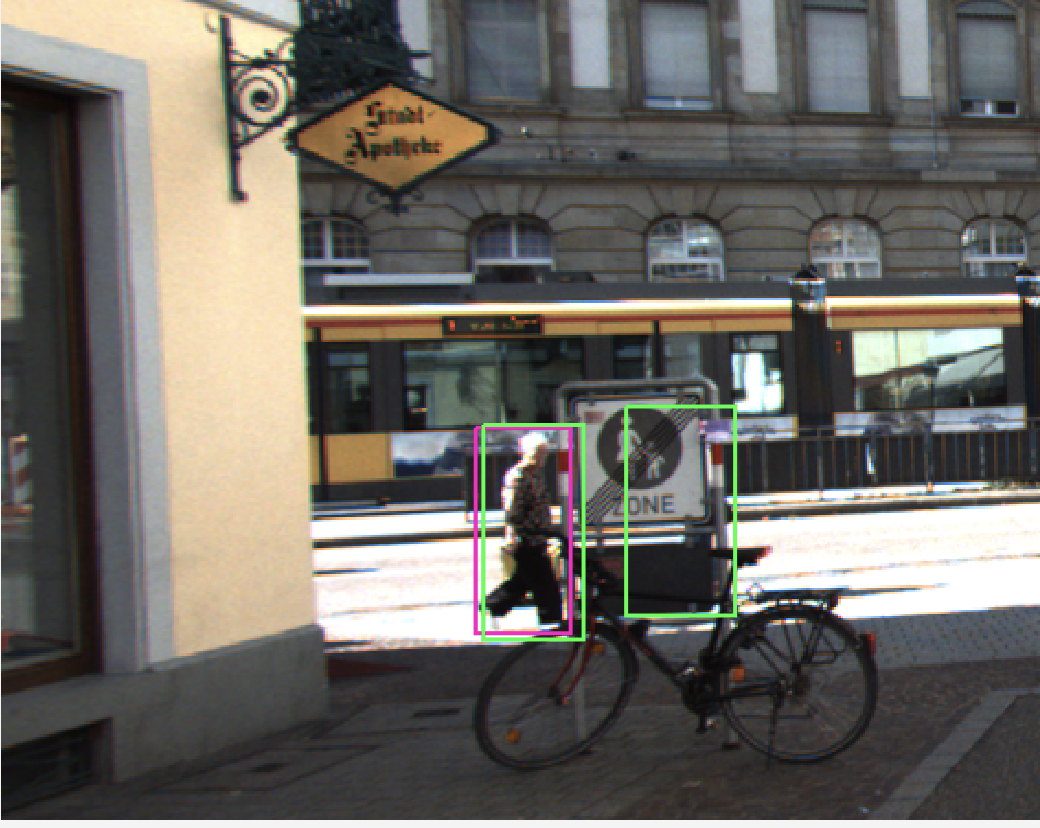}}
    \caption{False positive detection in the KITTI dataset: A bounding box is predicted on a background region where no pedestrian is present.}
    \label{fig:false_positive_2}
\end{figure}

\begin{figure}[!b]
    \centering
    \begin{subfigure}[t]{0.48\linewidth}
        \centering
        \scalebox{0.8}{\includegraphics[trim=200 200 200 200, clip, width=\linewidth]{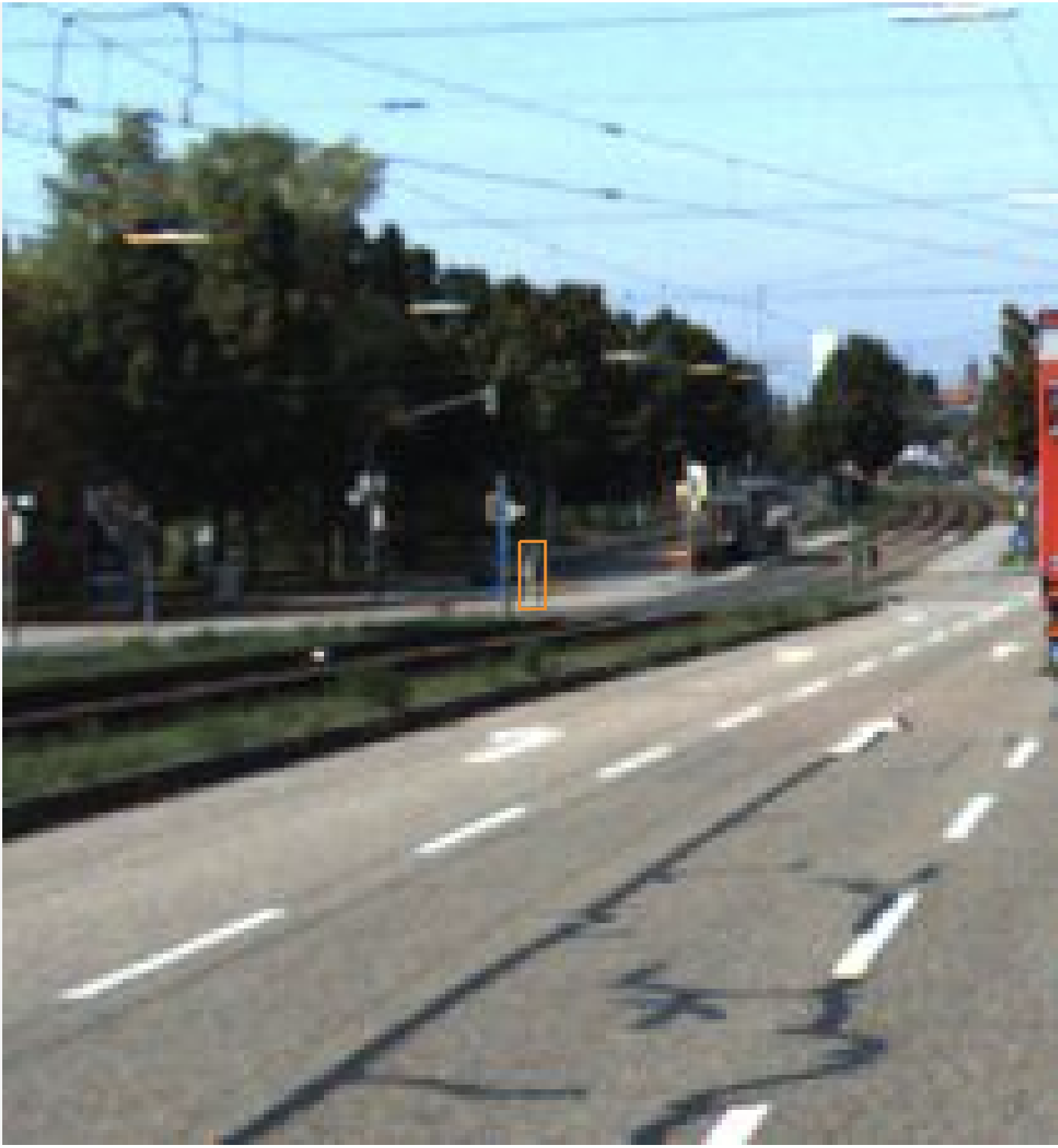}}
        \caption{}
    \end{subfigure}
    \begin{subfigure}[t]{0.48\linewidth}
        \centering
        \scalebox{0.8}{\includegraphics[trim=150 145 150 145, clip, width=\linewidth]{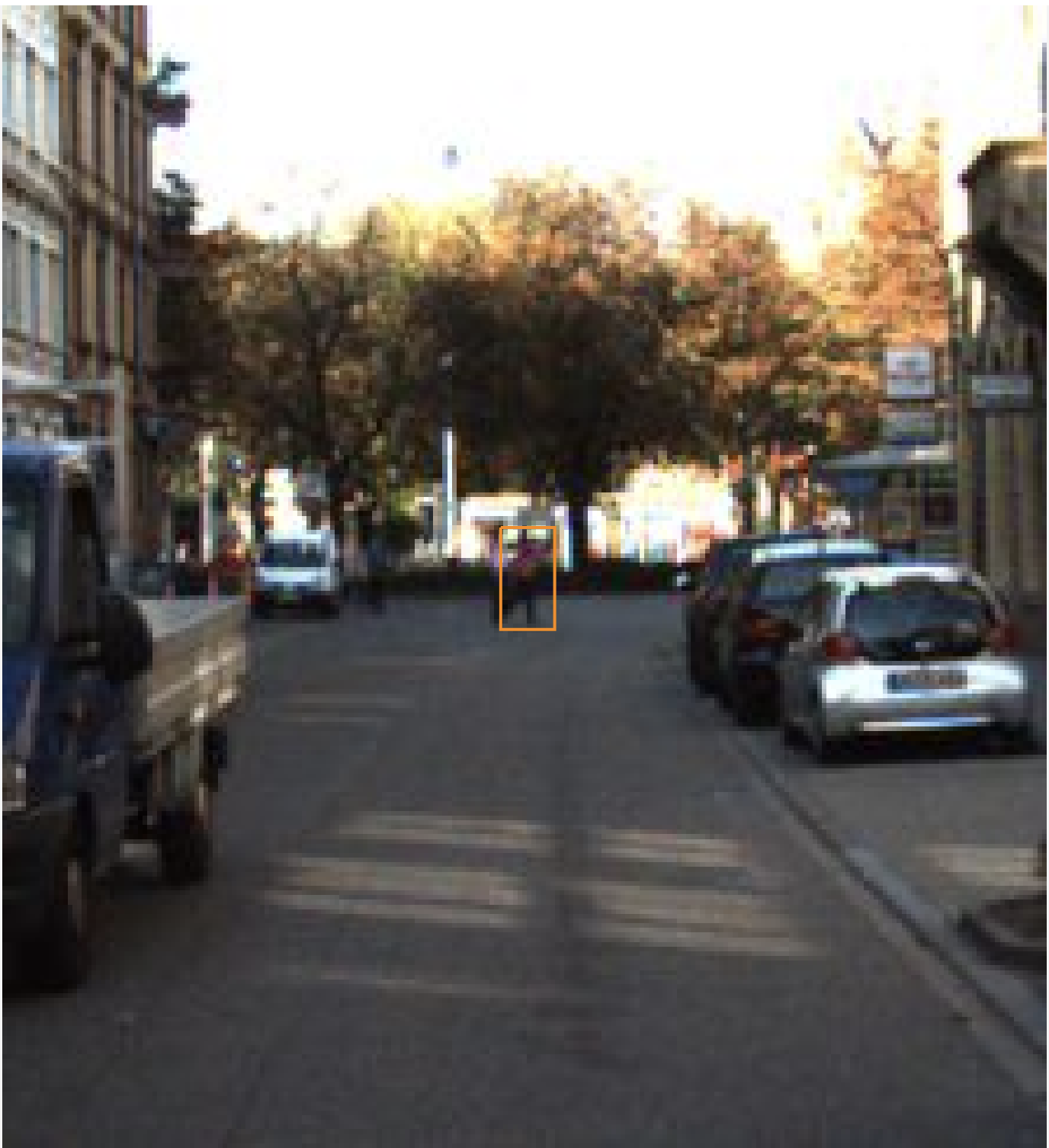}}
        \caption{}
    \end{subfigure}
    \par\medskip
    % \begin{subfigure}[t]{0.48\linewidth}
    %     \centering
    %     \scalebox{0.9}{\includegraphics[trim=100 200 100 200, clip, width=\linewidth]{imgs/Ambigious_3.png}}
    %     \caption{}
    % \end{subfigure}
    % \begin{subfigure}[t]{0.48\linewidth}
    %     \centering
    %     \scalebox{0.9}{\includegraphics[trim=220 250 200 220, clip, width=\linewidth]{imgs/Ambigious_4.png}}
    %     \caption{}
    % \end{subfigure}
    \caption{Example images with high ambiguity, where annotator disagreement is particularly pronounced.}
    \label{fig:ambigious_1_to_4}
\end{figure}

The primary focus of our work is identifying false negatives, \ie, instances where pedestrians are present in the scene but missing from the original annotations. This task is significantly more challenging, as it requires scanning the entire image to find pedestrians that were not detected at all. In contrast, identifying false positives is comparatively easier, as it only involves verifying whether an annotated bounding box in the original ground truth actually contains a pedestrian.

Although not the central goal of this study, we still identified several 85 false positives in the original KITTI dataset based on our validated ground truth (see Figures~\ref{fig:false_positive_1} and~\ref{fig:false_positive_2}). These findings suggest that our approach can also identify overlabeling errors that would otherwise remain unnoticed.

\subsubsection{Less Restrictive Analysis}

In \autoref{fig:cost_error_comparison_less_strict} we present a less strictly filtered analysis as in the main paper. The analysis of this graph is already conducted in the main paper and thus not repeated here.

\subsection{Visual Examples}

\subsubsection{Analysis of Ambiguity and Uncertainty}
\label{subsec:numerical results, analysis of ambiguity and uncertainty}
We analyze cases with low inter-annotator agreement, shown in ~\autoref{fig:ambigious_1_to_4}, and identify common sources of ambiguity in pedestrian detection: distant pedestrians, poor lighting conditions, and crowded scenes with occlusions.
Prior work has shown that incorporating soft labels and uncertainty measures can improve model calibration, detection accuracy, and downstream performance~\cite{schmarje2022benchmark, Davani2022BeyondMajority}. These methods are particularly effective in crowded or visually complex scenes, where binary labels fail to capture annotation uncertainty. Training with soft labels has also been shown to reduce the impact of missing ground truth labels and provide richer training signals by encoding probabilistic information.

\newif\ifshowimages
% \showimagesfalse
\showimagestrue

\FloatBarrier
\newpage
\onecolumn

 \begin{figure*}[htbp]
    \centering

    \begin{subfigure}[b]{0.41\textwidth}
        \includegraphics[width=\textwidth]{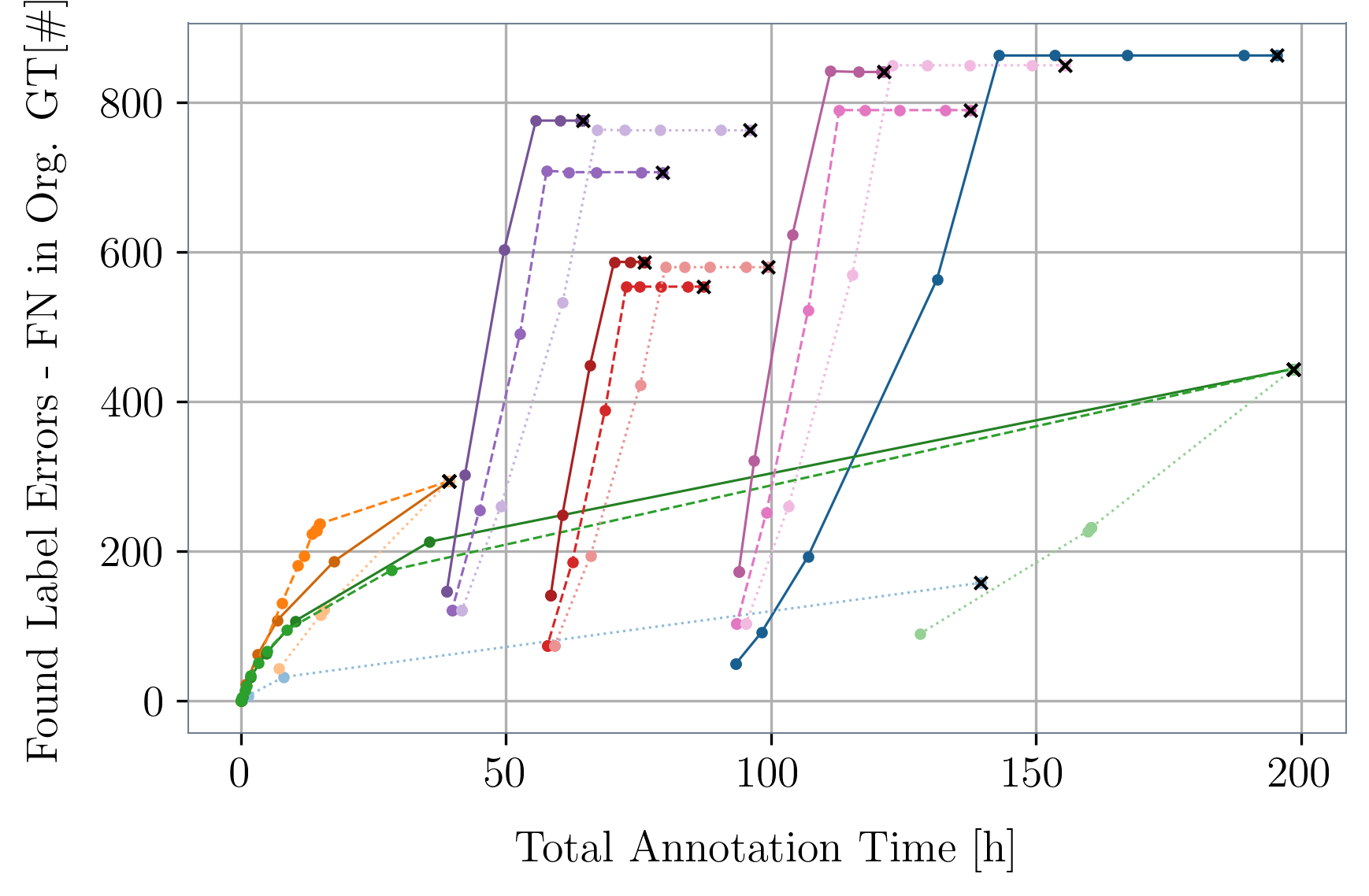}
        \caption{Found Label Errors - FN in original GT}
        \label{fig:cost_error_comparison_less_strict-fn}
    \end{subfigure}
    \hfill
    \begin{subfigure}[b]{0.41\textwidth}
        \includegraphics[width=\textwidth]{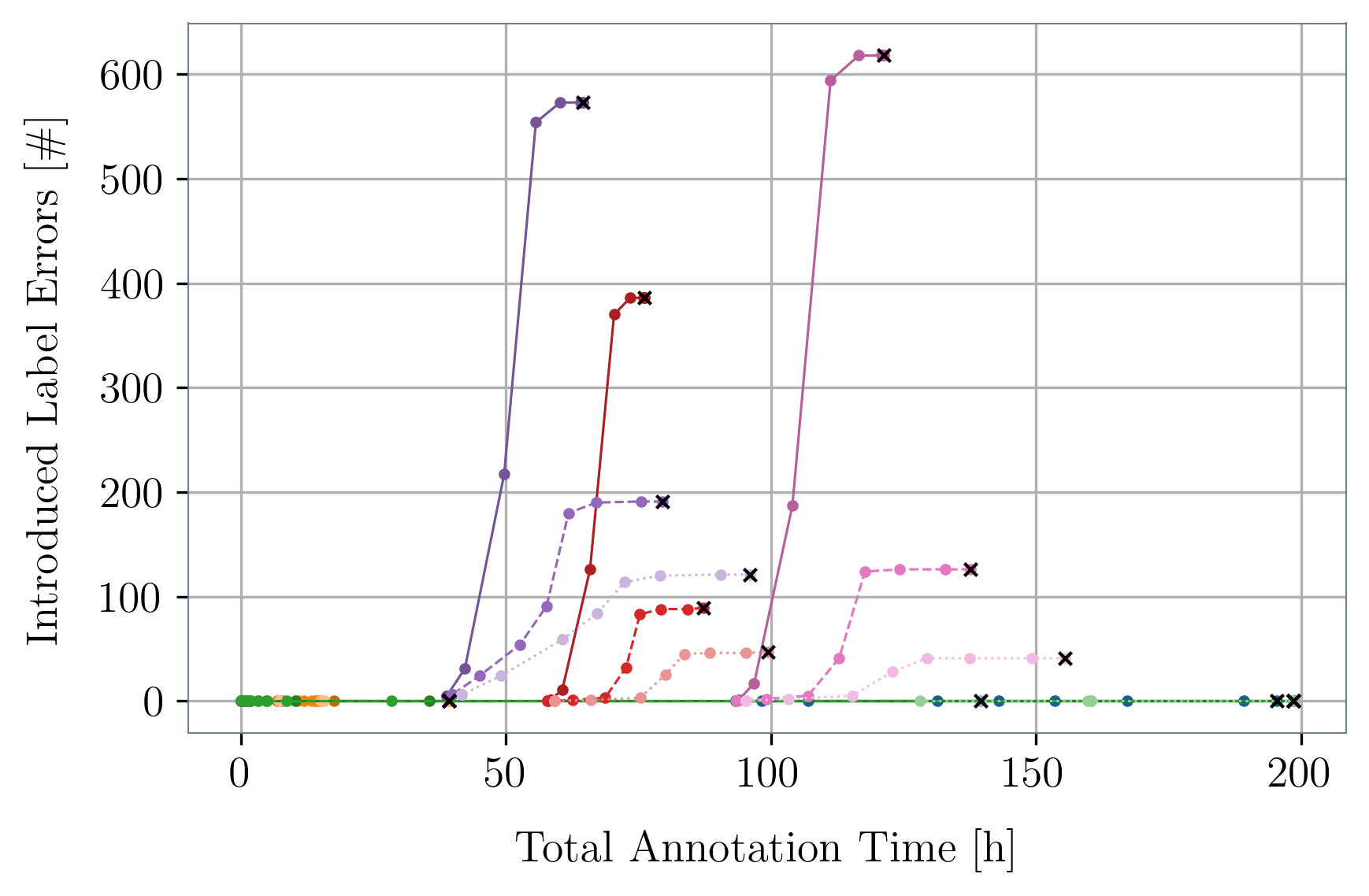}
        \caption{Introduced Label Errors due to low quality labeling strategy}
        \label{fig:cost_error_comparison_less_strict-introduced}
    \end{subfigure}
    \hfill
    \begin{subfigure}[b]{0.15\textwidth}
        \includegraphics[width=\textwidth]{imgs/legend_second_version.png}
        \caption{Legend}
        \label{fig:cost_error_comparison_less_strict-legend}
    \end{subfigure}
    \vspace{0.3cm}
    \caption{
This figure is a less strict version of figure~\ref{fig:cost_error_comparison} where we additionally consider small objects i.e. no filtering according to a minimum pixel height is conducted and a soft label probability of only $0.5$ is required. Consequently, a larger number of label errors in the original annotations can be identified. However, due to the constraints associated with the object detector proposals, we cannot identify all of them automatically.
}
    \label{fig:cost_error_comparison_less_strict}
\end{figure*}

\captionsetup{skip=0pt} % reduce space between caption and image
\setlength{\parskip}{0pt}

\subsubsection{Visual Comparison of Original and Validated GT}

Here, we display examples of the label errors we identified by comparing the validated annotations with the original ones. These examples correspond to the values listed in table~\ref{tab:number_of_label_errors} of the main manuscript. In the following, we consider only those label errors outside of \emph{don't care} regions of the dataset, \ie, the lower half of table~\ref{tab:number_of_label_errors}. First, we display all of the relatively large overlooked pedestrians and inaccurate annotations for which our validated annotations exhibit a high confidence. Then, we show random examples of other label errors, \eg representing smaller objects or ambiguous cases, as distinguished in table~\ref{tab:number_of_label_errors}. We set a common random seed for reproducibility.

\textbf{Most evident overlooked objects.} In the following, we display the overlooked objects (in red color) that are not contained in original annotations (green color). These examples have a soft label probability of $0.8$ or higher as well as a bounding box height of $40$ pixels or more. We identified $42$ such cases and order them according to the soft label probability of the validated annotations and include the original filename for verification purposes.

\ifshowimages
\noindent
\begin{minipage}[t]{0.24\linewidth}
      \centering
      \includegraphics[width=\linewidth, , height=\linewidth, keepaspectratio]{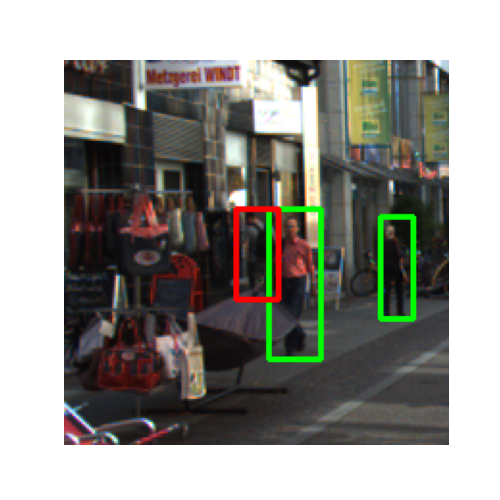}
      \captionsetup{labelformat=empty, hypcap=false}
      \captionof{figure}{007265.png}
    \end{minipage}
\begin{minipage}[t]{0.24\linewidth}
      \centering
      \includegraphics[width=\linewidth, , height=\linewidth, keepaspectratio]{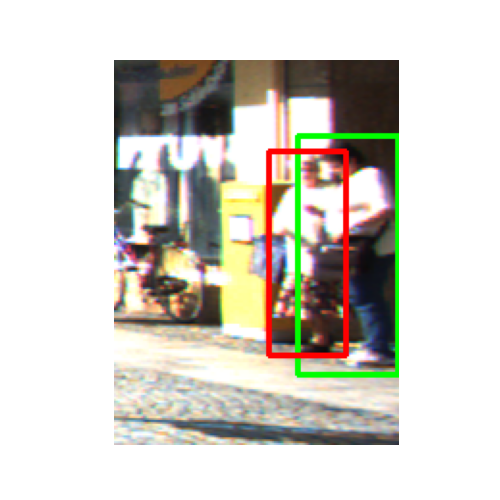}
      \captionsetup{labelformat=empty, hypcap=false}
      \captionof{figure}{007286.png}
    \end{minipage}
\begin{minipage}[t]{0.24\linewidth}
      \centering
      \includegraphics[width=\linewidth, , height=\linewidth, keepaspectratio]{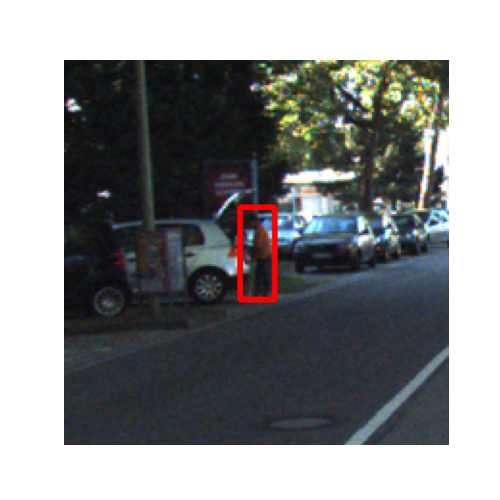}
      \captionsetup{labelformat=empty, hypcap=false}
      \captionof{figure}{005367.png}
    \end{minipage}
\begin{minipage}[t]{0.24\linewidth}
      \centering
      \includegraphics[width=\linewidth, , height=\linewidth, keepaspectratio]{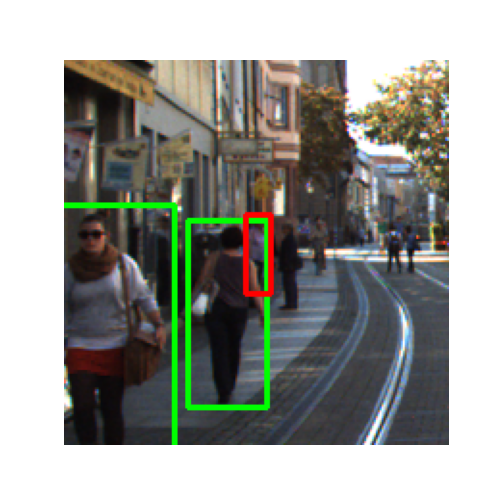}
      \captionsetup{labelformat=empty, hypcap=false}
      \captionof{figure}{003018.png}
    \end{minipage}
\par
\noindent
\begin{minipage}[t]{0.24\linewidth}
      \centering
      \includegraphics[width=\linewidth, , height=\linewidth, keepaspectratio]{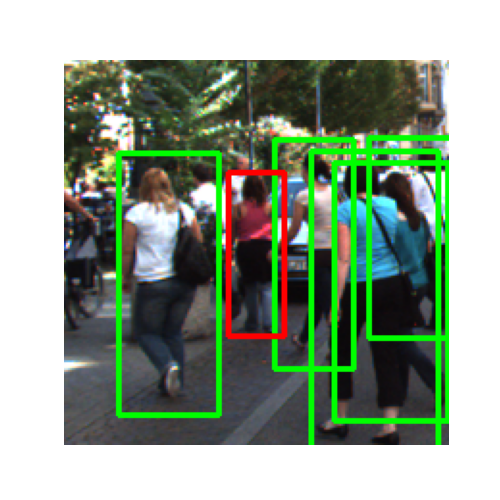}
      \captionsetup{labelformat=empty, hypcap=false}
      \captionof{figure}{006427.png}
    \end{minipage}
\begin{minipage}[t]{0.24\linewidth}
      \centering
      \includegraphics[width=\linewidth, , height=\linewidth, keepaspectratio]{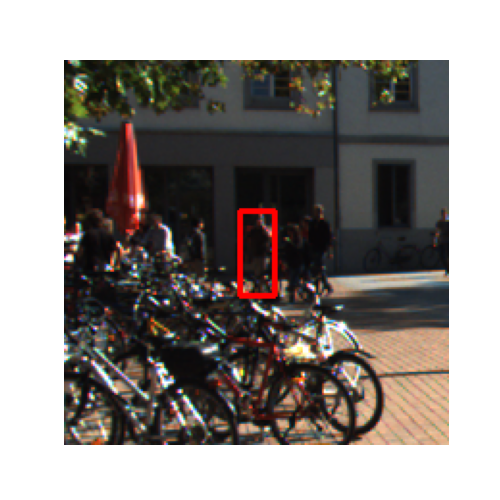}
      \captionsetup{labelformat=empty, hypcap=false}
      \captionof{figure}{000189.png}
    \end{minipage}
\begin{minipage}[t]{0.24\linewidth}
      \centering
      \includegraphics[width=\linewidth, , height=\linewidth, keepaspectratio]{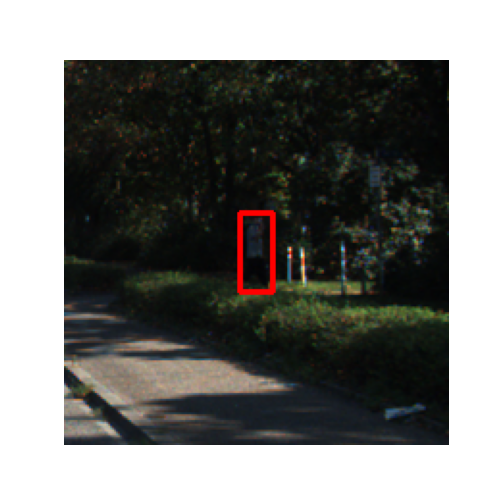}
      \captionsetup{labelformat=empty, hypcap=false}
      \captionof{figure}{006762.png}
    \end{minipage}
\begin{minipage}[t]{0.24\linewidth}
      \centering
      \includegraphics[width=\linewidth, , height=\linewidth, keepaspectratio]{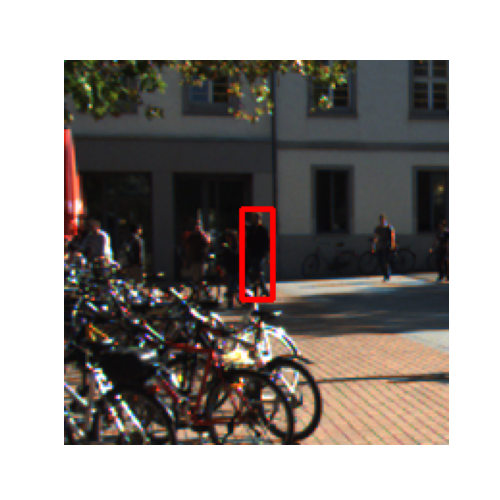}
      \captionsetup{labelformat=empty, hypcap=false}
      \captionof{figure}{000189.png}
    \end{minipage}
\par
\noindent
\begin{minipage}[t]{0.24\linewidth}
      \centering
      \includegraphics[width=\linewidth, , height=\linewidth, keepaspectratio]{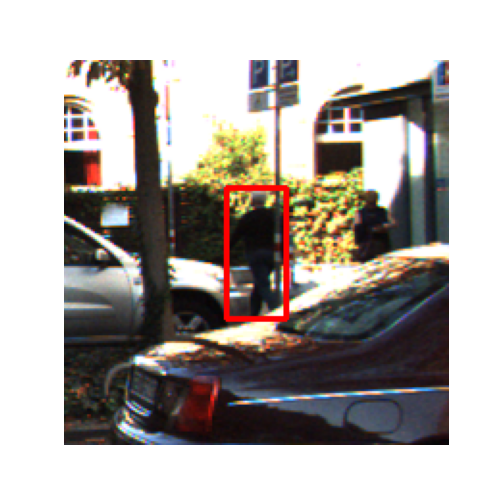}
      \captionsetup{labelformat=empty, hypcap=false}
      \captionof{figure}{003206.png}
    \end{minipage}
\begin{minipage}[t]{0.24\linewidth}
      \centering
      \includegraphics[width=\linewidth, , height=\linewidth, keepaspectratio]{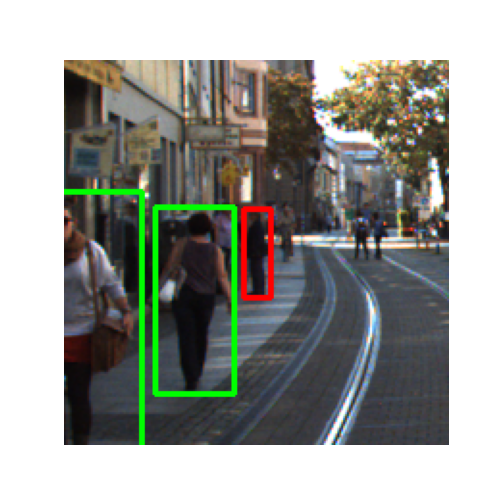}
      \captionsetup{labelformat=empty, hypcap=false}
      \captionof{figure}{003018.png}
    \end{minipage}
\begin{minipage}[t]{0.24\linewidth}
      \centering
      \includegraphics[width=\linewidth, , height=\linewidth, keepaspectratio]{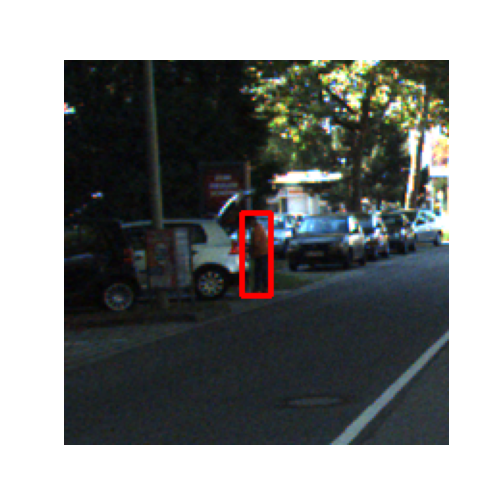}
      \captionsetup{labelformat=empty, hypcap=false}
      \captionof{figure}{000441.png}
    \end{minipage}
\begin{minipage}[t]{0.24\linewidth}
      \centering
      \includegraphics[width=\linewidth, , height=\linewidth, keepaspectratio]{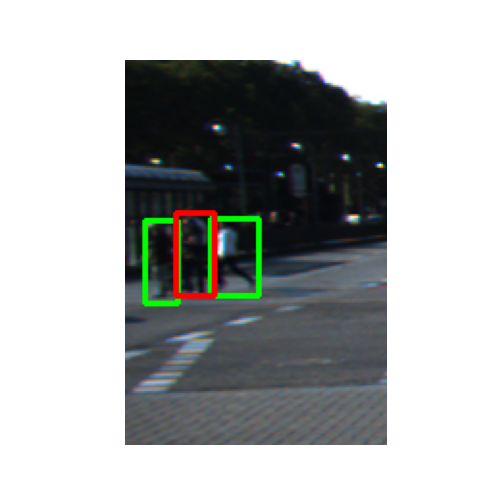}
      \captionsetup{labelformat=empty, hypcap=false}
      \captionof{figure}{004215.png}
    \end{minipage}
\par
\noindent
\begin{minipage}[t]{0.24\linewidth}
      \centering
      \includegraphics[width=\linewidth, , height=\linewidth, keepaspectratio]{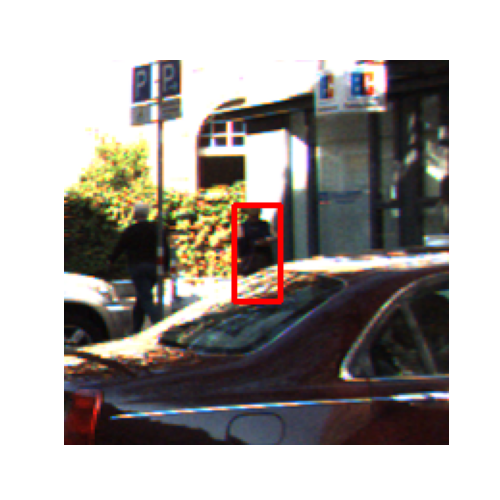}
      \captionsetup{labelformat=empty, hypcap=false}
      \captionof{figure}{003206.png}
    \end{minipage}
\begin{minipage}[t]{0.24\linewidth}
      \centering
      \includegraphics[width=\linewidth, , height=\linewidth, keepaspectratio]{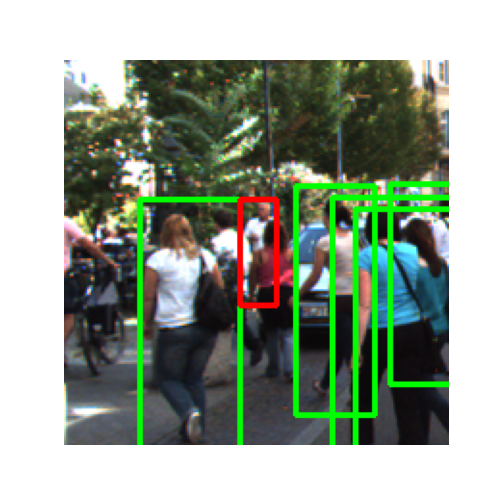}
      \captionsetup{labelformat=empty, hypcap=false}
      \captionof{figure}{006427.png}
    \end{minipage}
\begin{minipage}[t]{0.24\linewidth}
      \centering
      \includegraphics[width=\linewidth, , height=\linewidth, keepaspectratio]{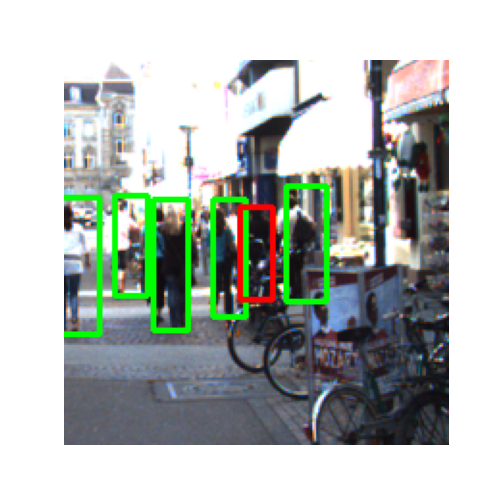}
      \captionsetup{labelformat=empty, hypcap=false}
      \captionof{figure}{007420.png}
    \end{minipage}
\begin{minipage}[t]{0.24\linewidth}
      \centering
      \includegraphics[width=\linewidth, , height=\linewidth, keepaspectratio]{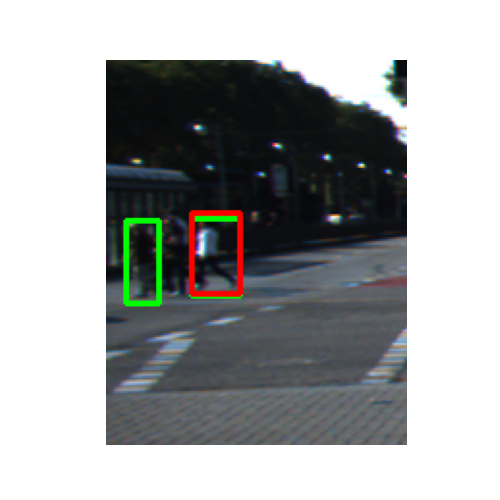}
      \captionsetup{labelformat=empty, hypcap=false}
      \captionof{figure}{004215.png}
    \end{minipage}
\par
\noindent
\begin{minipage}[t]{0.24\linewidth}
      \centering
      \includegraphics[width=\linewidth, , height=\linewidth, keepaspectratio]{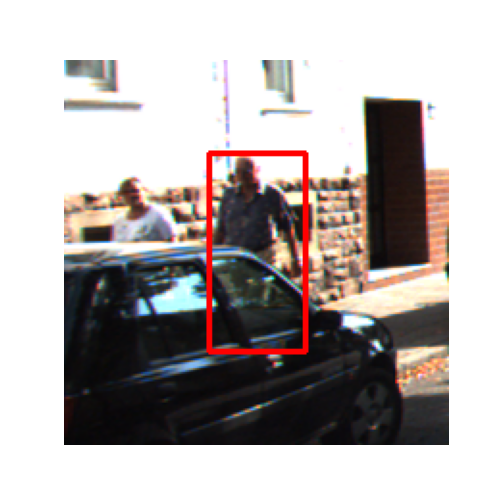}
      \captionsetup{labelformat=empty, hypcap=false}
      \captionof{figure}{005157.png}
    \end{minipage}
\begin{minipage}[t]{0.24\linewidth}
      \centering
      \includegraphics[width=\linewidth, , height=\linewidth, keepaspectratio]{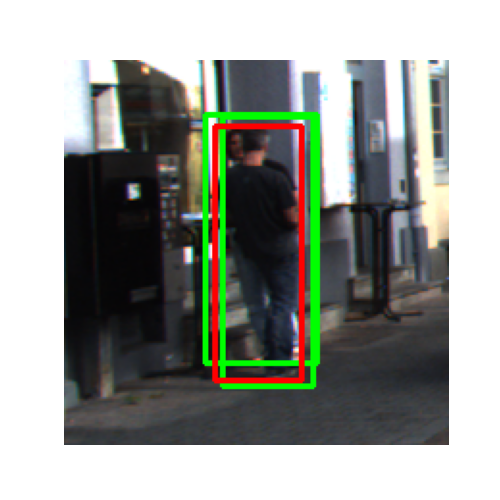}
      \captionsetup{labelformat=empty, hypcap=false}
      \captionof{figure}{000341.png}
    \end{minipage}
\begin{minipage}[t]{0.24\linewidth}
      \centering
      \includegraphics[width=\linewidth, , height=\linewidth, keepaspectratio]{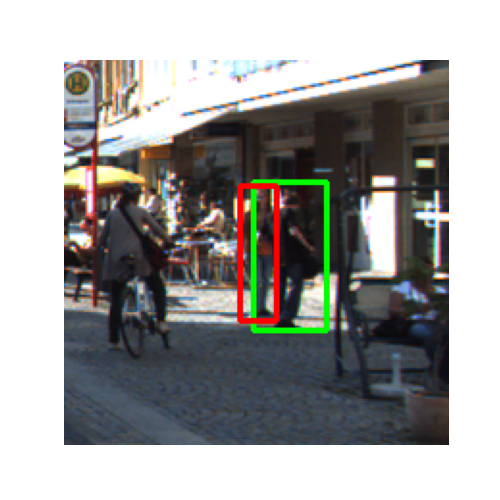}
      \captionsetup{labelformat=empty, hypcap=false}
      \captionof{figure}{001981.png}
    \end{minipage}
\begin{minipage}[t]{0.24\linewidth}
      \centering
      \includegraphics[width=\linewidth, , height=\linewidth, keepaspectratio]{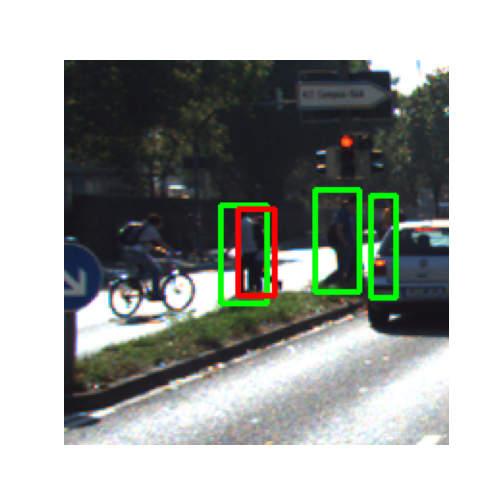}
      \captionsetup{labelformat=empty, hypcap=false}
      \captionof{figure}{001640.png}
    \end{minipage}
\par
\noindent
\begin{minipage}[t]{0.24\linewidth}
      \centering
      \includegraphics[width=\linewidth, , height=\linewidth, keepaspectratio]{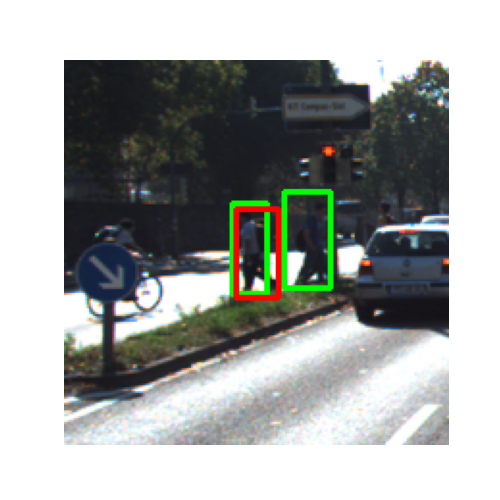}
      \captionsetup{labelformat=empty, hypcap=false}
      \captionof{figure}{004788.png}
    \end{minipage}
\begin{minipage}[t]{0.24\linewidth}
      \centering
      \includegraphics[width=\linewidth, , height=\linewidth, keepaspectratio]{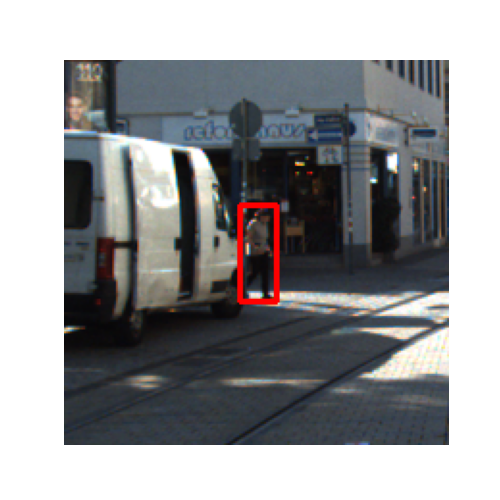}
      \captionsetup{labelformat=empty, hypcap=false}
      \captionof{figure}{007286.png}
    \end{minipage}
\begin{minipage}[t]{0.24\linewidth}
      \centering
      \includegraphics[width=\linewidth, , height=\linewidth, keepaspectratio]{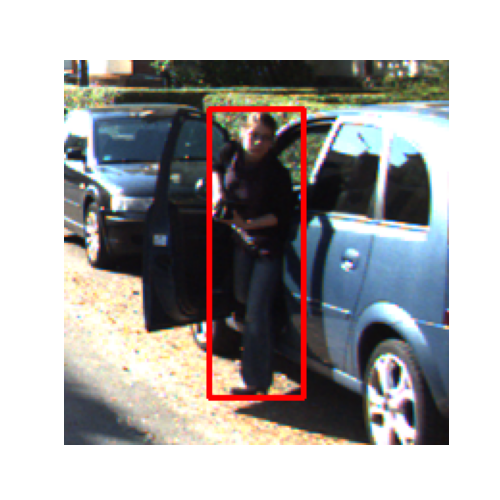}
      \captionsetup{labelformat=empty, hypcap=false}
      \captionof{figure}{001035.png}
    \end{minipage}
\begin{minipage}[t]{0.24\linewidth}
      \centering
      \includegraphics[width=\linewidth, , height=\linewidth, keepaspectratio]{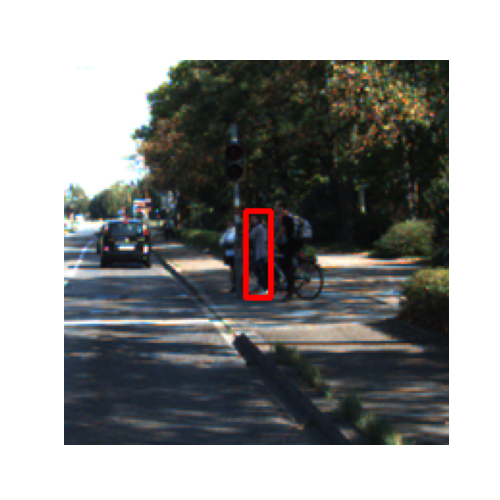}
      \captionsetup{labelformat=empty, hypcap=false}
      \captionof{figure}{003302.png}
    \end{minipage}
\par
\noindent
\begin{minipage}[t]{0.24\linewidth}
      \centering
      \includegraphics[width=\linewidth, , height=\linewidth, keepaspectratio]{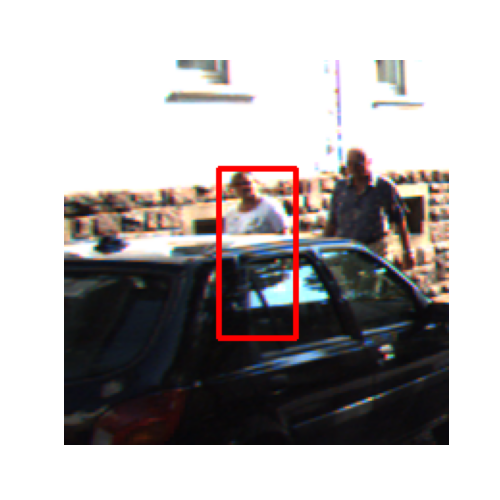}
      \captionsetup{labelformat=empty, hypcap=false}
      \captionof{figure}{005157.png}
    \end{minipage}
\begin{minipage}[t]{0.24\linewidth}
      \centering
      \includegraphics[width=\linewidth, , height=\linewidth, keepaspectratio]{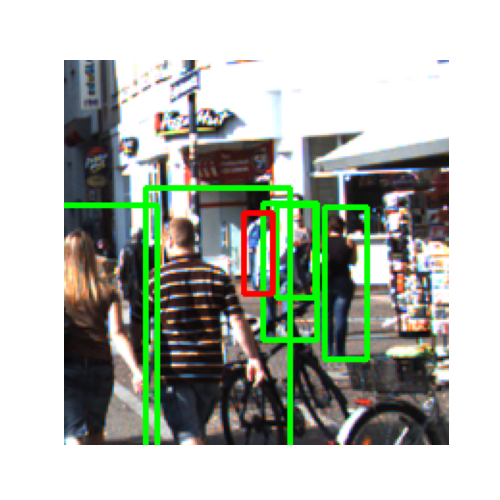}
      \captionsetup{labelformat=empty, hypcap=false}
      \captionof{figure}{000559.png}
    \end{minipage}
\begin{minipage}[t]{0.24\linewidth}
      \centering
      \includegraphics[width=\linewidth, , height=\linewidth, keepaspectratio]{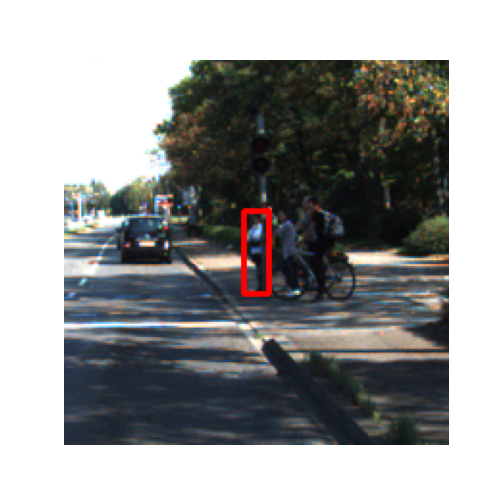}
      \captionsetup{labelformat=empty, hypcap=false}
      \captionof{figure}{000804.png}
    \end{minipage}
\begin{minipage}[t]{0.24\linewidth}
      \centering
      \includegraphics[width=\linewidth, , height=\linewidth, keepaspectratio]{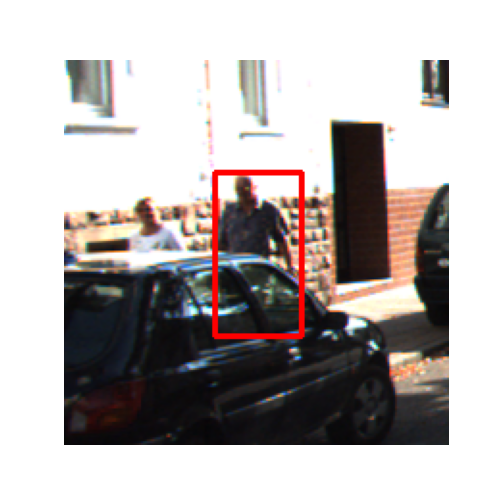}
      \captionsetup{labelformat=empty, hypcap=false}
      \captionof{figure}{000736.png}
    \end{minipage}
\par
\noindent
\begin{minipage}[t]{0.24\linewidth}
      \centering
      \includegraphics[width=\linewidth, , height=\linewidth, keepaspectratio]{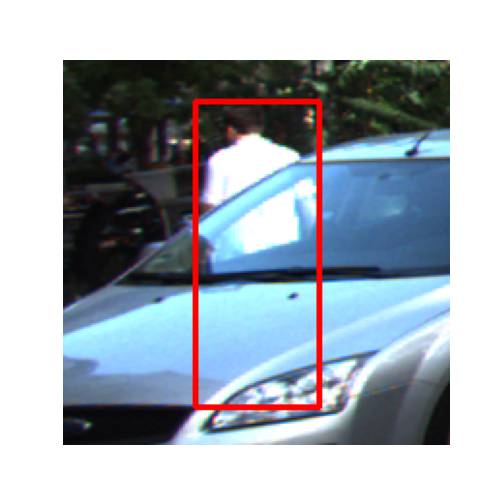}
      \captionsetup{labelformat=empty, hypcap=false}
      \captionof{figure}{007161.png}
    \end{minipage}
\begin{minipage}[t]{0.24\linewidth}
      \centering
      \includegraphics[width=\linewidth, , height=\linewidth, keepaspectratio]{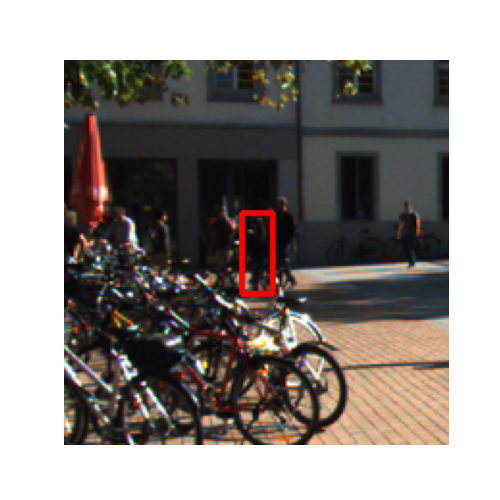}
      \captionsetup{labelformat=empty, hypcap=false}
      \captionof{figure}{000189.png}
    \end{minipage}
\begin{minipage}[t]{0.24\linewidth}
      \centering
      \includegraphics[width=\linewidth, , height=\linewidth, keepaspectratio]{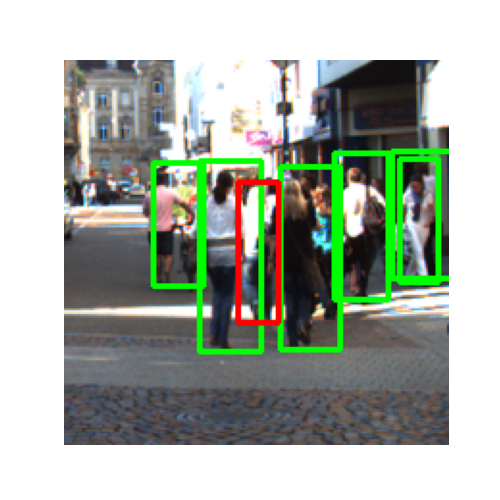}
      \captionsetup{labelformat=empty, hypcap=false}
      \captionof{figure}{002775.png}
    \end{minipage}
\begin{minipage}[t]{0.24\linewidth}
      \centering
      \includegraphics[width=\linewidth, , height=\linewidth, keepaspectratio]{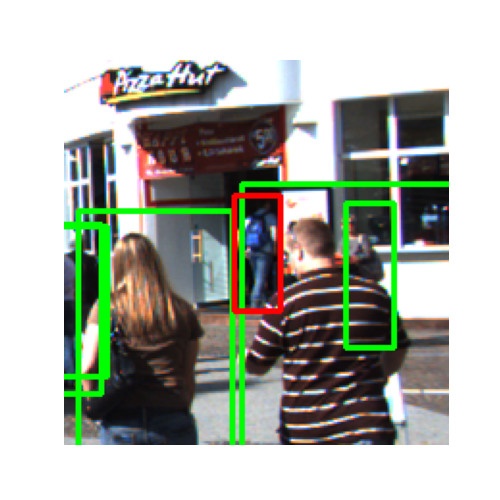}
      \captionsetup{labelformat=empty, hypcap=false}
      \captionof{figure}{006937.png}
    \end{minipage}
\par
\noindent
\begin{minipage}[t]{0.24\linewidth}
      \centering
      \includegraphics[width=\linewidth, , height=\linewidth, keepaspectratio]{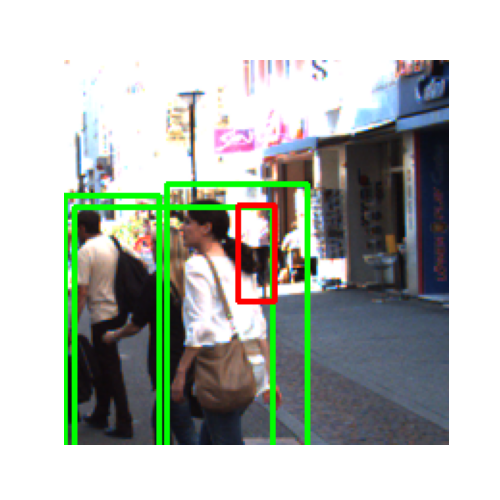}
      \captionsetup{labelformat=empty, hypcap=false}
      \captionof{figure}{006427.png}
    \end{minipage}
\begin{minipage}[t]{0.24\linewidth}
      \centering
      \includegraphics[width=\linewidth, , height=\linewidth, keepaspectratio]{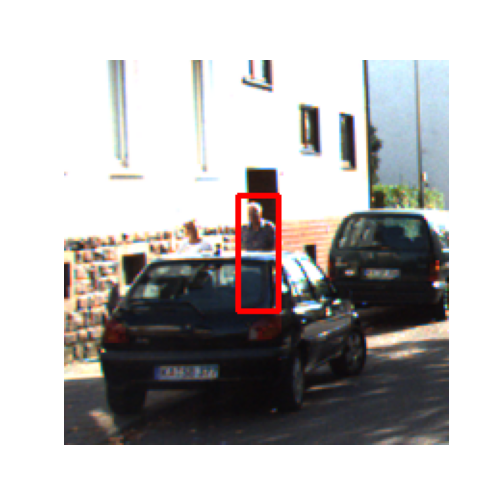}
      \captionsetup{labelformat=empty, hypcap=false}
      \captionof{figure}{005982.png}
    \end{minipage}
\begin{minipage}[t]{0.24\linewidth}
      \centering
      \includegraphics[width=\linewidth, , height=\linewidth, keepaspectratio]{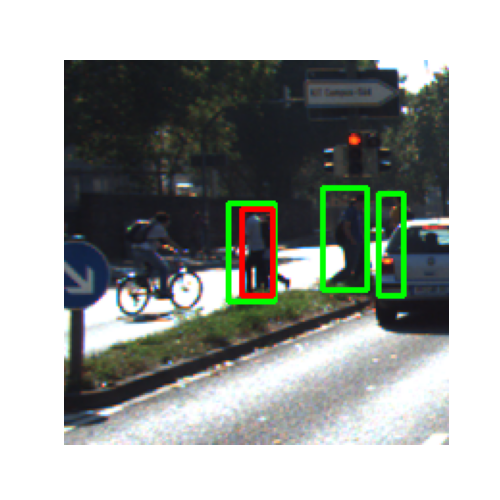}
      \captionsetup{labelformat=empty, hypcap=false}
      \captionof{figure}{001640.png}
    \end{minipage}
\begin{minipage}[t]{0.24\linewidth}
      \centering
      \includegraphics[width=\linewidth, , height=\linewidth, keepaspectratio]{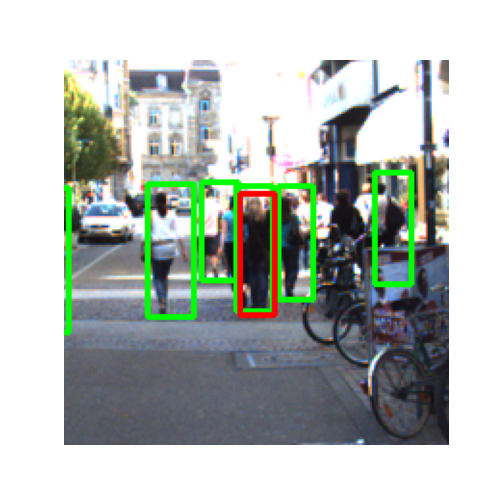}
      \captionsetup{labelformat=empty, hypcap=false}
      \captionof{figure}{003718.png}
    \end{minipage}
\par
\noindent
\begin{minipage}[t]{0.24\linewidth}
      \centering
      \includegraphics[width=\linewidth, , height=\linewidth, keepaspectratio]{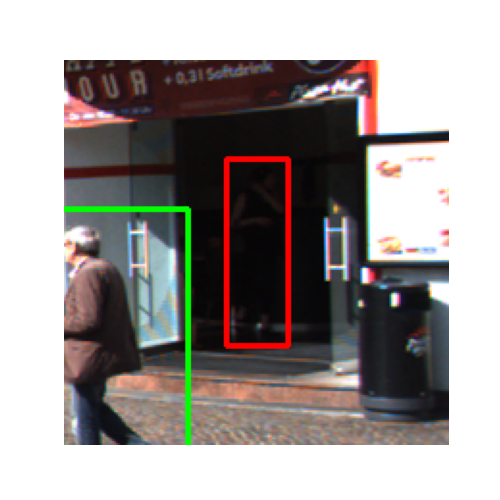}
      \captionsetup{labelformat=empty, hypcap=false}
      \captionof{figure}{004028.png}
    \end{minipage}
\begin{minipage}[t]{0.24\linewidth}
      \centering
      \includegraphics[width=\linewidth, , height=\linewidth, keepaspectratio]{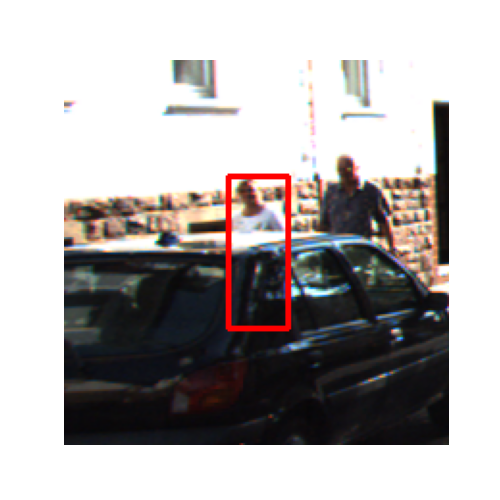}
      \captionsetup{labelformat=empty, hypcap=false}
      \captionof{figure}{000736.png}
    \end{minipage}
\begin{minipage}[t]{0.24\linewidth}
      \centering
      \includegraphics[width=\linewidth, , height=\linewidth, keepaspectratio]{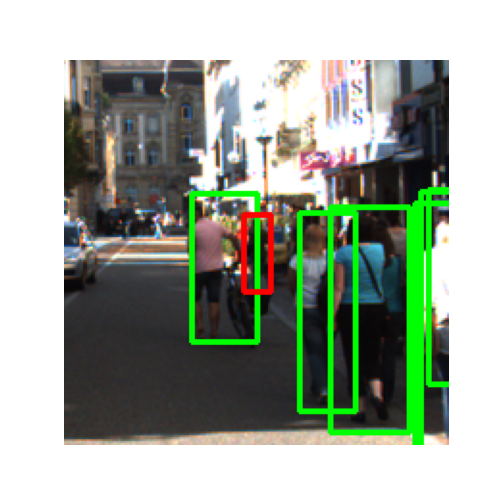}
      \captionsetup{labelformat=empty, hypcap=false}
      \captionof{figure}{002315.png}
    \end{minipage}
\begin{minipage}[t]{0.24\linewidth}
      \centering
      \includegraphics[width=\linewidth, , height=\linewidth, keepaspectratio]{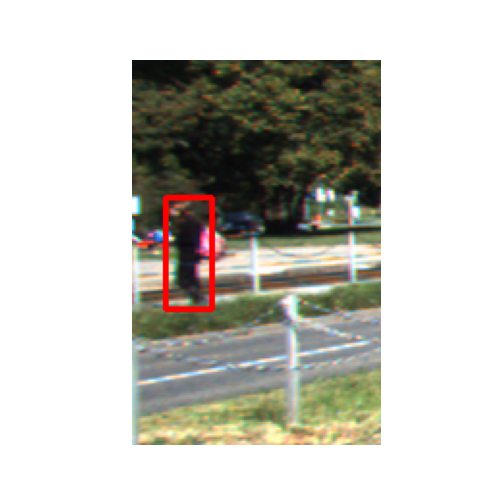}
      \captionsetup{labelformat=empty, hypcap=false}
      \captionof{figure}{007157.png}
    \end{minipage}
\par
\noindent
\begin{minipage}[t]{0.23\linewidth}
      \centering
      \includegraphics[width=\linewidth, , height=\linewidth, keepaspectratio]{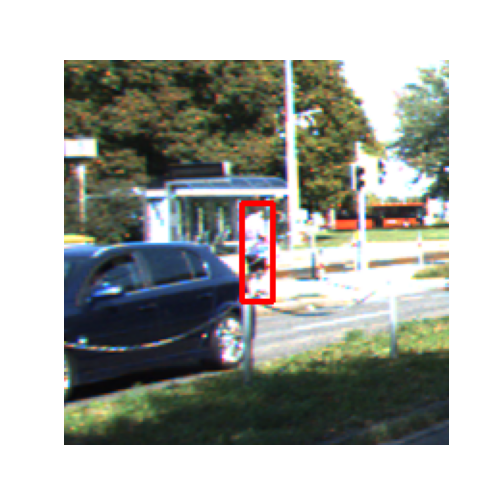}
      \captionsetup{labelformat=empty, hypcap=false}
      \captionof{figure}{003302.png}
    \end{minipage}
\begin{minipage}[t]{0.23\linewidth}
      \centering
      \includegraphics[width=\linewidth, , height=\linewidth, keepaspectratio]{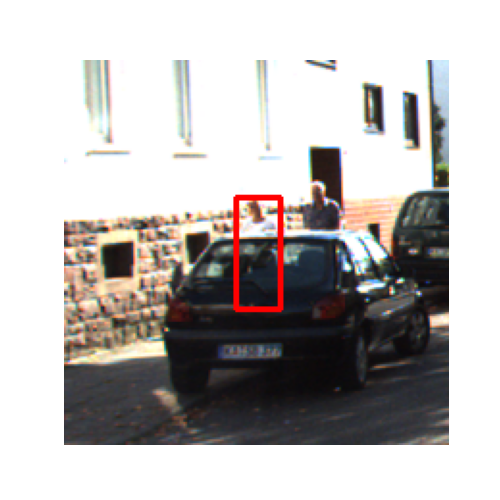}
      \captionsetup{labelformat=empty, hypcap=false}
      \captionof{figure}{005982.png}
    \end{minipage}
\fi

\textbf{Most evident misfitting bounding boxes.} 
Original annotations may be inaccurate. If the original annotation does not exhibit an $\iou$ of $0.5$ with the validated one, we consider the original bounding box as misfitting. This label error is less obvious and may depend on subjectivity. We list the first $20$ of the $87$ most evident mismatches (according to minimal bounding box height and soft label probability) we identified, sorted by the difference in $\iou$ with the VGT. Note that many of these errors can be interpreted as overlooked pedestrians but they overlap with an original bounding box annotation, e.g., due to duplicate GT bounding boxes or overlooked pedestrians being close to annotated ones.

\ifshowimages
\noindent
\begin{minipage}[t]{0.24\linewidth}
      \centering
      \includegraphics[width=\linewidth, , height=\linewidth, keepaspectratio]{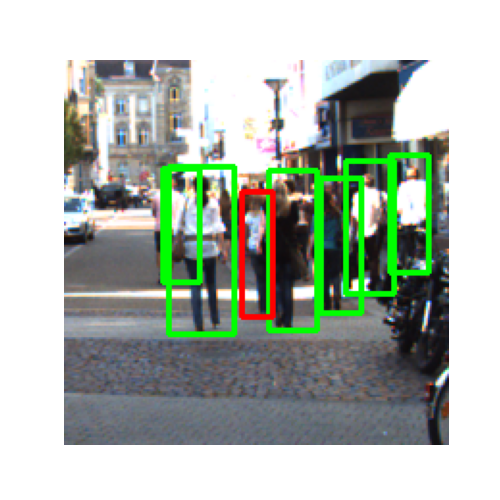}
      \captionsetup{labelformat=empty, hypcap=false}
      \captionof{figure}{005510.png}
    \end{minipage}
\begin{minipage}[t]{0.24\linewidth}
      \centering
      \includegraphics[width=\linewidth, , height=\linewidth, keepaspectratio]{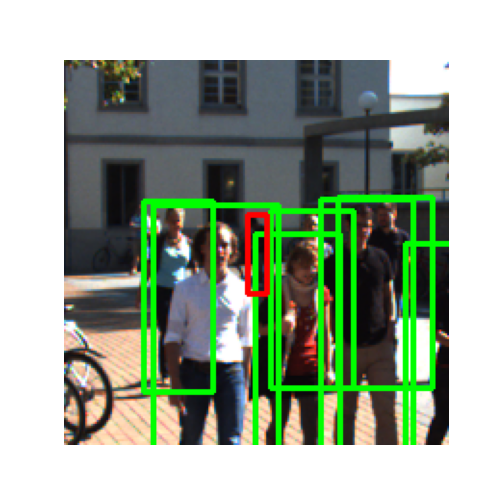}
      \captionsetup{labelformat=empty, hypcap=false}
      \captionof{figure}{000966.png}
    \end{minipage}
\begin{minipage}[t]{0.24\linewidth}
      \centering
      \includegraphics[width=\linewidth, , height=\linewidth, keepaspectratio]{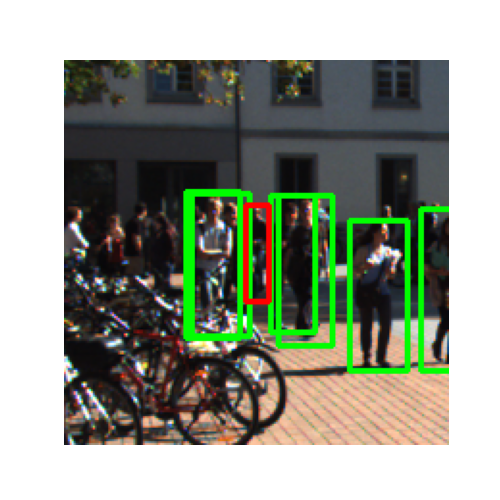}
      \captionsetup{labelformat=empty, hypcap=false}
      \captionof{figure}{005532.png}
    \end{minipage}
\begin{minipage}[t]{0.24\linewidth}
      \centering
      \includegraphics[width=\linewidth, , height=\linewidth, keepaspectratio]{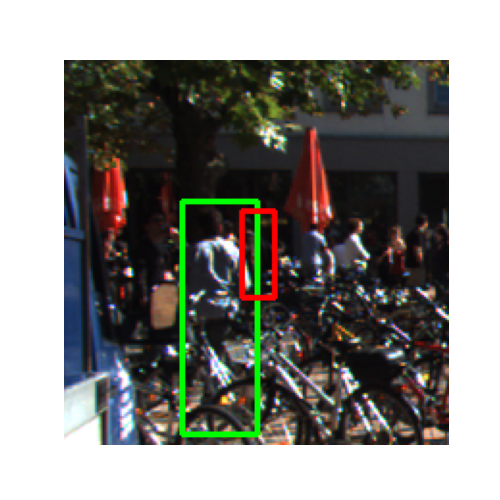}
      \captionsetup{labelformat=empty, hypcap=false}
      \captionof{figure}{005532.png}
    \end{minipage}
\par
\noindent
\begin{minipage}[t]{0.24\linewidth}
      \centering
      \includegraphics[width=\linewidth, , height=\linewidth, keepaspectratio]{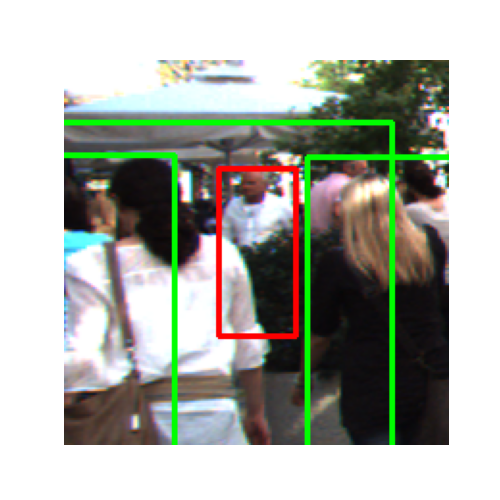}
      \captionsetup{labelformat=empty, hypcap=false}
      \captionof{figure}{004443.png}
    \end{minipage}
\begin{minipage}[t]{0.24\linewidth}
      \centering
      \includegraphics[width=\linewidth, , height=\linewidth, keepaspectratio]{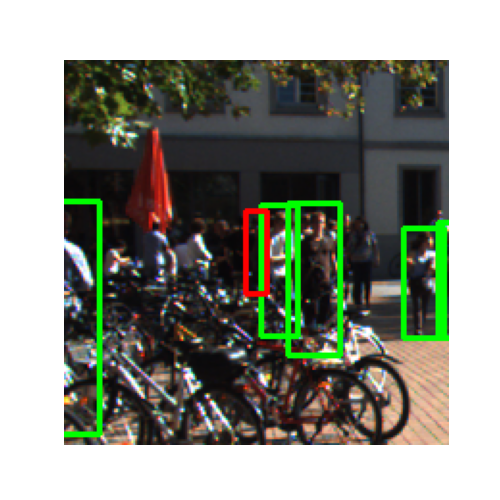}
      \captionsetup{labelformat=empty, hypcap=false}
      \captionof{figure}{000625.png}
    \end{minipage}
\begin{minipage}[t]{0.24\linewidth}
      \centering
      \includegraphics[width=\linewidth, , height=\linewidth, keepaspectratio]{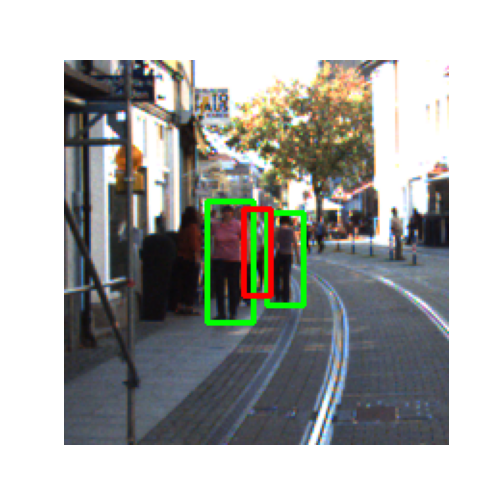}
      \captionsetup{labelformat=empty, hypcap=false}
      \captionof{figure}{000937.png}
    \end{minipage}
\begin{minipage}[t]{0.24\linewidth}
      \centering
      \includegraphics[width=\linewidth, , height=\linewidth, keepaspectratio]{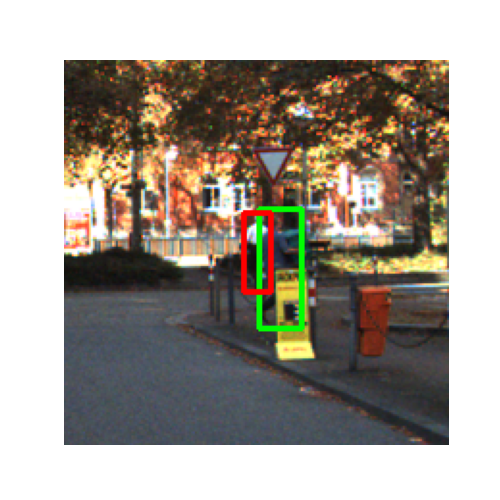}
      \captionsetup{labelformat=empty, hypcap=false}
      \captionof{figure}{004273.png}
    \end{minipage}
\par
\noindent
\begin{minipage}[t]{0.24\linewidth}
      \centering
      \includegraphics[width=\linewidth, , height=\linewidth, keepaspectratio]{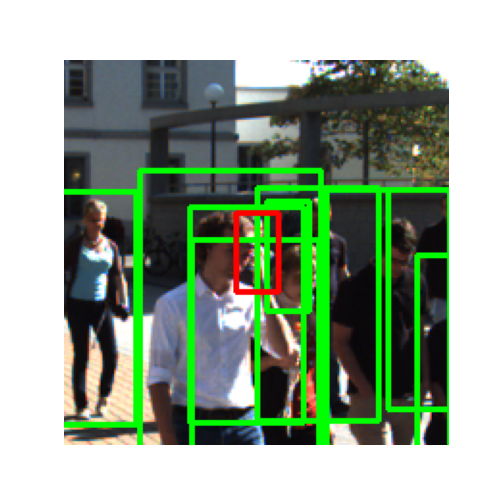}
      \captionsetup{labelformat=empty, hypcap=false}
      \captionof{figure}{003502.png}
    \end{minipage}
\begin{minipage}[t]{0.24\linewidth}
      \centering
      \includegraphics[width=\linewidth, , height=\linewidth, keepaspectratio]{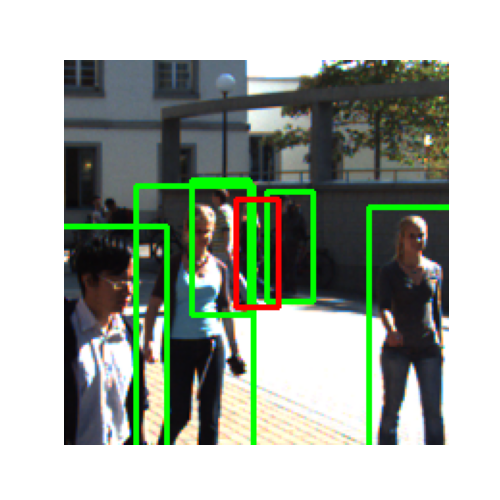}
      \captionsetup{labelformat=empty, hypcap=false}
      \captionof{figure}{002979.png}
    \end{minipage}
\begin{minipage}[t]{0.24\linewidth}
      \centering
      \includegraphics[width=\linewidth, , height=\linewidth, keepaspectratio]{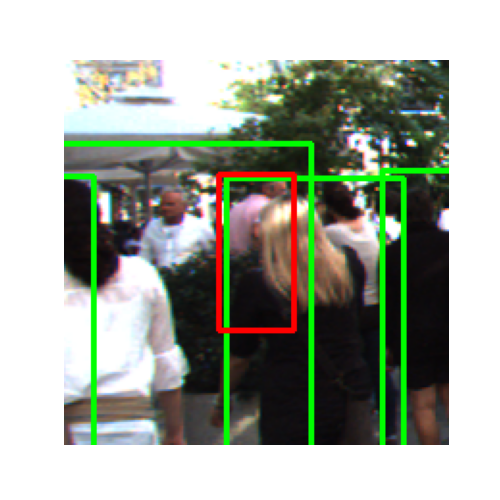}
      \captionsetup{labelformat=empty, hypcap=false}
      \captionof{figure}{004443.png}
    \end{minipage}
\begin{minipage}[t]{0.24\linewidth}
      \centering
      \includegraphics[width=\linewidth, , height=\linewidth, keepaspectratio]{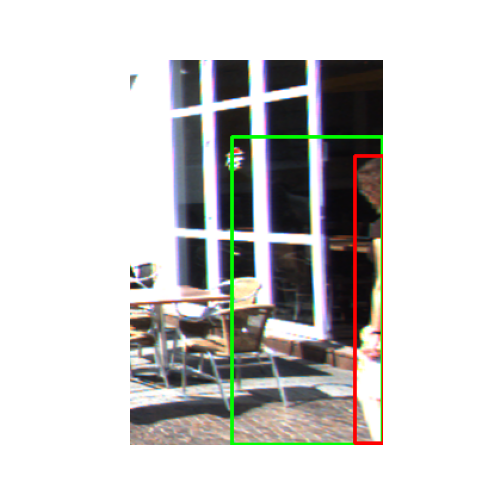}
      \captionsetup{labelformat=empty, hypcap=false}
      \captionof{figure}{000727.png}
    \end{minipage}
\par
\noindent
\begin{minipage}[t]{0.24\linewidth}
      \centering
      \includegraphics[width=\linewidth, , height=\linewidth, keepaspectratio]{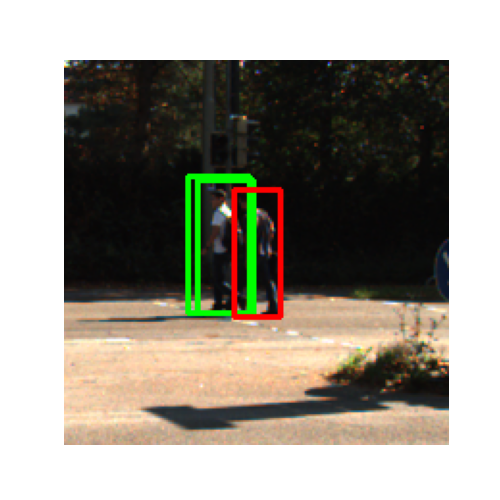}
      \captionsetup{labelformat=empty, hypcap=false}
      \captionof{figure}{006051.png}
    \end{minipage}
\begin{minipage}[t]{0.24\linewidth}
      \centering
      \includegraphics[width=\linewidth, , height=\linewidth, keepaspectratio]{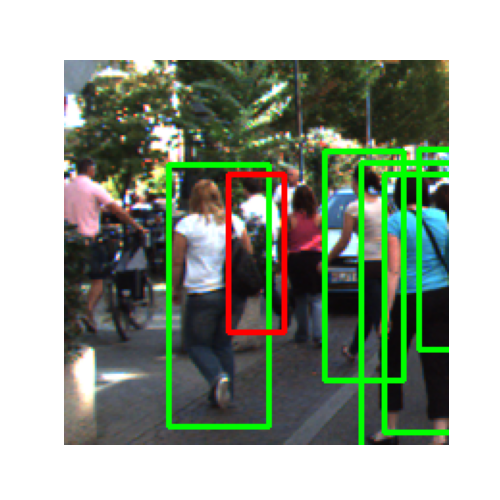}
      \captionsetup{labelformat=empty, hypcap=false}
      \captionof{figure}{006427.png}
    \end{minipage}
\begin{minipage}[t]{0.24\linewidth}
      \centering
      \includegraphics[width=\linewidth, , height=\linewidth, keepaspectratio]{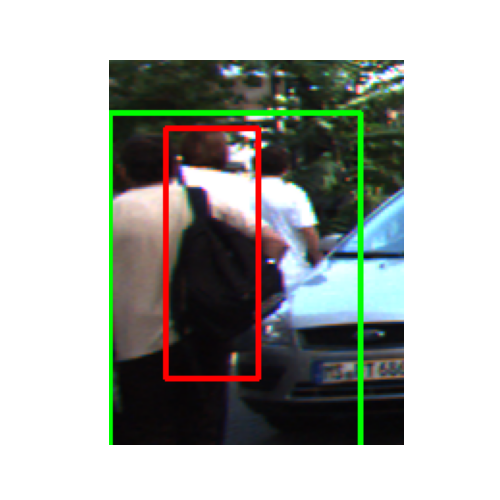}
      \captionsetup{labelformat=empty, hypcap=false}
      \captionof{figure}{005336.png}
    \end{minipage}
\begin{minipage}[t]{0.24\linewidth}
      \centering
      \includegraphics[width=\linewidth, , height=\linewidth, keepaspectratio]{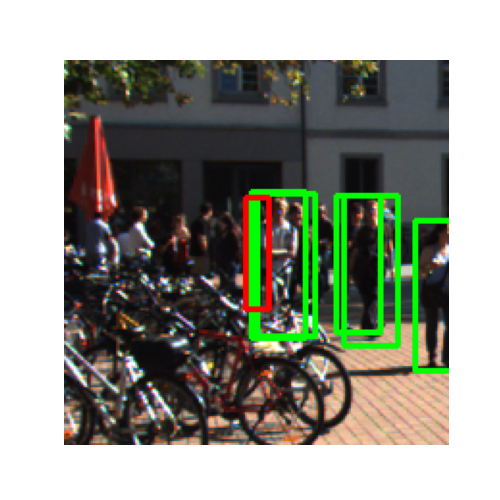}
      \captionsetup{labelformat=empty, hypcap=false}
      \captionof{figure}{005532.png}
    \end{minipage}
\par
\noindent
\begin{minipage}[t]{0.24\linewidth}
      \centering
      \includegraphics[width=\linewidth, , height=\linewidth, keepaspectratio]{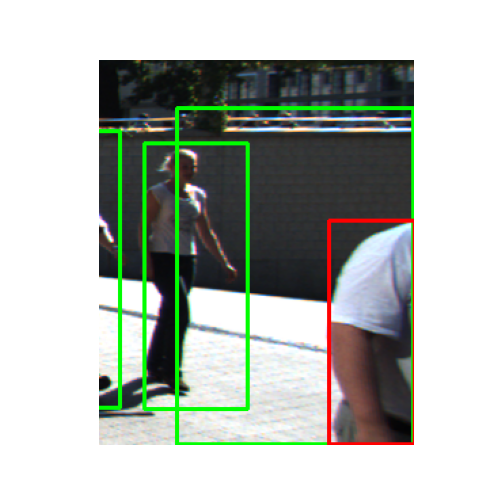}
      \captionsetup{labelformat=empty, hypcap=false}
      \captionof{figure}{000966.png}
    \end{minipage}
\begin{minipage}[t]{0.24\linewidth}
      \centering
      \includegraphics[width=\linewidth, , height=\linewidth, keepaspectratio]{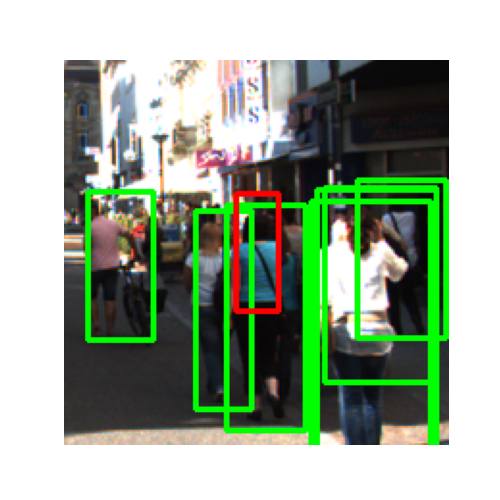}
      \captionsetup{labelformat=empty, hypcap=false}
      \captionof{figure}{002315.png}
    \end{minipage}
\begin{minipage}[t]{0.24\linewidth}
      \centering
      \includegraphics[width=\linewidth, , height=\linewidth, keepaspectratio]{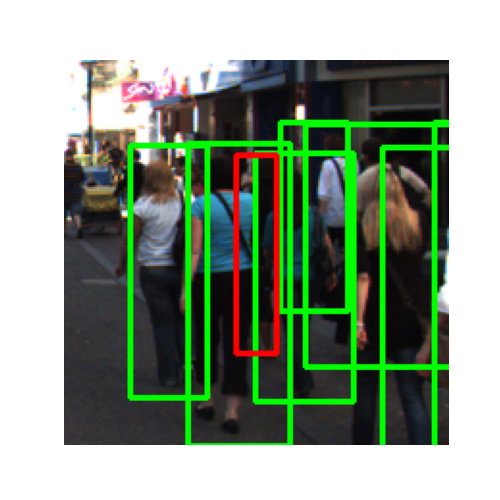}
      \captionsetup{labelformat=empty, hypcap=false}
      \captionof{figure}{002583.png}
    \end{minipage}
\begin{minipage}[t]{0.24\linewidth}
      \centering
      \includegraphics[width=\linewidth, , height=\linewidth, keepaspectratio]{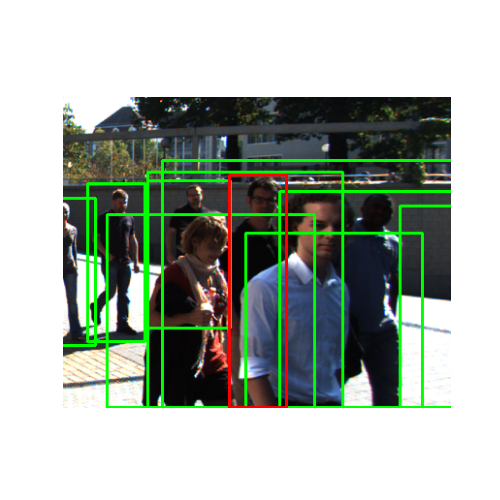}
      \captionsetup{labelformat=empty, hypcap=false}
      \captionof{figure}{002979.png}
    \end{minipage}
\par
\fi

\vspace{18pt}

\textbf{Examples of smaller but unambiguous label errors.} 
Here, we consider a random sample of overlooked pedestrians, that were assigned a soft label probability exceeding $0.8$ and have a bounding box height of less than 40 pixels.

\ifshowimages
\noindent
\begin{minipage}[t]{0.24\linewidth}
      \centering
      \includegraphics[width=\linewidth, , height=\linewidth, keepaspectratio]{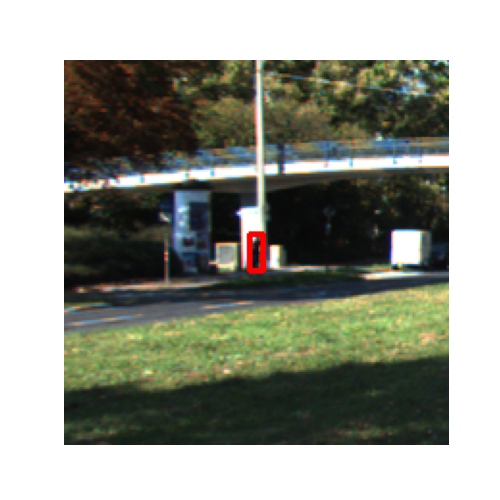}
      \captionsetup{labelformat=empty, hypcap=false}
      \captionof{figure}{000326.png}
    \end{minipage}
\begin{minipage}[t]{0.24\linewidth}
      \centering
      \includegraphics[width=\linewidth, , height=\linewidth, keepaspectratio]{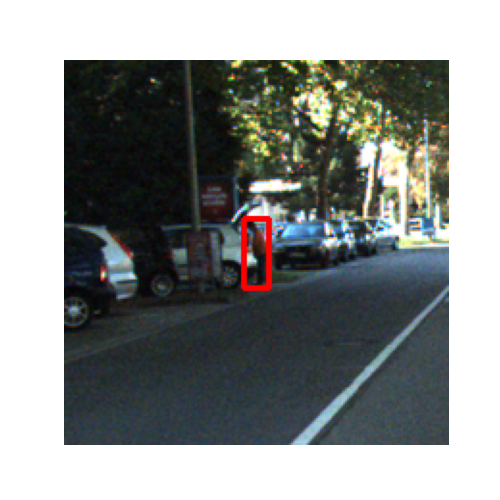}
      \captionsetup{labelformat=empty, hypcap=false}
      \captionof{figure}{002705.png}
    \end{minipage}
\begin{minipage}[t]{0.24\linewidth}
      \centering
      \includegraphics[width=\linewidth, , height=\linewidth, keepaspectratio]{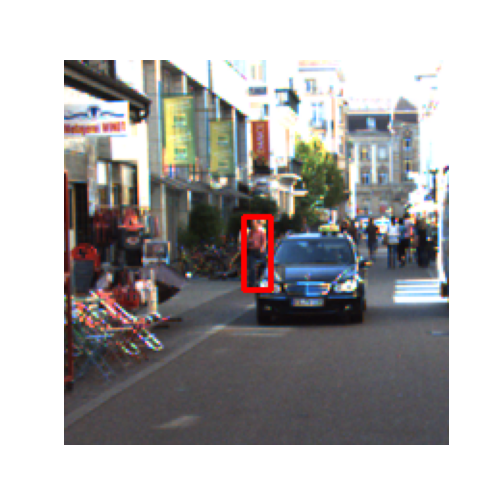}
      \captionsetup{labelformat=empty, hypcap=false}
      \captionof{figure}{003511.png}
    \end{minipage}
\begin{minipage}[t]{0.24\linewidth}
      \centering
      \includegraphics[width=\linewidth, , height=\linewidth, keepaspectratio]{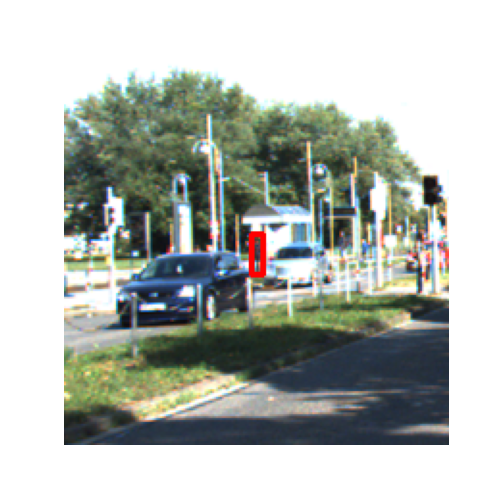}
      \captionsetup{labelformat=empty, hypcap=false}
      \captionof{figure}{000376.png}
    \end{minipage}
\par
\noindent
\begin{minipage}[t]{0.24\linewidth}
      \centering
      \includegraphics[width=\linewidth, , height=\linewidth, keepaspectratio]{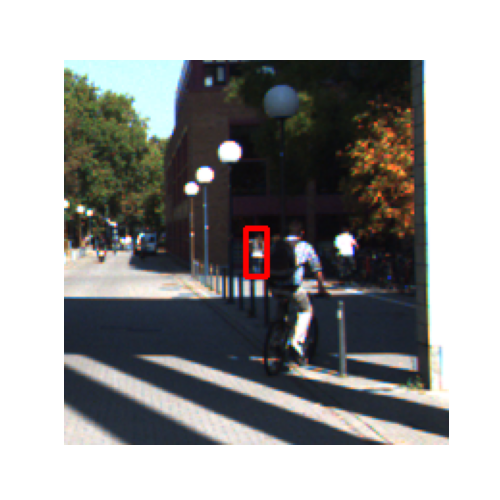}
      \captionsetup{labelformat=empty, hypcap=false}
      \captionof{figure}{004261.png}
    \end{minipage}
\begin{minipage}[t]{0.24\linewidth}
      \centering
      \includegraphics[width=\linewidth, , height=\linewidth, keepaspectratio]{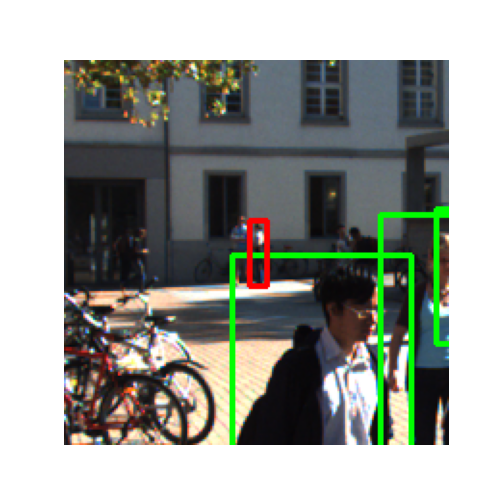}
      \captionsetup{labelformat=empty, hypcap=false}
      \captionof{figure}{002979.png}
    \end{minipage}
\begin{minipage}[t]{0.24\linewidth}
      \centering
      \includegraphics[width=\linewidth, , height=\linewidth, keepaspectratio]{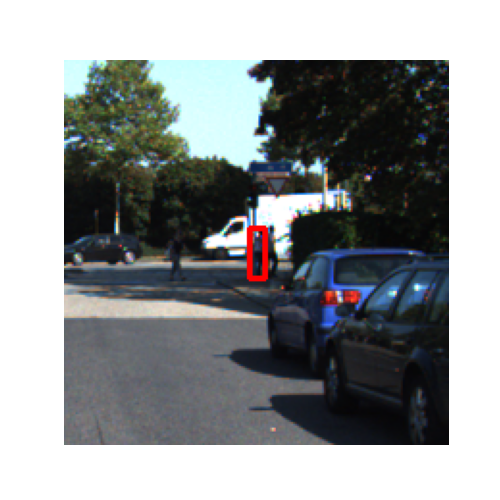}
      \captionsetup{labelformat=empty, hypcap=false}
      \captionof{figure}{003056.png}
    \end{minipage}
\begin{minipage}[t]{0.24\linewidth}
      \centering
      \includegraphics[width=\linewidth, , height=\linewidth, keepaspectratio]{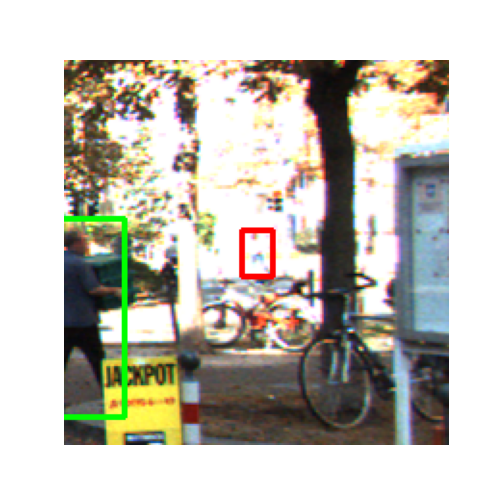}
      \captionsetup{labelformat=empty, hypcap=false}
      \captionof{figure}{004827.png}
    \end{minipage}
\par
\noindent
\begin{minipage}[t]{0.24\linewidth}
      \centering
      \includegraphics[width=\linewidth, , height=\linewidth, keepaspectratio]{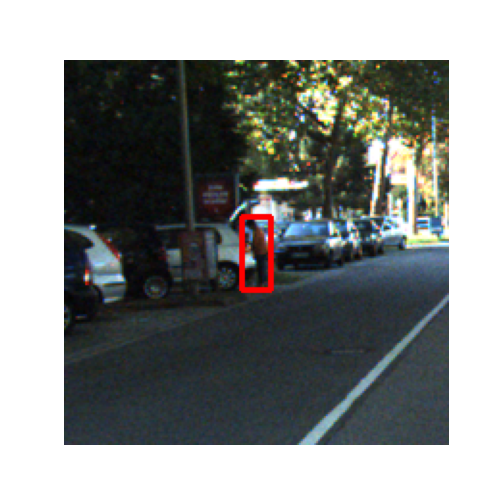}
      \captionsetup{labelformat=empty, hypcap=false}
      \captionof{figure}{000549.png}
    \end{minipage}
\begin{minipage}[t]{0.24\linewidth}
      \centering
      \includegraphics[width=\linewidth, , height=\linewidth, keepaspectratio]{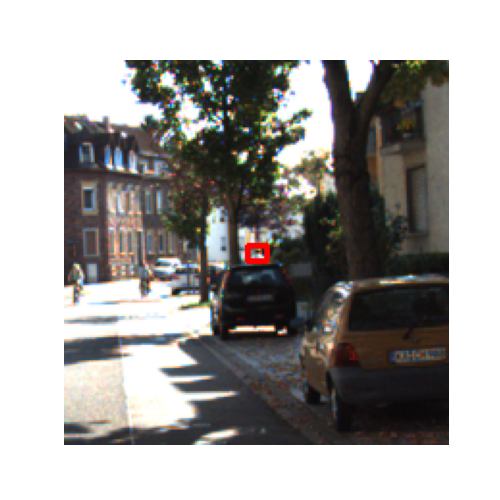}
      \captionsetup{labelformat=empty, hypcap=false}
      \captionof{figure}{004948.png}
    \end{minipage}
\begin{minipage}[t]{0.24\linewidth}
      \centering
      \includegraphics[width=\linewidth, , height=\linewidth, keepaspectratio]{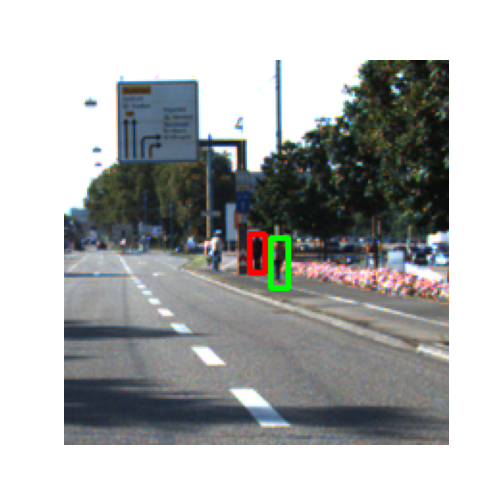}
      \captionsetup{labelformat=empty, hypcap=false}
      \captionof{figure}{005316.png}
    \end{minipage}
\begin{minipage}[t]{0.24\linewidth}
      \centering
      \includegraphics[width=\linewidth, , height=\linewidth, keepaspectratio]{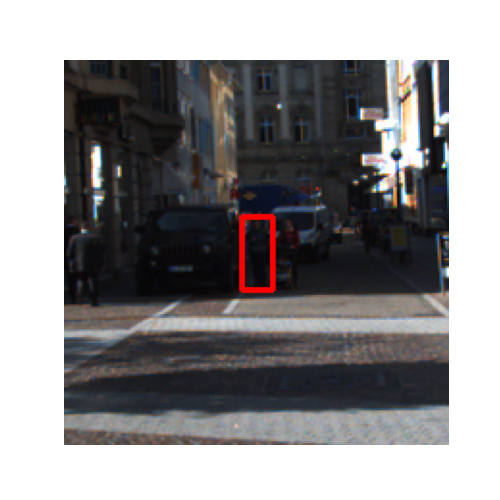}
      \captionsetup{labelformat=empty, hypcap=false}
      \captionof{figure}{006937.png}
    \end{minipage}
\par
\fi

\newpage
\textbf{Examples of somewhat ambiguous label errors.} 
Here, we consider a random sample of overlooked pedestrians, that were assigned a soft label probability between $0.5$ and $0.8$ meaning that they represent somewhat ambiguous instances.

\ifshowimages
\noindent
\begin{minipage}[t]{0.24\linewidth}
      \centering
      \includegraphics[width=\linewidth, , height=\linewidth, keepaspectratio]{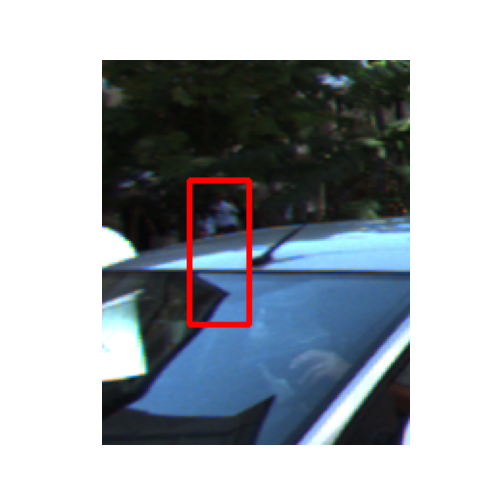}
      \captionsetup{labelformat=empty, hypcap=false}
      \captionof{figure}{000015.png}
    \end{minipage}
\begin{minipage}[t]{0.24\linewidth}
      \centering
      \includegraphics[width=\linewidth, , height=\linewidth, keepaspectratio]{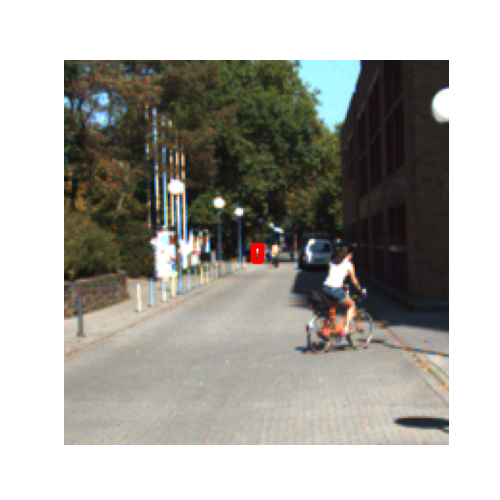}
      \captionsetup{labelformat=empty, hypcap=false}
      \captionof{figure}{003437.png}
    \end{minipage}
\begin{minipage}[t]{0.24\linewidth}
      \centering
      \includegraphics[width=\linewidth, , height=\linewidth, keepaspectratio]{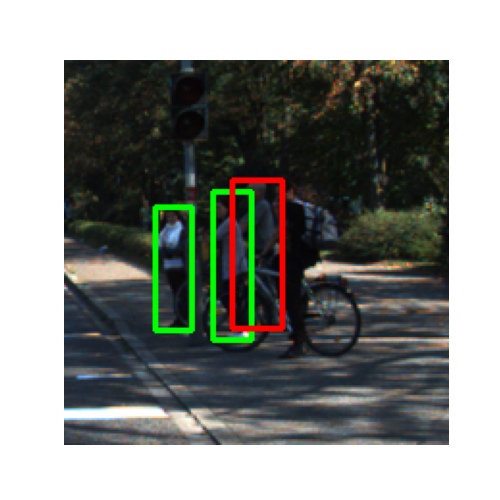}
      \captionsetup{labelformat=empty, hypcap=false}
      \captionof{figure}{006785.png}
    \end{minipage}
\begin{minipage}[t]{0.24\linewidth}
      \centering
      \includegraphics[width=\linewidth, , height=\linewidth, keepaspectratio]{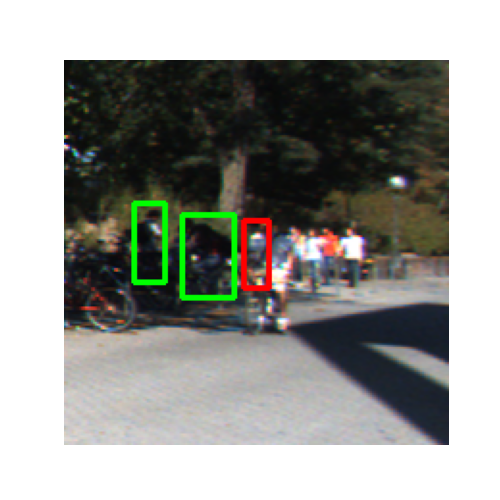}
      \captionsetup{labelformat=empty, hypcap=false}
      \captionof{figure}{002920.png}
    \end{minipage}
\par
\noindent
\begin{minipage}[t]{0.24\linewidth}
      \centering
      \includegraphics[width=\linewidth, , height=\linewidth, keepaspectratio]{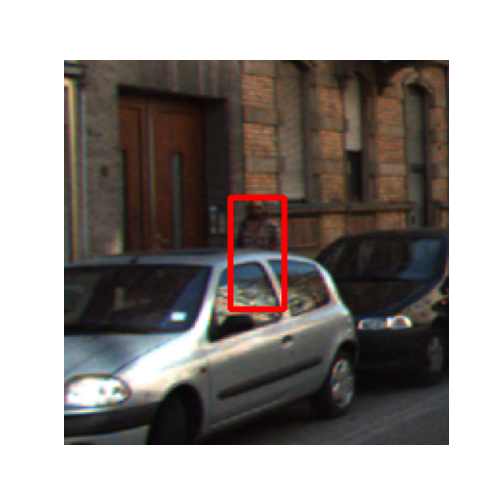}
      \captionsetup{labelformat=empty, hypcap=false}
      \captionof{figure}{000254.png}
    \end{minipage}
\begin{minipage}[t]{0.24\linewidth}
      \centering
      \includegraphics[width=\linewidth, , height=\linewidth, keepaspectratio]{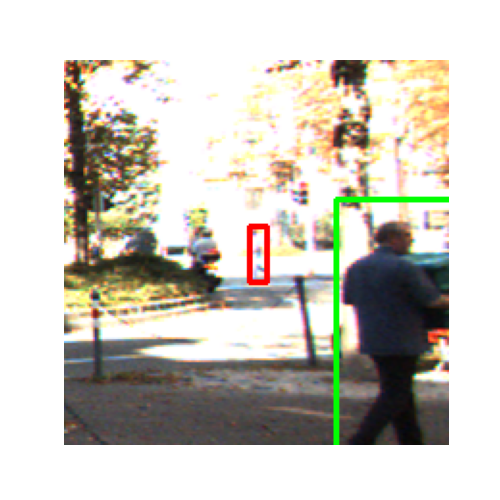}
      \captionsetup{labelformat=empty, hypcap=false}
      \captionof{figure}{006645.png}
    \end{minipage}
\begin{minipage}[t]{0.24\linewidth}
      \centering
      \includegraphics[width=\linewidth, , height=\linewidth, keepaspectratio]{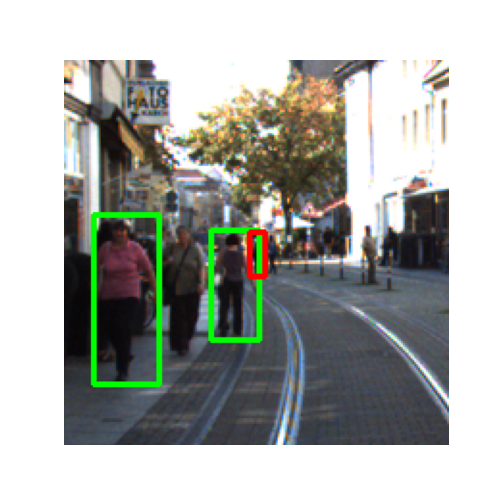}
      \captionsetup{labelformat=empty, hypcap=false}
      \captionof{figure}{001572.png}
    \end{minipage}
\begin{minipage}[t]{0.24\linewidth}
      \centering
      \includegraphics[width=\linewidth, , height=\linewidth, keepaspectratio]{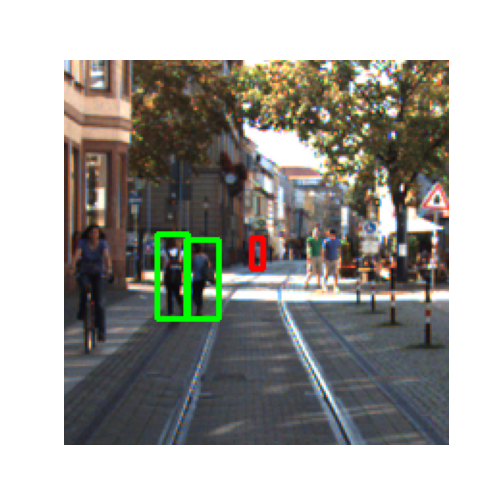}
      \captionsetup{labelformat=empty, hypcap=false}
      \captionof{figure}{002536.png}
    \end{minipage}
\par
\noindent
\begin{minipage}[t]{0.24\linewidth}
      \centering
      \includegraphics[width=\linewidth, , height=\linewidth, keepaspectratio]{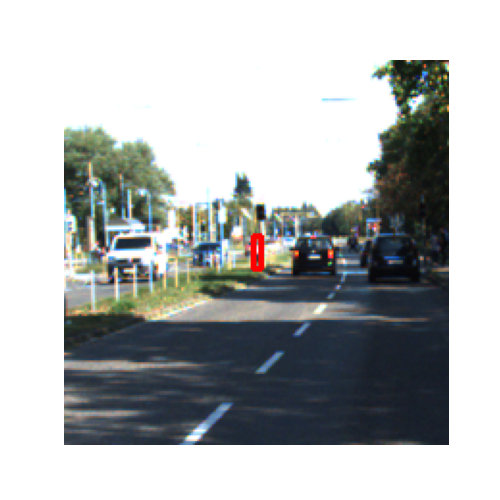}
      \captionsetup{labelformat=empty, hypcap=false}
      \captionof{figure}{003181.png}
    \end{minipage}
\begin{minipage}[t]{0.24\linewidth}
      \centering
      \includegraphics[width=\linewidth, , height=\linewidth, keepaspectratio]{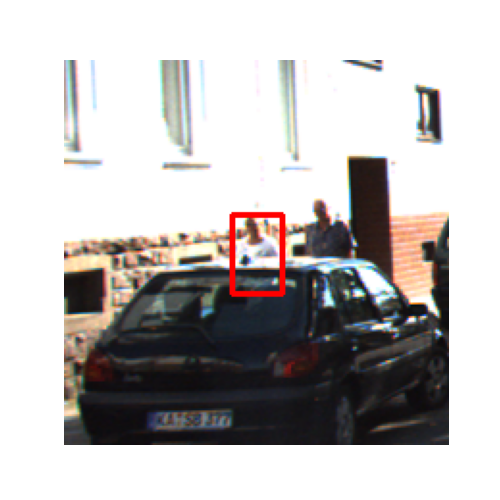}
      \captionsetup{labelformat=empty, hypcap=false}
      \captionof{figure}{004948.png}
    \end{minipage}
\begin{minipage}[t]{0.24\linewidth}
      \centering
      \includegraphics[width=\linewidth, , height=\linewidth, keepaspectratio]{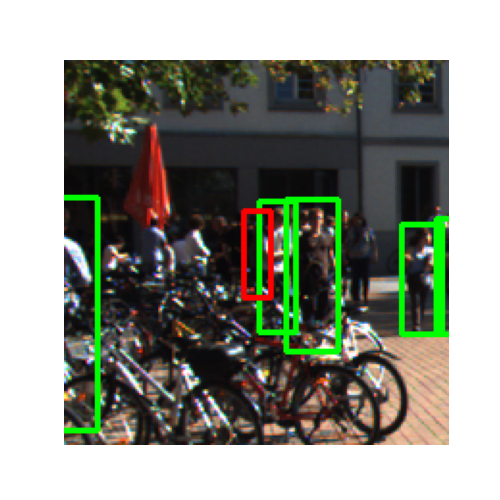}
      \captionsetup{labelformat=empty, hypcap=false}
      \captionof{figure}{000625.png}
    \end{minipage}
\begin{minipage}[t]{0.24\linewidth}
      \centering
      \includegraphics[width=\linewidth, , height=\linewidth, keepaspectratio]{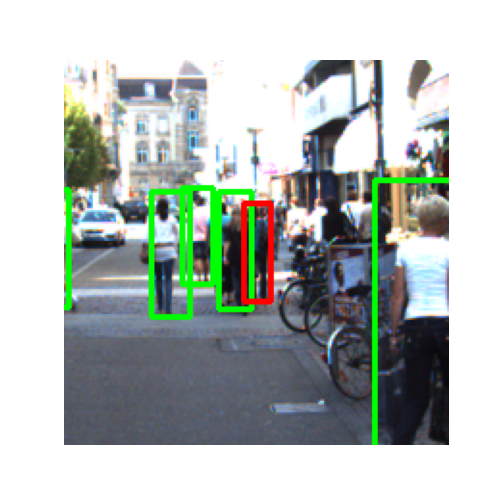}
      \captionsetup{labelformat=empty, hypcap=false}
      \captionof{figure}{006980.png}
    \end{minipage}
\par
\fi


\begin{thebibliography}{10}
\providecommand{\url}[1]{#1}
\csname url@samestyle\endcsname
\providecommand{\newblock}{\relax}
\providecommand{\bibinfo}[2]{#2}
\providecommand{\BIBentrySTDinterwordspacing}{\spaceskip=0pt\relax}
\providecommand{\BIBentryALTinterwordstretchfactor}{4}
\providecommand{\BIBentryALTinterwordspacing}{\spaceskip=\fontdimen2\font plus
\BIBentryALTinterwordstretchfactor\fontdimen3\font minus \fontdimen4\font\relax}
\providecommand{\BIBforeignlanguage}[2]{{%
\expandafter\ifx\csname l@#1\endcsname\relax
\typeout{** WARNING: IEEEtran.bst: No hyphenation pattern has been}%
\typeout{** loaded for the language `#1'. Using the pattern for}%
\typeout{** the default language instead.}%
\else
\language=\csname l@#1\endcsname
\fi
#2}}
\providecommand{\BIBdecl}{\relax}
\BIBdecl

\bibitem{kuutti2020survey}
S.~Kuutti, R.~Bowden, Y.~Jin, P.~Barber, and S.~Fallah, ``A survey of deep learning applications to autonomous vehicle control,'' \emph{IEEE Transactions on Intelligent Transportation Systems}, vol.~22, no.~2, pp. 712--733, 2020.

\bibitem{feng2020deep}
D.~Feng, C.~Haase-Sch{\"u}tz, L.~Rosenbaum, H.~Hertlein, C.~Glaeser, F.~Timm, W.~Wiesbeck, and K.~Dietmayer, ``Deep multi-modal object detection and semantic segmentation for autonomous driving: Datasets, methods, and challenges,'' \emph{IEEE Transactions on Intelligent Transportation Systems}, vol.~22, no.~3, pp. 1341--1360, 2020.

\bibitem{lundervold2019mri}
\BIBentryALTinterwordspacing
A.~S. Lundervold and A.~Lundervold, ``An overview of deep learning in medical imaging focusing on mri,'' \emph{Zeitschrift für Medizinische Physik}, vol.~29, no.~2, pp. 102--127, 2019, special Issue: Deep Learning in Medical Physics. [Online]. Available: \url{https://www.sciencedirect.com/science/article/pii/S0939388918301181}
\BIBentrySTDinterwordspacing

\bibitem{schmarje2021semi-supervised}
L.~Schmarje, M.~Santarossa, S.-M. Schroder, and R.~Koch, ``{A Survey on Semi-, Self- and Unsupervised Learning for Image Classification},'' \emph{IEEE Access}, vol.~9, pp. 82\,146--82\,168, 2021.

\bibitem{hussain2018autonomous}
R.~Hussain and S.~Zeadally, ``Autonomous cars: Research results, issues, and future challenges,'' \emph{IEEE Communications Surveys \& Tutorials}, vol.~21, no.~2, pp. 1275--1313, 2018.

\bibitem{oktay2018attention}
\BIBentryALTinterwordspacing
O.~Oktay, J.~Schlemper, L.~L. Folgoc, M.~Lee, M.~Heinrich, K.~Misawa, K.~Mori, S.~McDonagh, N.~Y. Hammerla, B.~Kainz, B.~Glocker, and D.~Rueckert, ``Attention u-net: Learning where to look for the pancreas,'' in \emph{Medical Imaging with Deep Learning}, 2018. [Online]. Available: \url{https://openreview.net/forum?id=Skft7cijM}
\BIBentrySTDinterwordspacing

\bibitem{schmarje2022mros}
L.~Schmarje, S.~Reinhold, E.~Orwoll, C.-C. Gl{\"{u}}er, and R.~Koch, ``{Opportunistic hip fracture risk prediction in Men from X-ray: Findings from the Osteoporosis in Men (MrOS) Study},'' \emph{Predictive Intelligence in Medicine. PRIME 2022. Lecture Notes in Computer Science, vol 13564}, vol. MICCAI 202, pp. 103--114, 2022.

\bibitem{Kamnitsas2017Efficient}
\BIBentryALTinterwordspacing
K.~Kamnitsas, C.~Ledig, V.~F. Newcombe, J.~P. Simpson, A.~D. Kane, D.~K. Menon, D.~Rueckert, and B.~Glocker, ``Efficient multi-scale 3d cnn with fully connected crf for accurate brain lesion segmentation,'' \emph{Medical Image Analysis}, vol.~36, pp. 61--78, 2017. [Online]. Available: \url{https://www.sciencedirect.com/science/article/pii/S1361841516301839}
\BIBentrySTDinterwordspacing

\bibitem{sun2017revisiting}
C.~Sun, A.~Shrivastava, S.~Singh, and A.~Gupta, ``Revisiting unreasonable effectiveness of data in deep learning era,'' in \emph{Proceedings of the IEEE international conference on computer vision}, 2017, pp. 843--852.

\bibitem{schmarje2023highquality}
L.~Schmarje, V.~Grossmann, T.~Michels, J.~Nazarenus, M.~Santarossa, C.~Zelenka, and R.~Koch, ``Label smarter, not harder: Cleverlabel for faster annotation of ambiguous image classification with higher quality,'' in \emph{German Conference of Pattern Recognition (GCPR)}, 2023, pp. 459--475.

\bibitem{kaur2021survey}
A.~Kaur, Y.~Singh, N.~Neeru, L.~Kaur, and A.~Singh, ``A survey on deep learning approaches to medical images and a systematic look up into real-time object detection,'' \emph{Archives of Computational Methods in Engineering}, pp. 1--41, 2021.

\bibitem{Agnew2024AnnotQual}
C.~Agnew, A.~Scanlan, P.~Denny, E.~M. Grua, P.~van~de Ven, and C.~Eising, ``Annotation quality versus quantity for object detection and instance segmentation,'' \emph{IEEE Access}, vol.~12, pp. 140\,958--140\,977, 2024.

\bibitem{Kovashka2016Crowdsourcing}
A.~Kovashka, O.~Russakovsky, L.~Fei-Fei, and K.~Grauman, ``Crowdsourcing in computer vision,'' \emph{Found. Trends. Comput. Graph. Vis.}, vol.~10, no.~3, p. 177–243, Nov. 2016.

\bibitem{CrowdsourcingApplications}
M.~Stojmenovi{\'{c}}, ``Crowdsourcing applications and techniques in computer vision,'' in \emph{Mobile Crowdsourcing: From Theory to Practice}, J.~Wu and E.~Wang, Eds.\hskip 1em plus 0.5em minus 0.4em\relax Cham: Springer International Publishing, 2023, pp. 409--431.

\bibitem{Marchesoni2023}
\BIBentryALTinterwordspacing
F.~Marchesoni-Acland and G.~Facciolo, ``Iadet: Simplest human-in-the-loop object detection,'' 2023. [Online]. Available: \url{https://arxiv.org/abs/2307.01582}
\BIBentrySTDinterwordspacing

\bibitem{Human-in-the-loop-systems}
\BIBentryALTinterwordspacing
E.~Mosqueira-Rey, E.~Hern\'{a}ndez-Pereira, D.~Alonso-R\'{\i}os, J.~Bobes-Bascar\'{a}n, and A.~Fern\'{a}ndez-Leal, ``Human-in-the-loop machine learning: a state of the art,'' \emph{Artif. Intell. Rev.}, vol.~56, no.~4, p. 3005–3054, Aug. 2022. [Online]. Available: \url{https://doi.org/10.1007/s10462-022-10246-w}
\BIBentrySTDinterwordspacing

\bibitem{Song2023LearningFromNoisyLabels}
\BIBentryALTinterwordspacing
H.~Song, M.~Kim, D.~Park, Y.~Shin, and J.-G. Lee, ``Learning from noisy labels with deep neural networks: A survey,'' \emph{IEEE transactions on neural networks and learning systems}, vol.~34, no.~11, p. 8135—8153, November 2023. [Online]. Available: \url{https://doi.org/10.1109/TNNLS.2022.3152527}
\BIBentrySTDinterwordspacing

\bibitem{Nahum2024LLM}
\BIBentryALTinterwordspacing
O.~Nahum, N.~Calderon, O.~Keller, I.~Szpektor, and R.~Reichart, ``Are llms better than reported? detecting label errors and mitigating their effect on model performance,'' 2024. [Online]. Available: \url{https://arxiv.org/abs/2410.18889}
\BIBentrySTDinterwordspacing

\bibitem{jakubik2024Improve}
\BIBentryALTinterwordspacing
J.~Jakubik, M.~Vössing, M.~Maskey, C.~Wölfle, and G.~Satzger, ``Improving label error detection and elimination with uncertainty quantification,'' 2024. [Online]. Available: \url{https://arxiv.org/abs/2405.09602}
\BIBentrySTDinterwordspacing

\bibitem{Barragan2021QualityQuantity}
A.~M. Barrag{\'{a}}n-Montero, M.~Thomas, G.~Defraene, S.~Michiels, K.~Haustermans, J.~A. Lee, and E.~Sterpin, ``{Deep learning dose prediction for IMRT of esophageal cancer: The effect of data quality and quantity on model performance},'' \emph{Physica Medica}, vol.~83, pp. 52--63, 2021.

\bibitem{deep_fish}
A.~Saleh, I.~H. Laradji, D.~A. Konovalov, M.~Bradley, D.~Vazquez, and M.~Sheaves, ``{A realistic fish-habitat dataset to evaluate algorithms for underwater visual analysis},'' \emph{Scientific Reports}, vol.~10, no.~1, pp. 1--10, 2020.

\bibitem{noisy-labels-comparison}
D.~Karimi, H.~Dou, S.~K. Warfield, and A.~Gholipour, ``{Deep learning with noisy labels: exploring techniques and remedies in medical image analysis},'' \emph{Medical Image Analysis}, vol.~65, 2020.

\bibitem{Wei2021cifar10n}
J.~Wei, Z.~Zhu, H.~Cheng, T.~Liu, G.~Niu, and Y.~Liu, ``{Learning with Noisy Labels Revisited: A Study Using Real-World Human Annotations},'' \emph{ICLR}, pp. 1--23, 2021.

\bibitem{Tomasev2022Relicv2}
N.~Tomasev, I.~Bica, B.~McWilliams, L.~Buesing, R.~Pascanu, C.~Blundell, and J.~Mitrovic, ``{Pushing the limits of self-supervised ResNets: Can we outperform supervised learning without labels on ImageNet?}'' \emph{arXiv preprint arXiv:2201.05119}, 2022.

\bibitem{northcutt2021confident}
C.~Northcutt, L.~Jiang, and I.~Chuang, ``Confident learning: Estimating uncertainty in dataset labels,'' \emph{Journal of Artificial Intelligence Research}, vol.~70, pp. 1373--1411, 2021.

\bibitem{northcutt2021pervasive}
\BIBentryALTinterwordspacing
C.~G. Northcutt, A.~Athalye, and J.~Mueller, ``Pervasive label errors in test sets destabilize machine learning benchmarks,'' in \emph{Thirty-fifth Conference on Neural Information Processing Systems Datasets and Benchmarks Track (Round 1)}, 2021. [Online]. Available: \url{https://openreview.net/forum?id=XccDXrDNLek}
\BIBentrySTDinterwordspacing

\bibitem{thyagarajan2022}
\BIBentryALTinterwordspacing
A.~Thyagarajan, E.~Snorrason, C.~Northcutt, and J.~Mueller, ``Identifying incorrect annotations in multi-label classification data,'' 2022. [Online]. Available: \url{https://arxiv.org/abs/2211.13895}
\BIBentrySTDinterwordspacing

\bibitem{Rottmann2023AutomatedDetection}
\BIBentryALTinterwordspacing
M.~Rottmann and M.~Reese, ``{ Automated Detection of Label Errors in Semantic Segmentation Datasets via Deep Learning and Uncertainty Quantification },'' in \emph{2023 IEEE/CVF Winter Conference on Applications of Computer Vision (WACV)}.\hskip 1em plus 0.5em minus 0.4em\relax Los Alamitos, CA, USA: IEEE Computer Society, Jan. 2023, pp. 3213--3222. [Online]. Available: \url{https://doi.ieeecomputersociety.org/10.1109/WACV56688.2023.00323}
\BIBentrySTDinterwordspacing

\bibitem{hu2022probability}
Z.~Hu, K.~Gao, X.~Zhang, J.~Wang, H.~Wang, and J.~Han, ``Probability differential-based class label noise purification for object detection in aerial images,'' \emph{IEEE Geoscience and Remote Sensing Letters}, vol.~19, pp. 1--5, 2022.

\bibitem{Jeon2025Active}
\BIBentryALTinterwordspacing
Y.~Jeon, K.~Cho, S.~Woo, and E.~Kim, ``A$^2$lc: Active and automated label correction for semantic segmentation,'' 2025. [Online]. Available: \url{https://arxiv.org/abs/2506.11599}
\BIBentrySTDinterwordspacing

\bibitem{Ma2022Correction}
J.~Ma, Y.~Ushiku, and M.~Sagara, ``The effect of improving annotation quality on object detection datasets: A preliminary study,'' in \emph{2022 IEEE/CVF Conference on Computer Vision and Pattern Recognition Workshops (CVPRW)}, 2022, pp. 4849--4858.

\bibitem{Kim2024Correction}
H.~Kim, S.~Hwang, S.~Kwak, and J.~Ok, ``Active label correction for semantic segmentation with foundation models,'' in \emph{Proceedings of the 41st International Conference on Machine Learning}, ser. ICML'24.\hskip 1em plus 0.5em minus 0.4em\relax JMLR.org, 2024.

\bibitem{metadetect}
M.~Schubert, K.~Kahl, and M.~Rottmann, ``{MetaDetect}: {Uncertainty} {Quantification} and {Prediction} {Quality} {Estimates} for {Object} {Detection},'' in \emph{2021 {International} {Joint} {Conference} on {Neural} {Networks} ({IJCNN})}, Jul. 2021, pp. 1--10.

\bibitem{loss-based-method}
M.~Schubert, T.~Riedlinger, K.~Kahl, D.~Kr\"oll, S.~Schoenen, S.~\v{S}egvi\'c, and M.~Rottmann, ``Identifying label errors in object detection datasets by loss inspection,'' in \emph{Proceedings of the IEEE/CVF Winter Conference on Applications of Computer Vision (WACV)}, January 2024, pp. 4582--4591.

\bibitem{objectlab}
U.~Tkachenko, A.~Thyagarajan, and J.~Mueller, ``Objectlab: Automated diagnosis of mislabeled images in object detection data,'' in \emph{ICML Workshop on Data-centric Machine Learning Research}, 2023.

\bibitem{kitti}
A.~Geiger, P.~Lenz, and R.~Urtasun, ``Are we ready for autonomous driving? the kitti vision benchmark suite,'' in \emph{2012 IEEE Conference on Computer Vision and Pattern Recognition}, 2012, pp. 3354--3361.

\bibitem{Castello2020}
V.~O. Castelló, O.~del Tejo~Catalá, I.~S. Igual, and J.-C. Perez-Cortes, ``Real-time on-board pedestrian detection using generic single-stage algorithms and on-road databases,'' \emph{International Journal of Advanced Robotic Systems}, vol.~17, no.~5, p. 1729881420929175, 2020.

\bibitem{Lmd2023GCPR}
T.~Riedlinger, M.~Schubert, S.~Penquitt, J.-M. Kezmann, P.~Colling, K.~Kahl, L.~Roese-Koerner, M.~Arnold, U.~Zimmermann, and M.~Rottmann, ``Lmd: Light-weight prediction quality estimation for object detection in lidar point clouds,'' in \emph{Pattern Recognition}.\hskip 1em plus 0.5em minus 0.4em\relax Cham: Springer Nature Switzerland, 2024, pp. 85--99.

\bibitem{Lmd2025IJCV}
------, ``Lmd: Light-weight prediction quality estimation for object detection in lidar point clouds,'' \emph{International Journal of Computer Vision}, vol. 133, pp. 4349--4365, 02 2025.

\bibitem{Bär2023Noisy}
A.~Bär, J.~Uhrig, J.~P. Umesh, M.~Cordts, and T.~Fingscheidt, ``A novel benchmark for refinement of noisy localization labels in autolabeled datasets for object detection,'' in \emph{2023 IEEE/CVF Conference on Computer Vision and Pattern Recognition Workshops (CVPRW)}, 2023, pp. 3851--3860.

\bibitem{Bernhardt2022Clean}
M.~Bernhardt, D.~Coelho~de Castro, R.~Tanno, A.~Schwaighofer, K.~Tezcan, M.~Monteiro, S.~Bannur, M.~Lungren, A.~Nori, B.~Glocker, J.~Alvarez-Valle, and O.~Oktay, ``Active label cleaning for improved dataset quality under resource constraints,'' \emph{Nature Communications}, vol.~13, 03 2022.

\bibitem{goh2022crowdlab}
H.~W. Goh, U.~Tkachenko, and J.~Mueller, ``Crowdlab: Supervised learning to infer consensus labels and quality scores for data with multiple annotators,'' in \emph{NeurIPS Human in the Loop Learning Workshop}, 2022.

\bibitem{lin2014microsoft}
T.-Y. Lin, M.~Maire, S.~Belongie, J.~Hays, P.~Perona, D.~Ramanan, P.~Doll{\'a}r, and C.~L. Zitnick, ``Microsoft coco: Common objects in context,'' in \emph{European conference on computer vision}.\hskip 1em plus 0.5em minus 0.4em\relax Springer, 2014, pp. 740--755.

\bibitem{GoogleData}
A.~Kuznetsova, H.~Rom, N.~Alldrin, J.~Uijlings, I.~Krasin, J.~Pont-Tuset, S.~Kamali, S.~Popov, M.~Malloci, A.~Kolesnikov, T.~Duerig, and V.~Ferrari, ``The open images dataset v4,'' in \emph{International Journal of Computer Vision}, vol. 128, no.~7, 2020, pp. 1956--1981.

\bibitem{Yun2021Correction}
S.~Yun, S.~J. Oh, B.~Heo, D.~Han, J.~Choe, and S.~Chun, ``Re-labeling imagenet: from single to multi-labels, from global to localized labels,'' in \emph{2021 IEEE/CVF Conference on Computer Vision and Pattern Recognition (CVPR)}, 2021, pp. 2340--2350.

\bibitem{tailception}
J.~Br{\"{u}}nger, S.~Dippel, R.~Koch, and C.~Veit, ``{‘Tailception': using neural networks for assessing tail lesions on pictures of pig carcasses},'' \emph{Animal}, vol.~13, no.~5, pp. 1030--1036, 2019.

\bibitem{mammo-variability}
E.~A. Ooms, H.~M. Zonderland, M.~J.~C. Eijkemans, M.~Kriege, B.~{Mahdavian Delavary}, C.~W. Burger, and A.~C. Ansink, ``{Mammography: Interobserver variability in breast density assessment},'' \emph{The Breast}, vol.~16, no.~6, pp. 568--576, 2007.

\bibitem{schoening2020Megafauna}
T.~Schoening, A.~Purser, D.~Langenk{\"{a}}mper, I.~Suck, J.~Taylor, D.~Cuvelier, L.~Lins, E.~Simon-Lled{\'{o}}, Y.~Marcon, D.~O.~B. Jones, T.~Nattkemper, K.~K{\"{o}}ser, M.~Zurowietz, J.~Greinert, and J.~Gomes-Pereira, ``{Megafauna community assessment of polymetallic-nodule fields with cameras: platform and methodology comparison},'' \emph{Biogeosciences}, vol.~17, no.~12, pp. 3115--3133, 2020.

\bibitem{medicinecrowdsource}
F.~J.~C. dos Reis, S.~Lynn, H.~R. Ali, D.~Eccles, A.~Hanby, E.~Provenzano, C.~Caldas, W.~J. Howat, L.-A. McDuffus, B.~Liu, and Others, ``{Crowdsourcing the general public for large scale molecular pathology studies in cancer},'' \emph{EBioMedicine}, vol.~2, no.~7, pp. 681--689, 2015.

\bibitem{Vasudevan2022DoughBagel}
V.~Vasudevan, B.~Caine, R.~Gontijo-Lopes, S.~Fridovich-Keil, and R.~Roelofs, ``{When does dough become a bagel? Analyzing the remaining mistakes on ImageNet},'' \emph{Advances in Neural Information Processing Systems}, vol.~35, pp. 6720--6734, 2022.

\bibitem{planktonUncertain}
P.~Culverhouse, R.~Williams, B.~Reguera, V.~Herry, and S.~Gonz{\'{a}}lez-Gil, ``{Do experts make mistakes? A comparison of human and machine identification of dinoflagellates},'' \emph{Marine Ecology Progress Series}, vol. 247, pp. 17--25, 2003.

\bibitem{Davani2022BeyondMajority}
A.~M. Davani, M.~D{\'{i}}az, and V.~Prabhakaran, ``{Dealing with Disagreements: Looking Beyond the Majority Vote in Subjective Annotations},'' \emph{Transactions of the Association for Computational Linguistics}, vol.~10, pp. 92--110, 2022.

\bibitem{Basile2021ConsiderDisagree}
V.~Basile, M.~Fell, T.~Fornaciari, D.~Hovy, S.~Paun, B.~Plank, M.~Poesio, and A.~Uma, ``{We Need to Consider Disagreement in Evaluation},'' in \emph{Proceedings of the 1st workshop on benchmarking: past, present and future}, 2021, pp. 15--21.

\bibitem{schmarje2022dc3}
L.~Schmarje, M.~Santarossa, S.-M. Schr{\"{o}}der, C.~Zelenka, R.~Kiko, J.~Stracke, N.~Volkmann, and R.~Koch, ``{A data-centric approach for improving ambiguous labels with combined semi-supervised classification and clustering},'' \emph{Proceedings of the European Conference on Computer Vision (ECCV)}, 2022.

\bibitem{schmarje2022benchmark}
\BIBentryALTinterwordspacing
L.~Schmarje, V.~Grossmann, C.~Zelenka, S.~Dippel, R.~Kiko, M.~Oszust, M.~Pastell, J.~Stracke, A.~Valros, N.~Volkmann, and R.~Koch, ``{Is one annotation enough? A data-centric image classification benchmark for noisy and ambiguous label estimation},'' \emph{Advances in Neural Information Processing Systems}, vol.~35, pp. 33\,215--33\,232, 2022. [Online]. Available: \url{https://proceedings.neurips.cc/paper_files/paper/2022/file/d6c03035b8bc551f474f040fe8607cab-Paper-Datasets_and_Benchmarks.pdf}
\BIBentrySTDinterwordspacing

\bibitem{Peng2014LearningOP}
\BIBentryALTinterwordspacing
P.~Peng, R.~C.-W. Wong, and P.~S. Yu, ``Learning on probabilistic labels,'' in \emph{SDM}, 2014. [Online]. Available: \url{https://api.semanticscholar.org/CorpusID:17158052}
\BIBentrySTDinterwordspacing

\bibitem{Li2024SoftLabelsOD}
Y.~Li, S.~Ma, X.~Jiang, Y.~Luan, and Z.~Jiang, ``Probability based dynamic soft label assignment for object detection,'' \emph{Image and Vision Computing}, vol. 150, p. 105240, 2024.

\bibitem{Gao2024PSTTL}
\BIBentryALTinterwordspacing
Y.~Gao, Y.~Zhang, Z.~Huang, N.~Liu, and D.~Huang, ``Ps-ttl: Prototype-based soft-labels and test-time learning for few-shot object detection,'' in \emph{Proceedings of the 32nd ACM International Conference on Multimedia}, ser. MM '24.\hskip 1em plus 0.5em minus 0.4em\relax New York, NY, USA: Association for Computing Machinery, 2024, p. 8691–8700. [Online]. Available: \url{https://doi.org/10.1145/3664647.3681176}
\BIBentrySTDinterwordspacing

\bibitem{Nguyen2022LabelDistillation}
\BIBentryALTinterwordspacing
C.~H. Nguyen, T.~C. Nguyen, T.~N. Tang, and N.~L.~H. Phan, ``{ Improving Object Detection by Label Assignment Distillation },'' in \emph{2022 IEEE/CVF Winter Conference on Applications of Computer Vision (WACV)}.\hskip 1em plus 0.5em minus 0.4em\relax Los Alamitos, CA, USA: IEEE Computer Society, Jan. 2022, pp. 1322--1331. [Online]. Available: \url{https://doi.ieeecomputersociety.org/10.1109/WACV51458.2022.00139}
\BIBentrySTDinterwordspacing

\bibitem{yolox2021}
Z.~Ge, S.~Liu, F.~Wang, Z.~Li, and J.~Sun, ``Yolox: Exceeding yolo series in 2021,'' \emph{arXiv preprint arXiv:2107.08430}, 2021.

\bibitem{CascadeRCNN}
Z.~Cai and N.~Vasconcelos, ``Cascade r-cnn: Delving into high quality object detection,'' in \emph{Proceedings of the IEEE Conference on Computer Vision and Pattern Recognition (CVPR)}, June 2018.

\bibitem{GoodNWS15}
\BIBentryALTinterwordspacing
B.~M. Good, M.~Nanis, C.~Wu, and A.~I. Su, ``Microtask crowdsourcing for disease mention annotation in pubmed abstracts,'' in \emph{Biocomputing 2015: Proceedings of the Pacific Symposium, Kohala Coast, Hawaii, USA, January 4-8, 2015}, R.~B. Altman, A.~K. Dunker, L.~Hunter, T.~E. Klein, and M.~D. Ritchie, Eds., 2015, pp. 282--293. [Online]. Available: \url{http://psb.stanford.edu/psb-online/proceedings/psb15/good.pdf}
\BIBentrySTDinterwordspacing

\bibitem{mmdetection}
K.~Chen, J.~Wang, J.~Pang, Y.~Cao, Y.~Xiong, X.~Li, S.~Sun, W.~Feng, Z.~Liu, J.~Xu, Z.~Zhang, D.~Cheng, C.~Zhu, T.~Cheng, Q.~Zhao, B.~Li, X.~Lu, R.~Zhu, Y.~Wu, J.~Dai, J.~Wang, J.~Shi, W.~Ouyang, C.~C. Loy, and D.~Lin, ``{MMDetection}: Open mmlab detection toolbox and benchmark,'' \emph{arXiv preprint arXiv:1906.07155}, 2019.

\end{thebibliography}
\end{document}